\documentclass[final,5p,times,twocolumn]{elsarticle}

\usepackage{graphicx}
\usepackage{subcaption}
\usepackage{booktabs}

\usepackage{algorithm,algpseudocode}
\usepackage{multirow}
\usepackage{listings}
\usepackage{caption}

\usepackage[dvipsnames]{xcolor}

\newcommand{\startcompact}[1]{\par\vspace{-0.75em}\begin{#1}%
		\allowdisplaybreaks\ignorespaces}
	
\newcommand{\stopcompact}[1]{\end{#1}\ignorespaces}

\usepackage{amssymb}
\usepackage{amsmath}

\journal{Pattern Recognition}

\begin{document}

\begin{frontmatter}

\title{Part-based Quantitative Analysis for Heatmaps} %

\author[1]{Osman Tursun\corref{cor1}} 
\cortext[cor1]{Corresponding author}
\ead{osman.tursun@qut.edu.au}
\author[2]{Sinan Kalkan} 
\author[1]{Simon Denman}
\author[1]{Sridha Sridharan}
\author[1]{Clinton Fookes}

\address[1]{Signal Processing, Artificial Intelligence and Vision Technologies (SAIVT), \\ Queensland University of Technology, Australia}
\address[2]{Dept. of Computer Eng. and METU ROMER Robotics Center,\\
	Middle East Technical University, Ankara, Turkey}

\begin{abstract}
Heatmaps have been instrumental in helping understand deep network decisions, and are a common approach for Explainable AI (XAI). While significant progress has been made in enhancing the informativeness and accessibility of heatmaps, heatmap analysis is typically very subjective and limited to domain experts. As such, developing automatic, scalable, and numerical analysis methods to make heatmap-based XAI more objective, end-user friendly, and cost-effective is vital. In addition, there is a need for comprehensive evaluation metrics to assess heatmap quality at a granular level.
\end{abstract}


\end{frontmatter}

\section{Introduction}
\label{sec:intro}

A key challenge in deep learning is interpreting predictions made by deep networks \cite{ rio2020understanding,samek2016evaluating,zhou2016learning}. A common approach is to generate a heatmap visualization that highlights the regions of an input image responsible for a network's decision. This approach has seen heatmaps gain significant importance, not only for their interpretability for explainable artificial intelligence (XAI), but also for their utility in weakly supervised localization and segmentation \cite{jo2021puzzle,zhou2016learning}. Existing literature has predominately explored alternative heatmap generation techniques to improve their subjective visual quality and the efficiency of their generation \cite{tursun2022sess}, wherein the localisation accuracy of generated heatmaps has been the focus of considerable attention.

\begin{figure*}[t!]
	\centering
	\includegraphics[width=0.995\textwidth]{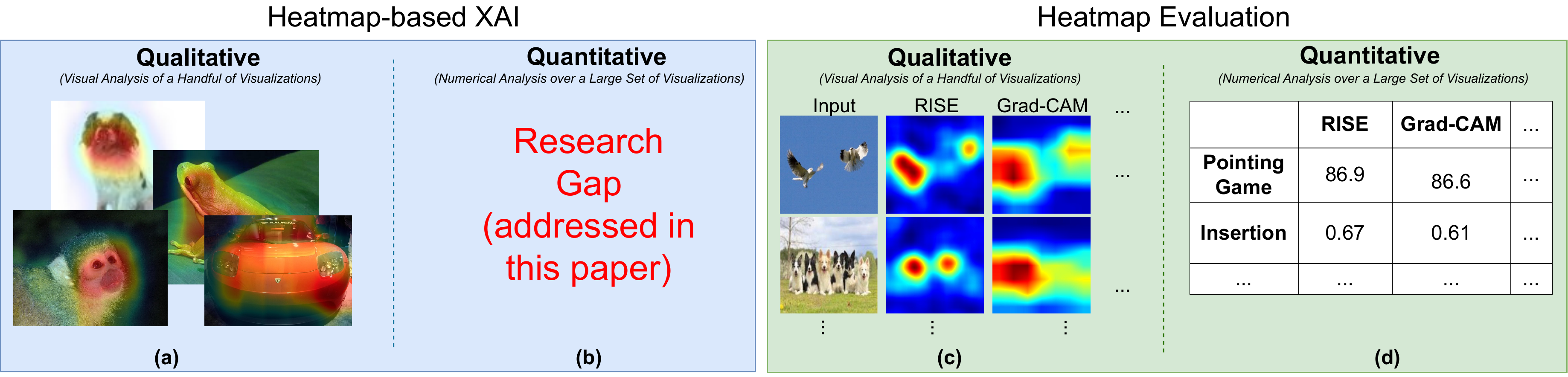}	
	\caption{An overview of the current landscape of heatmap-based XAI and heatmap evaluation methods. Conventional qualitative methods for explaining AI models through heatmaps (a), and evaluating the quality of these heatmaps (c), have predominantly concentrated on visualizing heatmaps for a limited number of examples. In contrast, a quantitative method for explaining AI model performance using heatmaps is currently lacking (b), and the existing quantitative methods for heatmap evaluation (d) rely on simplistic summary statistics which fail to consider detailed fine-grained information.}
	\label{fig:pqah}
\end{figure*}

Although there have been substantial advancements in heatmap generation techniques, research aimed at creating objective, user-friendly, and cost-effective XAI from heatmaps has not garnered equivalent attention or received a comprehensive exploration. We identify several notable research gaps (Figure \ref{fig:pqah}):

\begin{itemize}	
	\item {\bf No Quantitative Approach.} To the best of our knowledge, there is currently no established quantitative approach available for heatmap-based XAI. The absence of a quantitative approach poses challenges for both the scalability and authenticity of heatmap-based XAI.
	
	\item {\bf Reliance on Expert Knowledge.} While heatmaps highlight influential regions or elements that influence a model's prediction, a complete explanation often necessitates deep learning expertise and detailed knowledge of the model. The need for expert knowledge to accurately decipher heatmaps for XAI remains a significant hurdle, restricting their widespread use in XAI applications \cite{kim2022help}. As illustrated in Figure \ref{fig:pqah}a, visualized heatmaps may not be intuitively comprehensible to end-users without expert assistance.

\end{itemize}

Furthermore, the evaluation of heatmap quality remains an area that warrants further enhancement. Despite the presence of both qualitative and quantitative approaches, the following gaps are apparent:

\begin{itemize}
	\item {\bf Lack of Generalization.} Many studies heavily rely on qualitative evaluation methods that involve assessing a limited subset of heatmaps along with explanations provided by the method's authors. This approach is illustrated in Figure \ref{fig:pqah}c. It introduces inherent biases and does not scale to large datasets or multiple experiments, even though it provides granular information. Addressing this issue is essential for achieving unbiased and scalable qualitative evaluation processes.
	
	\item {\bf Lack of Granularity.} Presently, commonly used quantitative evaluation metrics such as insertion and deletion scores \cite{petsiuk2018rise} and pointing accuracy \cite{zhang2018top} primarily provide an overarching assessment of heatmap localization performance, but lack granular information that considers how well different object parts are covered by the heatmap. For instance, the quantitative results presented in Figure \ref{fig:pqah}d demonstrate that RISE outperforms Grad-CAM overall. However, these measures do not provide insights about the qualities of the heatmaps when considering different classes or specific regions of those classes. %
\end{itemize}

To address the aforementioned gaps, we propose a novel heatmap analysis approach called Part-based Quantitative Analysis of Heatmaps (PQAH)\footnote{Pronounced as ``pikah".}. To be specific, PQAH analyses the distribution of heatmaps over classes and their constituent parts, providing detailed quantitative insights. PQAH analysis can be used as an alternative evaluation measure to compare heatmaps extracted with different heatmap visualisation approaches. Similar to qualitative heatmap evaluation approaches, PQAH incorporates fine-grained details while maintaining objectivity and scalability, like existing quantitative evaluation methods. In addition however, PQAH can provide granular quantitative information to make heatmap-based XAI more objective, transparent, thus enabling quantitative assessment and improvement. For example, AI experts can leverage these numerical results to enhance the training pipeline and network architecture, while general end-users can access a crucial XAI report with the assistance of a large language model.

The only requirement of PQAH is a set of images with part-based annotations. Part annotations can be acquired via manual or automatic part segmentation. Although achieving fully automatic part-based segmentation is a challenging task, recent studies such as VLPART \cite{peize2023vlpart} and Semantic-SAM \cite{li2023semantic} have shown promising progress in this regard. Furthermore, PQAH requires a relatively small dataset, such as a test set, making manual labeling feasible and cost-effective.

In this paper, our contributions can be summarized as follows:

\begin{itemize}
	\item We introduce PQAH, a novel quantitative heatmap analysis approach that provides semantic and granular quantitative analysis, distinguishing it from existing approaches.

    \item We conduct a series of experiments to showcase the utility of PQAH in both heatmap-based XAI and heatmap evaluation.

    \item We perform experiments for generating user-friendly Explainable AI (XAI) reports and enhancing the model based on the training strategy obtained from the PQAH analysis. These efforts demonstrate PQAH's practical utility in addressing real-world problems.

\end{itemize}

\section{Related Work}
\label{sec:lit}

\textbf{Heatmap-based XAI.} In this study, our primary focus centers on heatmap-based eXplainable Artificial Intelligence (XAI) approaches tailored for black-box models, in particular neural networks. Following the success of Deep Neural Networks (DNNs), a multitude of techniques designed to generate heatmaps to help understand neural network behaviour have been proposed. Broadly, these techniques can be categorized into three main groups: gradient-based methods \cite{simonyan2013deep, sundararajan2017axiomatic}, class activation-based methods \cite{zhou2016learning, selvaraju2017grad}, and perturbation-based methods \cite{fong2017interpretable, dabkowski2017real}. Much research has been dedicated to improving the efficiency and quality of heatmap generation, with related studies making significant strides through the development of advanced techniques \cite{tursun2022sess, smilkov2017smoothgrad, wang2020score, petsiuk2018rise}.

{\noindent}\textbf{Assessing Heatmaps.} Concurrent with the development of methods to extract heatmaps, their assessment has become a subject of considerable attention. Evaluating heatmaps remains a challenging and ongoing endeavor of great importance. Researchers have put forth both qualitative and quantitative evaluation methods.

Qualitative assessment involves two main approaches. The first relies on aligning human prior knowledge with the regions highlighted by heatmaps, as seen in e.g. \cite{ribeiro2016should} and \cite{selvaraju2017grad}. The second measures how effectively the regions highlighted by heatmaps contribute to improving human decision-making, as demonstrated in \cite{nguyen2021effectiveness}. Qualitative assessment gives a high priority to the human perspective, which is inherently subjective and lacks scalability. 

In contrast, quantitative approaches offer automation, scalability, and objectivity, but the outcome of the quantitative approach might contradict human intuition, which may lead to the conclusion that a metric is poor as it does not conform to expectations. There are primarily two types of quantitative approach: \textit{supervised localisation based} methods and \textit{perturbation} methods. Supervised localisation methods leverage ground-truth object bounding box annotations and calculate global scores such as the Intersection over Union \cite{zhou2016learning}, Pointing Accuracy \cite{zhang2018top} and percentage of meaningful pixels outside the ground-truth bounding box \cite{rio2020understanding}. The second approach is perturbation-based methods. In these approaches, either all the important pixels are perturbed simultaneously, or the pixels are perturbed progressively in an ascending/descending order as suggested by the heatmaps \cite{chattopadhay2018grad, samek2016evaluating}. The changes in network output that result from this perturbation are then measured, and larger instant changes are considered to be indicative of better heatmaps. Several metrics including  Average Drop \cite{chattopadhay2018grad}, Area Over/Under MoRF/LoRF Curves \cite{samek2016evaluating,ancona2017unified}, and insertion and deletion scores \cite{petsiuk2018rise} are widely used for this assessment. 
PQAH could effectively evaluate heatmaps in a way comparable to supervised localization-based metrics. However, it offers a level of fine-grained differentiation that these metrics lack.

{\noindent}\textbf{Automating the Interpretation of Heatmaps} 
Another main use case of PQAH is automating the interpretability of the heatmaps themselves, which is less studied compared to heatmap generation and evaluation tasks. However, it's worth noting that recent works in the field have shown an increasing interest in automating the interpretation of heatmaps. For instance, research on verbalizing heatmaps \cite{feldhus2023saliency, tursun2023towards} and raising awareness about heatmap interpretability \cite{kim2022help} highlight this growing interest. Enhancing the interpretability of heatmaps holds significant value for various applications of XAI across different tasks. For instance, we generate end-user-friendly XAI reports that include comprehensive, readable analyses and technical suggestions for AI model improvement. In this study, we have produced such XAI reports by utilizing PQAH results.

\section{Part-based Quantitative Analysis for Heatmaps}
\label{sec:meth}

Heatmaps are a pivotal tool for interpreting the decision-making processes of Deep Neural Networks (DNNs), highlighting the input features that significantly influence the network's output. However, the intricate and often irregular patterns within heatmaps make quantitative analysis and comparison a formidable challenge, particularly when assessing multiple heatmaps across various test cases.

Our proposed method, PQAH, seeks to bridge the gap between qualitative interpretation and quantitative assessment, emulating the human approach to understanding heatmaps but with a numerical analysis. How do humans intuitively evaluate the quality of a heatmap? As pointed out by Samek et al. \cite{samek2016evaluating}, this typically involves aligning the heatmap with a prior understanding of what is considered relevant. In our approach, we posit that the semantic segmentation of an object's parts constitutes such a prior, which is essential for both explanation and quantitative evaluation. PQAH measures the degree of overlap between the heatmap and the semantic part segmentation, leveraging part annotations to quantify the heatmap activation associated with each object part. These annotations are thus integral to PQAH, enabling a detailed and quantifiable analysis of heatmap activations.

\begin{figure*}[tbh!]
	\centering
	\includegraphics[width=0.95\linewidth]{"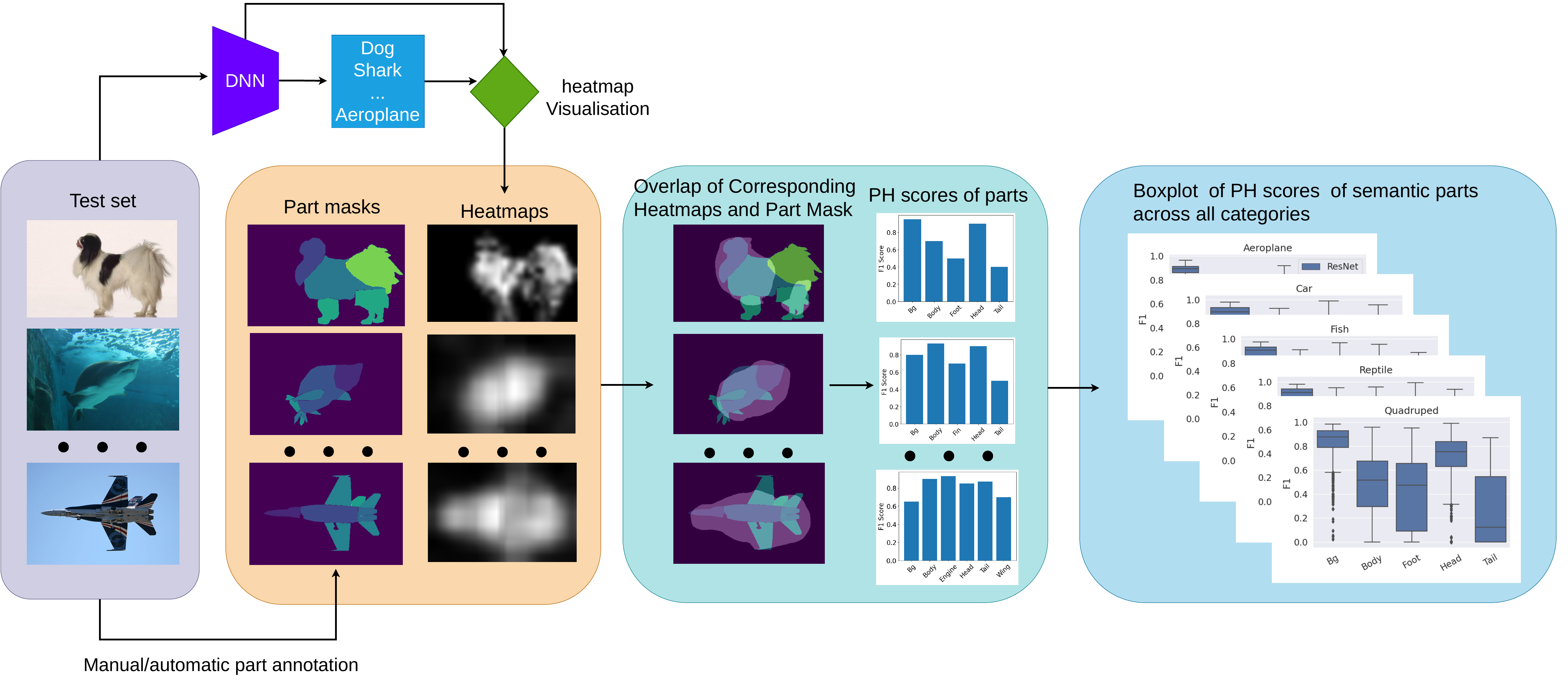"}
	\caption{Overview of the PQAH (Part-based Quantitative Analysis for Heatmaps) framework. The process involves (1) extracting part masks and heatmaps from the given image dataset, (2) computing the PQAH Overlap scores for each semantic part of the main object in the images, and (3) aggregating $\mathit{PH}$ scores across all semantic part categories to generate statistical summaries and visual representations.}%
	\label{fig:pqah_diagram}
\end{figure*}

The implementation of PQAH is visually depicted in Figure \ref{fig:pqah_diagram}. The process encompasses three main steps:

\begin{enumerate}
	\item Preparation of heatmaps and part-annotation masks for a designated set of images. Part-annotations are manually delineated to compensate for the shortcomings of automated part annotation methods. Heatmaps are generated using any standard heatmap visualization technique.
	\item Application of PQAH to obtain numerical results. For detailed information, refer to Section \ref{sec:pqah1}.
	\item Summarization and visualization of the numerical results obtained in step 2. For an in-depth explanation, see Section \ref{sec:pqah2}.
\end{enumerate}

\subsection{PQAH Algorithm}
\label{sec:pqah1}
\textbf{Problem Definition and Notation.} PQAH aims to quantify the overlap between each semantic part, $H^p$, in a heatmap $H$, and each corresponding part mask, $M^p$, for each image, $I$, from an image set, $D$. Here, $H^p$ represents the ground-truth portion of the heatmap, $H$, corresponding to part $p$. However, $H^p$ could not be generated by the network because its training is focused on processing objects rather than individual parts. $M^p$ denotes the binary mask for part $p$, with $M^p \in \{0, 1\}^{w \times h}$ where $w$ and $h$ are the width and height of $I$, respectively. The part masks, $\{M^p \in \{0, 1\}, p \in parts(I)\}$, correspond to distinct semantic parts within $I$, and the foreground mask $M = \sum_{p \in parts(I)} M^p$. Although the individual part heatmaps, $H^p$, are unknown, the composite heatmap, $H$, extracted from a DNN, $\mathcal{F}$, for image, $I$, by assigning each pixel, $i$ in $I$, a value $\mathcal{H}(\mathcal{F}, I, i) \in [0,1]$, according to some heatmap extraction method, $\mathcal{H}$. Please note that $H$ is binarized using a threshold, $\theta$, set by default to $0.5$. This default setting is based on the assumption that the calculation of $H$ incorporates a normalization step, such as min-max normalization, ensuring that its values are appropriately scaled.

{\bf Measuring PQAH Overlap.} PQAH Overlap refers to the overlap between $(M^p, H^p)$. This is denoted as $\mathit{PH}(M^p, H^p)$ for clarity. we can use evaluation measures commonly used in semantic segmentation, such as the Dice coefficient, Intersection over Union (IoU), and the $F1$ Score, due to their conceptual similarity to our task. However, due to the absence of $H^p$, the Dice coefficient and IoU may not be readily applicable to PQAH, as their calculation relies on the absolute value of $|H^p \cup M^p|$, $|H^p \cap M^p|$ and $|H^p|$. Therefore, we adopt the $F1$ Score as our evaluation measure, defined as:
\startcompact{small}
\begin{equation}
	\mathit{PH(M^P, H^P)} = \mathit{F_1(M^P, H^P)} = \frac{2 \cdot \mathit{Precision} \cdot \mathit{Recall}}{\mathit{Precision} + \mathit{Recall}}.
\end{equation}
\stopcompact{small}
The $Recall$ of a part $p$ can be easily defined as:
\startcompact{small}
\begin{equation}
	\mathit{Recall(M^p, H)}  = \frac{\mathit{TP}}{\mathit{TP} + \mathit{FN}},
\end{equation}
\stopcompact{small}
{\noindent}where $\mathit{TP} = \sum (M^p \odot H)$ represents the true positives, and $\mathit{FN} = \sum M^p - \mathit{TP}$ denotes the false negatives.

Precision for part $p$ is not directly observable since $H^p$ is not available. Nevertheless, given that both $\mathit{Precision}$ and $\mathit{Recall}$ are normalized between 0 and 1, we can approximate the $\mathit{Precision}$ for part $p$ using the overall Precision for the object in image $I$ to which part $p$ belongs. The approximate $\widetilde{\mathit{Precision}}$ is given by:
\startcompact{small}
\begin{equation}
	\mathit{\widetilde{Precision}(M, H)} = \frac{\mathit{TP}}{\mathit{TP} + \mathit{FP}},
\end{equation}
\stopcompact{small}
{\noindent}where $\mathit{TP} = \sum (M \odot H)$ and $\mathit{FP} = \sum H - \mathit{TP}$ are the true positives and false positives, respectively.

As a result, the $PH$ scores will range from 0 to 1, with higher values indicating a more accurate representation by the heatmap.

After computing the $\mathit{PH}$ scores for all parts within an image, the $\mathit{PH}$ score for the background is also determined to assess the heatmap's effectiveness in distinguishing between the foreground and the background. The heatmap for the background is represented as $1-H$, and the corresponding mask for the background is denoted as $1-M$. The computation of the $PH$ score for the background follows the same procedure as for the parts. However, the $\mathit{Precision}$ for the background can be directly calculated without the need for approximation, as the $\mathit{PH}$ score for the background is calculated at the object level.

\subsection{Summarising and Visualising PAQH Results}
\label{sec:pqah2}
The final step involves summarizing the $F_1$ scores across semantic parts of all categories by calculating numerical statistics, including Q1 (first quartile), median/Q2, and Q3 (third quartile). These are denoted as $\mathit{PH_{Q1}}$, $\mathit{PH_{Q2}}$, and $\mathit{PH_{Q3}}$ representively. These numerical metrics offer granular insights into the spatial distribution of heatmaps across the semantic parts of a class. They, therefore, are invaluable in the context of heatmap-based eXplainable Artificial Intelligence (XAI), where they facilitate quantitative explanations, contributing to the automation and objectivity of the XAI process. Furthermore, these metrics serve a dual purpose, as they can be employed to assess the quality of heatmaps. To deliver a holistic and insightful analysis, we have chosen to employ boxplots as the preferred visualization method for these numerical statistics.

\subsection{Critical Reporting with a LLM}
\label{sec:report}
The comprehensive numerical summary and accompanying graphical representation produced by PQAH facilitate an automated, scalable, objective, and accessible approach to XAI. Nevertheless, it is additionally helpful to automatically extract insights from PQAH's outputs. Such insights can involve pinpointing the strengths and weaknesses of the DNN and developing strategies to address its limitations, thus augmenting the practical value of XAI for end-users, especially for those who are not experts in deep learning. To accomplish this, employing the latest advancements in large-language models (LLM), such as GPT-4 \cite{openai2023gpt4}, proves invaluable. A large-language model like GPT-4 can not only distill PQAH's numerical data into a concise text-based XAI report but can also offer expert-level suggestions for potential enhancements. 

To create an XAI report using PQAH results with GPT-4, a JSON file containing the $PH_{Q1}$, $PH_{Q2}$, and $PH_{Q3}$ scores for all parts across all categories is input into GPT-4 with prompts designed to elicit an analysis of the DNN's strengths, weaknesses, and potential improvements. GPT-4 then processes this information to generate a concise, insightful XAI report, providing an expert-level understanding of the DNN's performance alongside actionable recommendations.

\section{Experimental Setup } 
\subsection{Datasets}

In this work, we use the PartImageNet \cite{he2022partimagenet} and PASCAL-Part \cite{chen2014detect} datasets for evaluation. Both datasets provide part-based annotations, which are segmentation masks of individual parts of objects. It is important to note that both datasets were initially designed for tasks related to segmenting object parts at a granular level. However, in our study, we utilise these datasets to assess PQAH's effectiveness.

{\bf{PartImageNet}} includes 11 super-categories which are created by grouping 158 classes from the original ImageNet dataset. We use the test set ($4,598$ images) of PartImageNet for evaluation. Each image typically includes a main object, and segmentation masks for two-four part-level annotations. For the complete part taxonomy, please refer to \cite{peize2023vlpart, he2022partimagenet}.

{\bf{PASCAL-Part}} is a modified version \cite{peize2023vlpart} of the original PASCAL-part dataset \cite{chen2014detect}. The original dataset is created by annotating parts for the PASCAL VOC 2010 dataset. The modified version includes 14 object categories in the validation set ($4,465$ images), and each category has between two and 12 parts. Note that we use the validation set (rather than the test set) as PASCAL does not have a public test set. %

\begin{figure*}[tb]
  \centering
  \begin{subfigure}{\textwidth}
    \centering
    \begin{minipage}{\textwidth}
      \begin{subfigure}{0.325\textwidth}
        \centering
        \includegraphics[width=\linewidth]{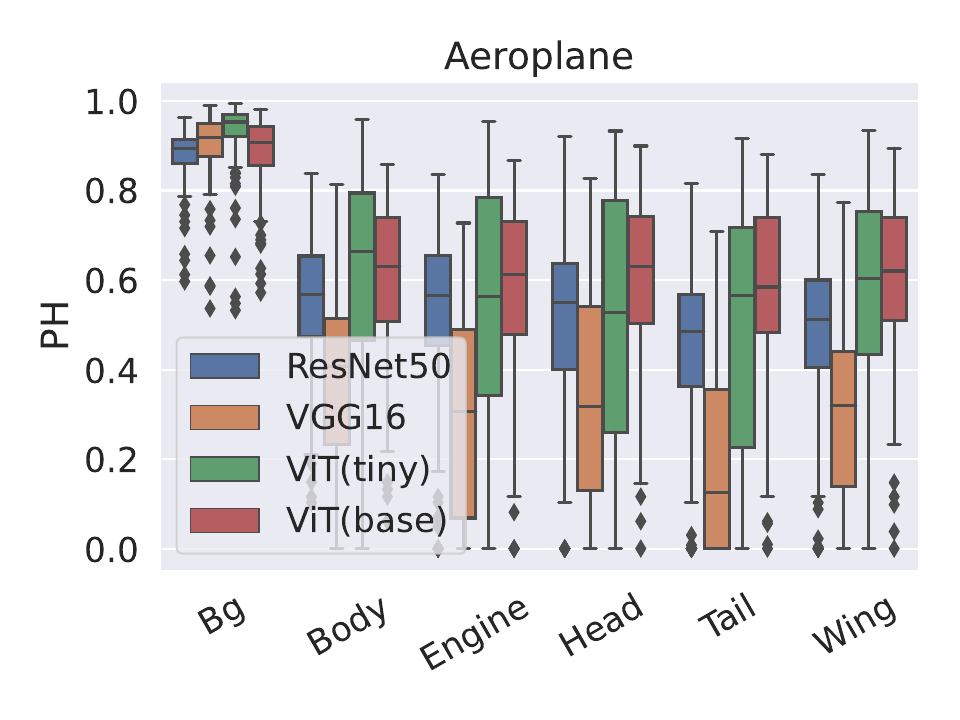}
      \end{subfigure}
      \hfill %
      \begin{subfigure}{0.325\textwidth}
        \centering
        \includegraphics[width=\linewidth]{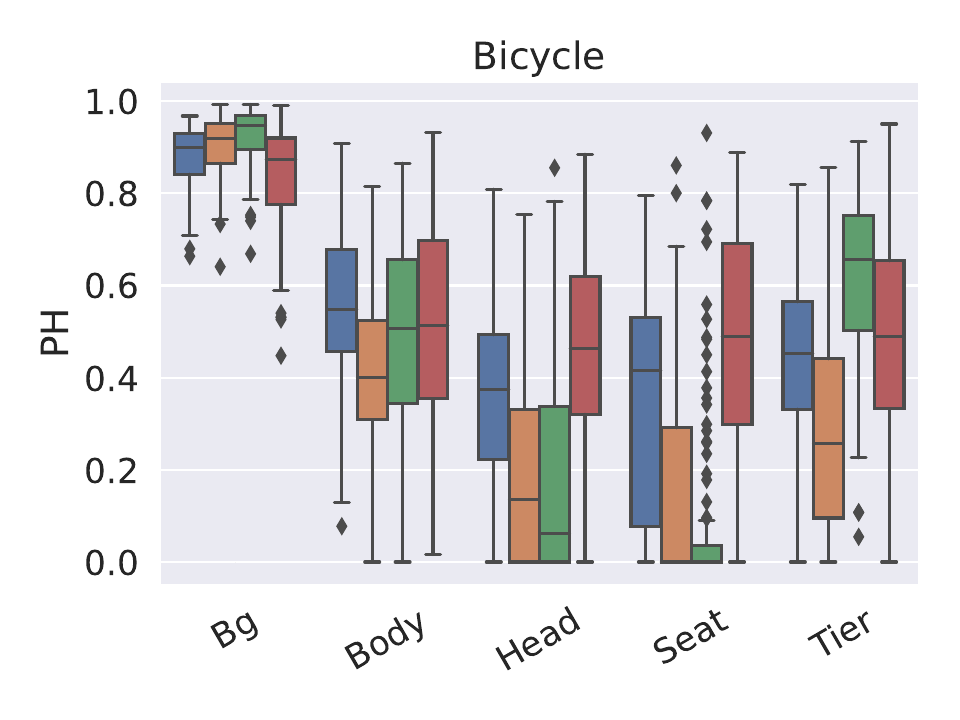}
      \end{subfigure}
      \hfill %
      \begin{subfigure}{0.325\textwidth}
        \centering
        \includegraphics[width=\linewidth]{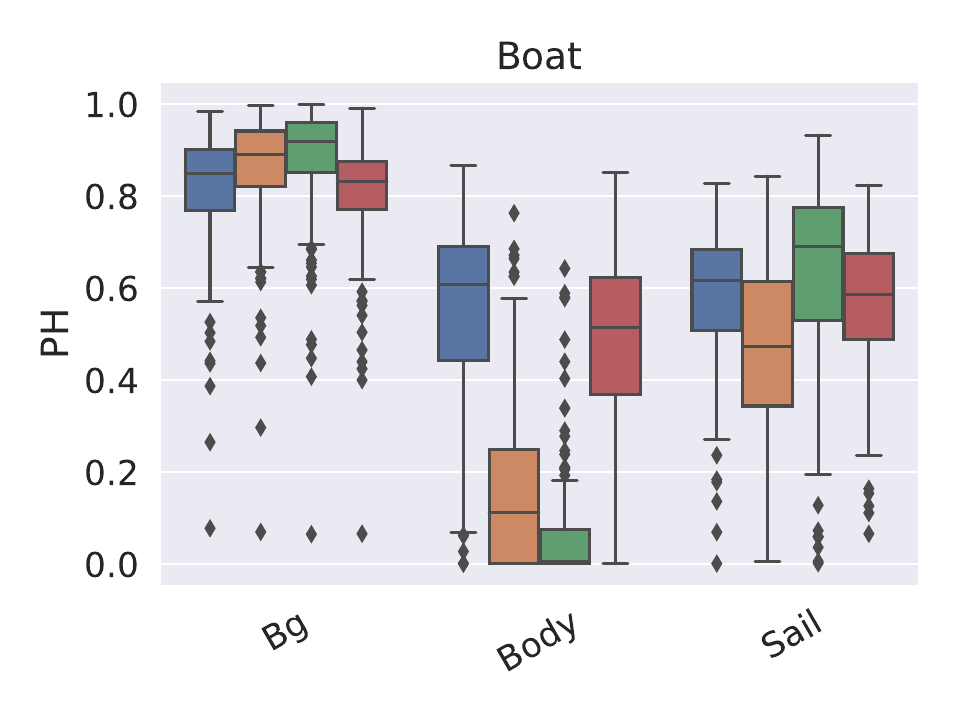}
      \end{subfigure}
    \end{minipage}
    \caption{PartImageNet}
    \label{fig:partimagenet_f1}
  \end{subfigure}

  \vspace{1em} %

  \begin{subfigure}{\textwidth}
    \centering
    \begin{minipage}{\textwidth}
      \begin{subfigure}{0.325\textwidth}
        \centering
        \includegraphics[width=\linewidth]{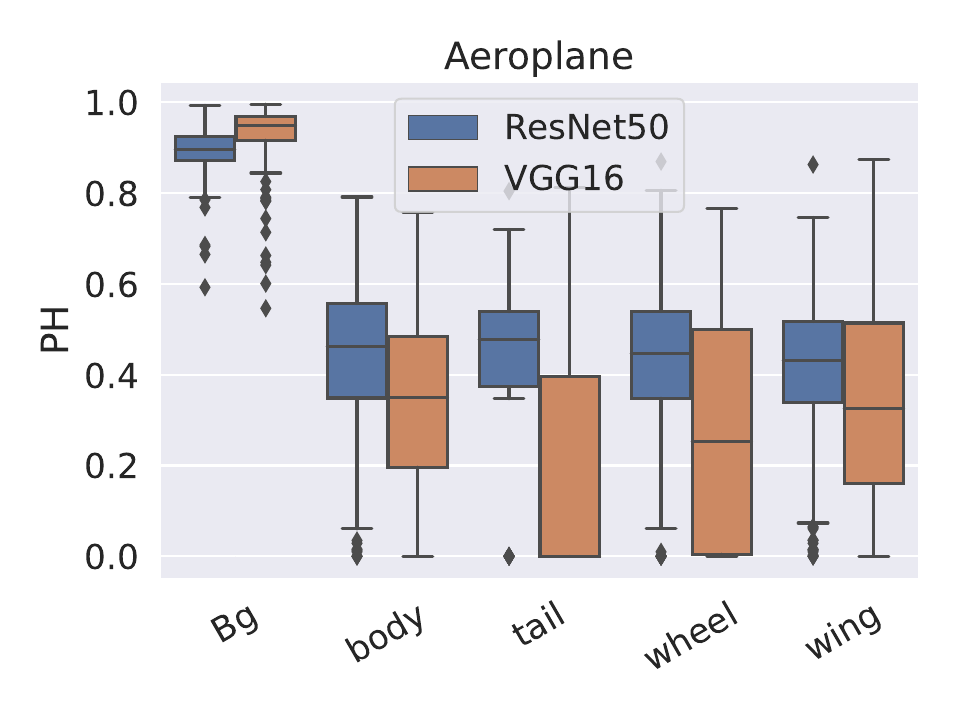}
      \end{subfigure}
      \hfill %
      \begin{subfigure}{0.325\textwidth}
        \centering
        \includegraphics[width=\linewidth]{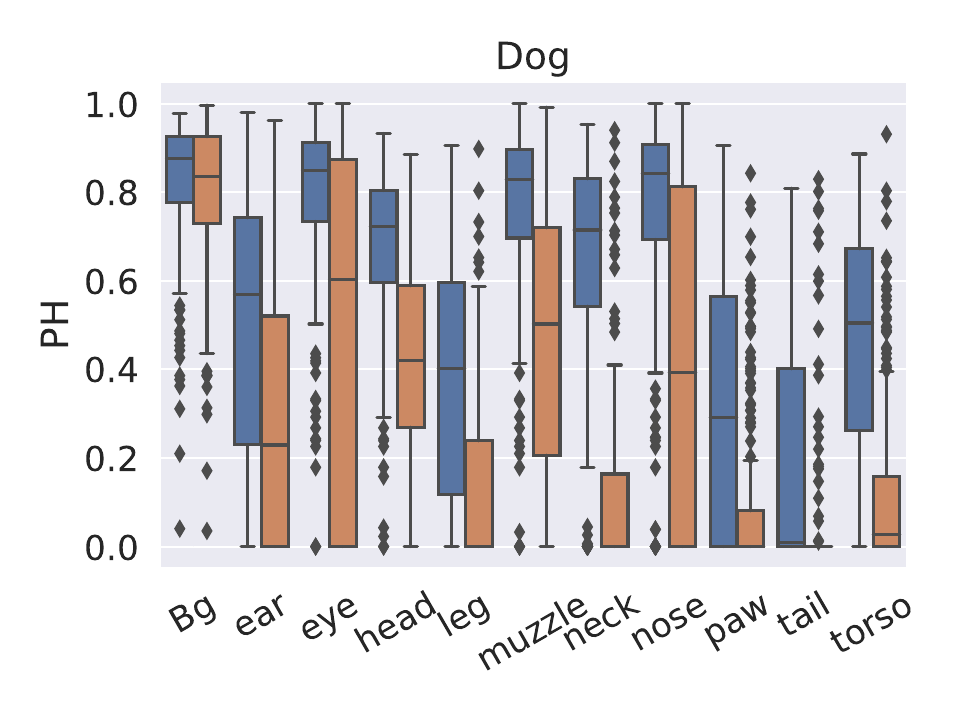}
      \end{subfigure}
      \hfill %
      \begin{subfigure}{0.325\textwidth}
        \centering
        \includegraphics[width=\linewidth]{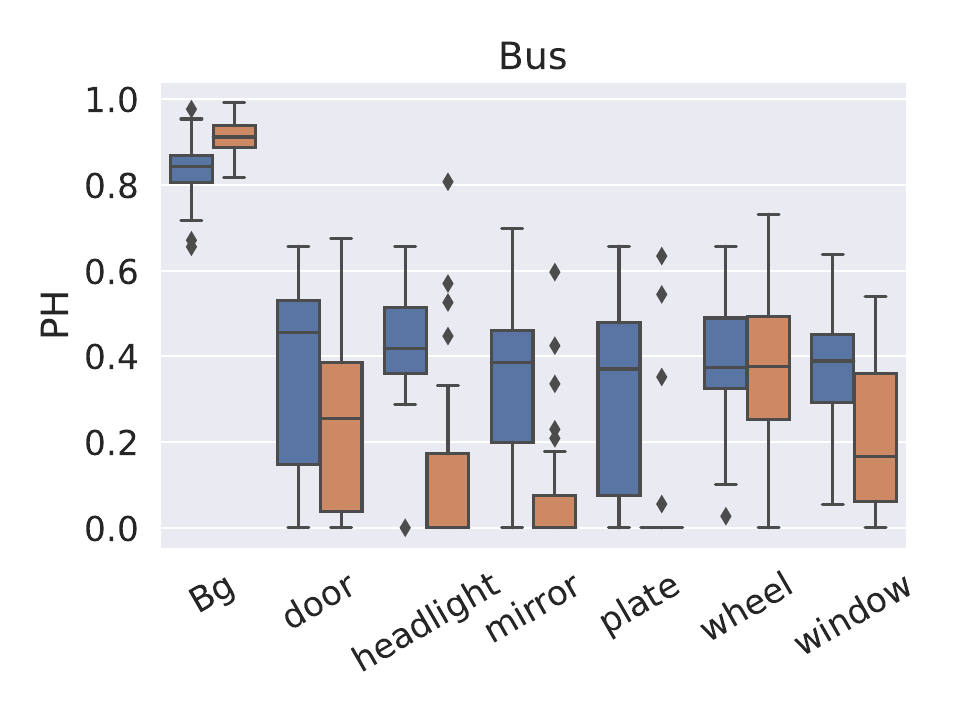}
      \end{subfigure}
    \end{minipage}
    \caption{Pascal-Part}
    \label{fig:pascalvoc_f1}
  \end{subfigure}
  \caption{Exp. 1: Representative examples of PQAH analysis for DNN models. On the X-axis, various parts are displayed, with `Bg' denoting the background. Complete PQAH analysis can be found in the supplementary materials.}
\end{figure*}

\subsection{Networks and Implementation Details}
Pre-trained networks including ResNet-50 \cite{he2016deep}, VGG-16 \cite{simonyan2014very} and Vision Transformer (ViT) \cite{dosovitskiy2020image} have been chosen as the backbone architectures for our study, as they epitomize the primary backbone architectures utilized in computer vision research. Moreover, each of these models comes equipped with publicly available pre-trained weights that have been obtained through training on the training sets of the datasets whose test/val sets are used in our evaluation.

\noindent{\bf Default Heatmap Extraction Method.} For extracting heatmaps, the default approach we employ is GradCAM \cite{selvaraju2017grad} in conjunction with SESS \cite{tursun2022sess}. GradCAM is one of the most popular heatmap visualisation techniques. Recognizing the diversity in test data input sizes and to enhance visualization quality, SESS is concurrently utilized. In the context of SESS, the default scale parameter is set to 2, and no prefiltering is applied. 

\section{Experiments and Results}
\label{sec:exp}

One of the applications of PQAH is to offer a quantitative and detailed analysis of DNNs by leveraging their heatmaps. To demonstrate the effectiveness of PQAH, we conducted comparisons that consider various aspects, including different network architectures, the same network trained with different data augmentation techniques, and the same network subjected to various saliency-enhancing approaches. The objectives of these experiments are threefold: (\textbf{i}) To assess whether PQAH aligns with established metrics like the Top-1 error rate. (\textbf{ii}) To determine if PQAH offers unique and detailed insights that are not captured by other existing metrics. (\textbf{iii}) To ascertain whether the PQAH analysis effectively measures the enhancements in generalization and attention of more advanced DNNs.

\label{sec:exp1}
\paragraph* {Exp. 1: Analyzing Networks with PQAH} ResNet-50, VGG-16 and ViT (tiny and base) are compared with PQAH over PartImageNet and PASCAL-Part datasets. Pre-trained weights\footnote{ResNet-50:https://download.pytorch.org/models/resnet50-11ad3fa6.pth, VGG-16:https://download.pytorch.org/models/vgg11-8a719046.pth, ViT:https://huggingface.co/facebook/deit-tiny-patch16-224} learned with ImageNet are used for the PartImageNet dataset. In comparison, pre-trained weights\footnote{https://github.com/ruthcfong/pointing\_game} learned on the PASCAL VOC 2007 dataset are used to compare the ResNet-50 and VGG-16 networks on PASCAL-Part dataset. Note that no publicly available ViT object classification model trained on the PASCAL VOC dataset was found, so ViT has not been included for PASCAL-Part dataset.

\begin{table}[tbh!]
    \centering
    \caption{Exp. 1: Summary Statistics of PQAH analysis of DNN models on the PartImageNet and PASCAL-VOC datasets.}
	\begin{tabular}{c|c|c|c|c}
	\hline 
	\multirow{2}{*}{\textbf{Networks}} & \textbf{Top-1} &{$\mathbf{PH_{Q1}}$ $\uparrow$} & $\mathbf{PH_{Q2}}$ $\uparrow$ & $\mathbf{PH_{Q3}}$ $\uparrow$  \\ \cline{3-5}
        & \textbf{Error $\downarrow$} & \multicolumn{3}{c}{\textbf{PartImageNet}} \\\hline
		ResNet-50 & 20.1 &0.47 & 0.61 & 0.73 \\
		VGG-16 & 28.4 & 0.26 & 0.39 & 0.56 \\
		ViT (tiny) & 27.8 & 0.34 & 0.50 & 0.67\\ %
        ViT (base) & \bf{18.2} & \bf{0.48} & \bf{0.63} & \bf{0.74}\\\hline
            & & \multicolumn{3}{c}{\textbf{Pascal-Part}} \\\hline
            ResNet-50 & - & \bf{0.37} & \bf{0.56} & \bf{0.69}\\
		VGG-16 & - & 0.15 & 0.30 & 0.48\\
		\hline
	\end{tabular}
  \label{tab:pqah-net}
\end{table}

\begin{figure}[tbh!]
	\centering
	\includegraphics[width=0.9\linewidth]{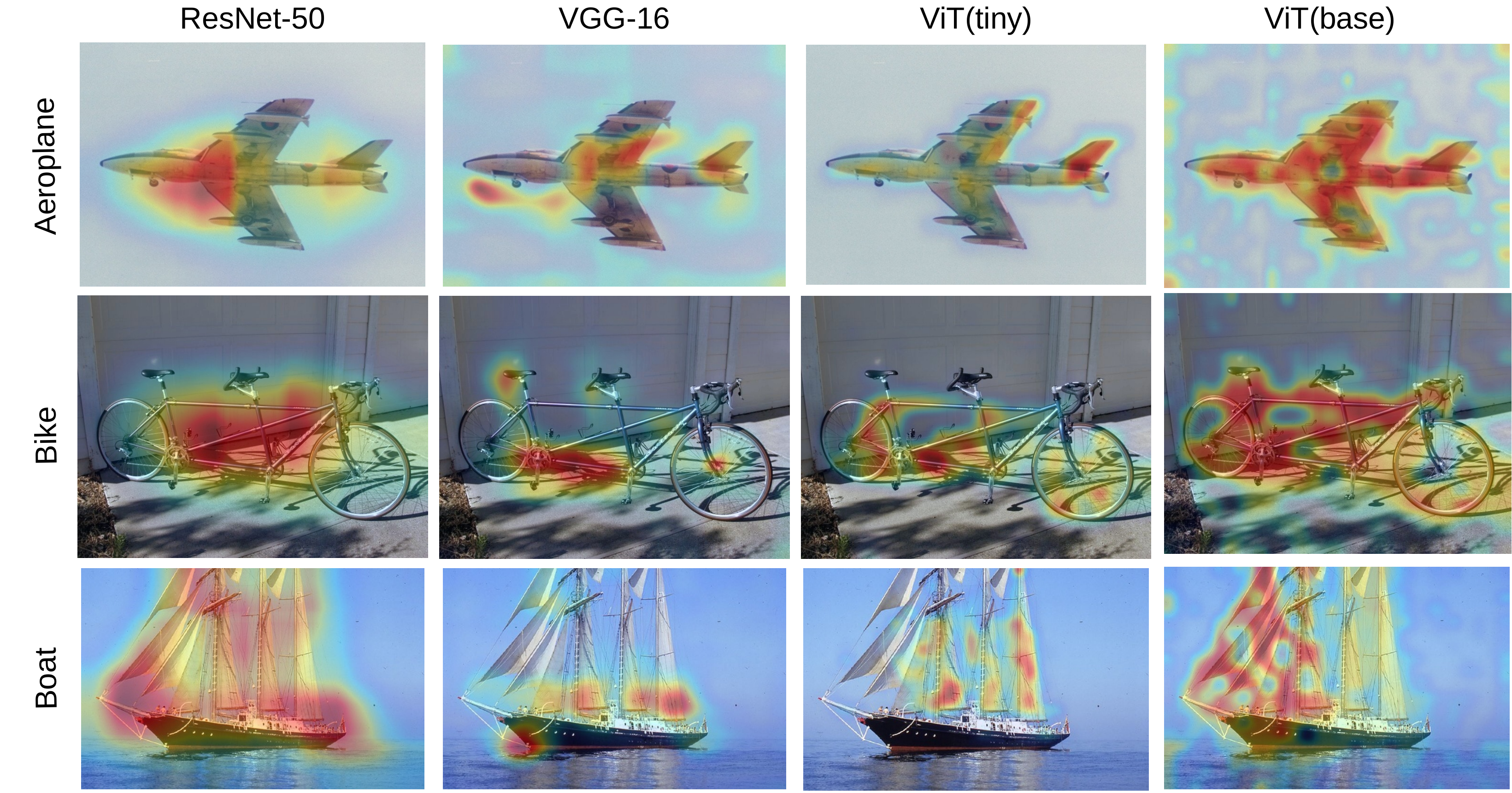}	
	\captionof{figure}{Exp. 1: Example heatmap visualizations. The visualization method is GradCam + SESS.}
	\label{fig:pin_heatmap}
\end{figure}

Following the PAQH analysis, we acquired 11 individual results corresponding to the 11 semantic classes of the PartImageNet dataset and 14 individual results for the 14 semantic classes of the PASCAL-Part dataset. To ensure clarity and accommodate space limitations, we have chosen to present only three representative results from each dataset. These can be seen in Fig. \ref{fig:partimagenet_f1} and \ref{fig:pascalvoc_f1}. Additionally, you can find example heatmap visualizations corresponding to the results displayed in Fig. \ref{fig:partimagenet_f1} in Fig. \ref{fig:pin_heatmap}. 

An overview of the PQAH analysis on the PartImageNet and PASCAL-Part datasets is provided as a summary in Table \ref{tab:pqah-net}. This table provides average values for $PH_{Q1}$, $PH_{Q2}$ and $PH_{Q3}$ in each category, as explained in Section \ref{sec:pqah2}. In this work, we emphasize that the provision of $PH_{Q1-3}$ aims to ascertain the alignment of overall PQAH outcomes with conventional metrics, as evidenced in Table \ref{tab:pqah-net}. Notably, an inverse relationship is observed, where higher $PH_{Q1-3}$ values correlate with a reduced Top-1 error rate. A secondary use is capturing overall PQAH results when individual region results are too numerous to display and consider individually.

Analysing the summary statistics, the ranking of $PH_{Q1-3}$ scores aligns with the ranking based on Top-1 error rates achieved on the ImageNet and PASCAL VOC datasets as shown in Table \ref{tab:pqah-net}, with ViT (base) achieving the lowest top-1 error, followed by ResNet-50. Additionally, both ViT (base) and ResNet-50's average median value (over all parts) exceeds 0.56, signifying that they take into account over half of the object area during classification.

Upon closer examination of the individual results in Fig. \ref{fig:partimagenet_f1} and \ref{fig:pascalvoc_f1}, we note that both ViT (base) and ViT (tiny) outperform ResNet-50 for specific objects, such as ``Aeroplane'', and for distinctive parts like ``Aeroplane Wing'', ``Bicycle Tire'', and ``Boat Sail''. Furthermore, both ViT and VGG-16 exhibit higher $\mathit{PH}$ scores in the ``background'' category compared to ResNet-50, suggesting the direct increased attention to foreground elements.

\begin{figure*}[!ht]
  \centering
  \begin{subfigure}{\textwidth}
  \centering
  \begin{minipage}{\textwidth}
      \begin{subfigure}{0.325\textwidth}
        \includegraphics[width=\linewidth]{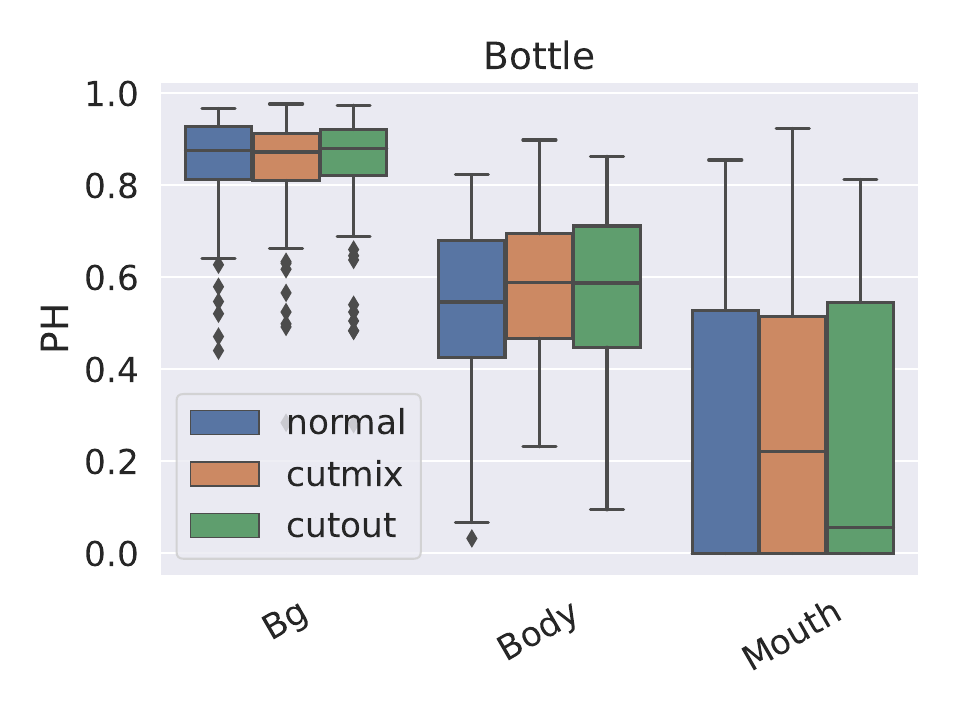}
      \end{subfigure}
      \hfill %
      \begin{subfigure}{0.325\textwidth}
        \includegraphics[width=\linewidth]{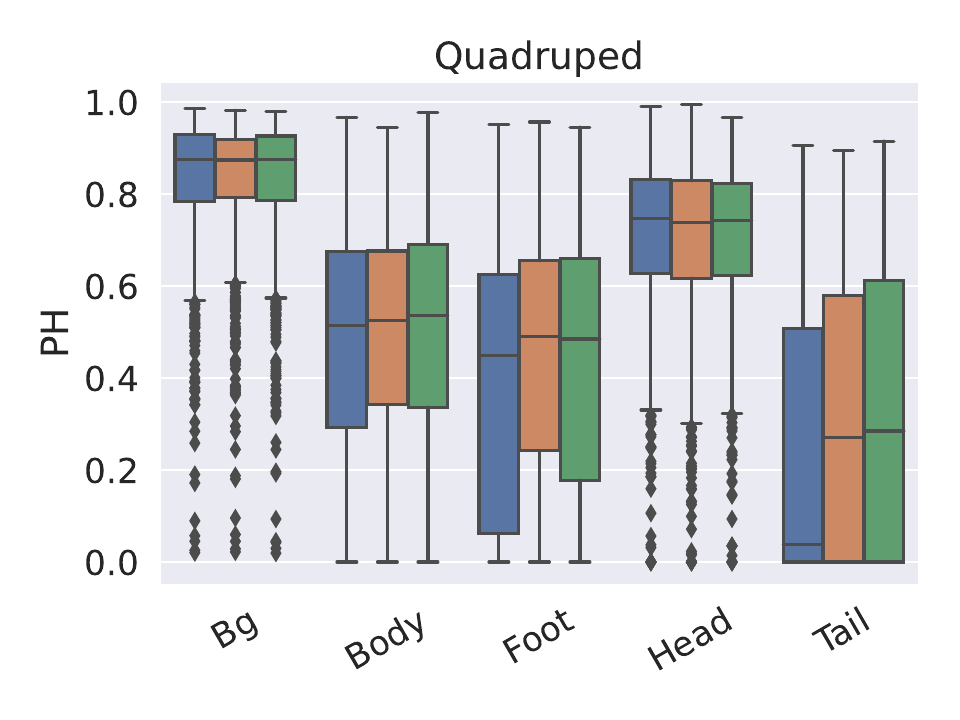}
      \end{subfigure}
      \hfill %
      \begin{subfigure}{0.325\textwidth}
        \includegraphics[width=\linewidth]{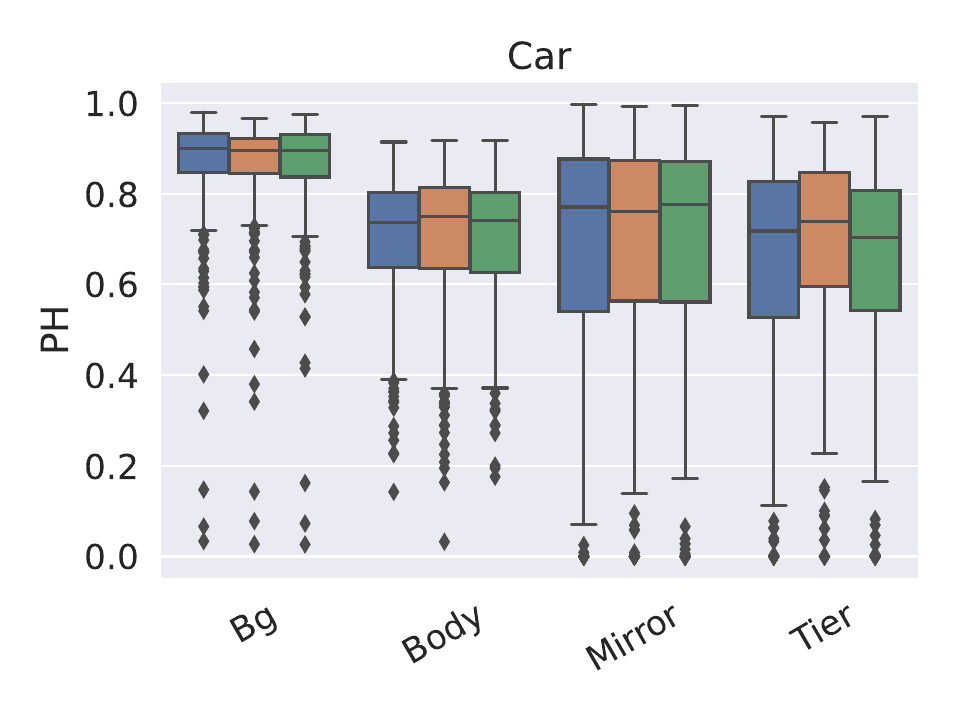}
      \end{subfigure}
  \end{minipage}
  \caption{Exp. 2: PQAH analysis of data augmentation approaches on the PartImageNet dataset.}
      \label{fig:dataaug_f1}
  \end{subfigure}

  \vspace{1em} %

\begin{subfigure}{\textwidth}
  \begin{minipage}{\textwidth}
      \begin{subfigure}{0.325\textwidth}
        \includegraphics[width=\linewidth]{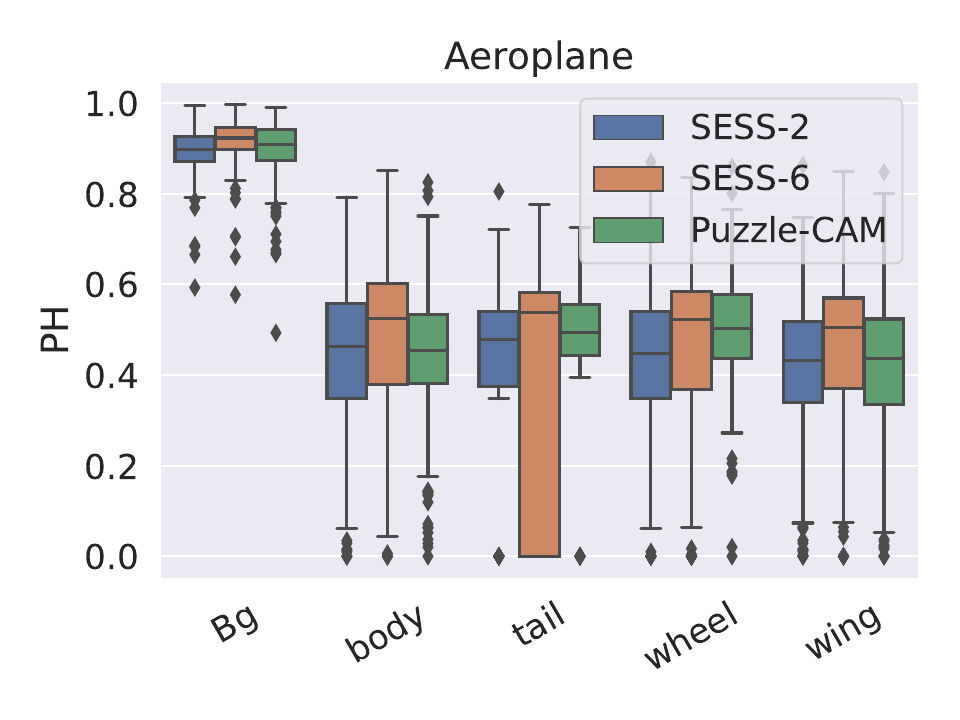}
      \end{subfigure}
      \hfill
      \begin{subfigure}{0.325\textwidth}
        \includegraphics[width=\linewidth]{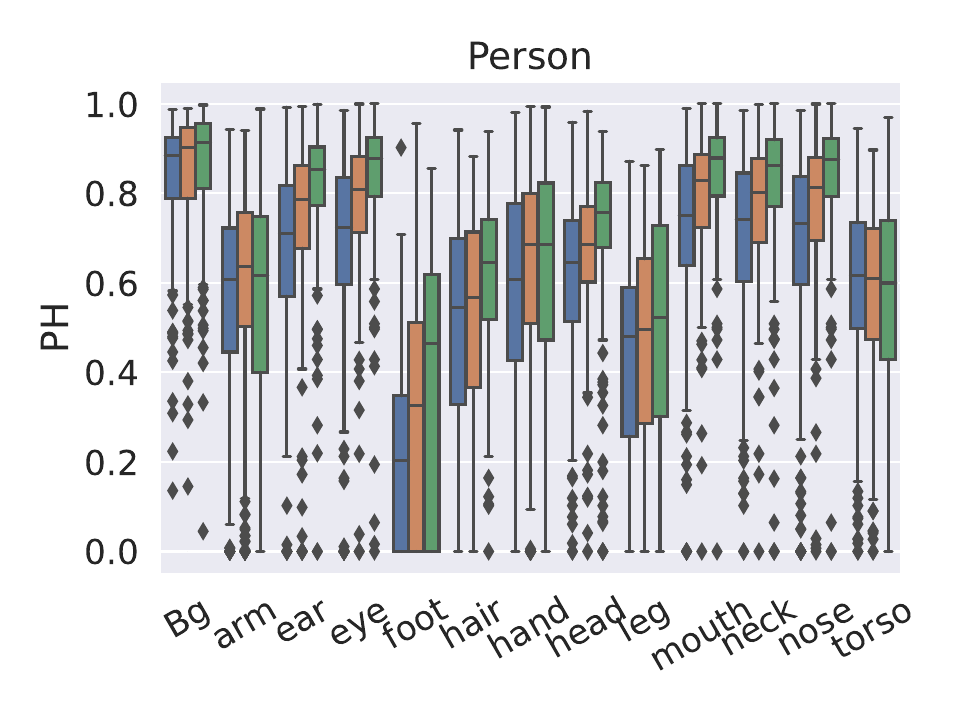}
      \end{subfigure}
      \hfill
      \begin{subfigure}{0.325\textwidth}
        \includegraphics[width=\linewidth]{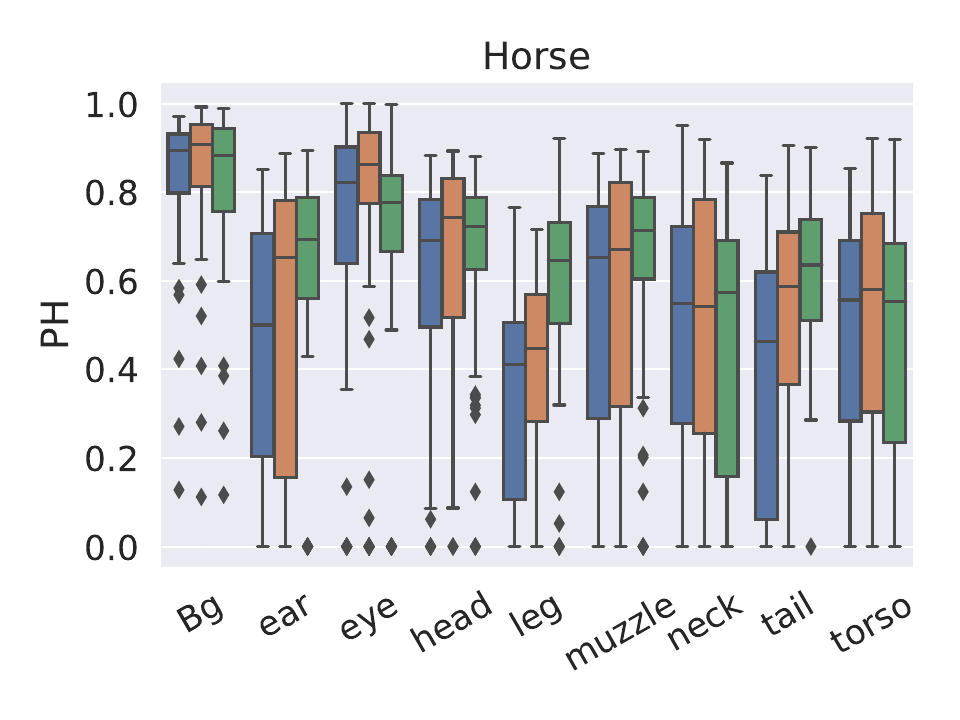}
      \end{subfigure}
  \end{minipage}
  \caption{Exp. 3: PQAH analysis of saliency enhancing approaches on the PASCAL-Part dataset.}
      \label{fig:sess-puz}
\end{subfigure}

\begin{subfigure}{\textwidth}
  \begin{minipage}{\textwidth}
      \begin{subfigure}{0.325\textwidth}
        \includegraphics[width=\linewidth]{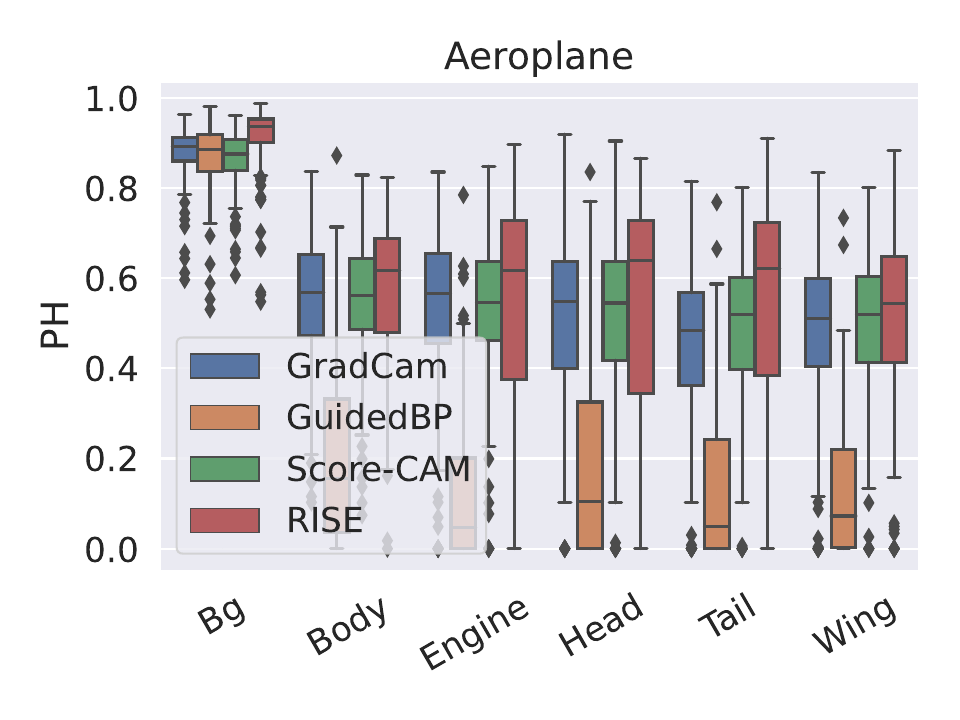}
      \end{subfigure}
      \hfill
      \begin{subfigure}{0.325\textwidth}
        \includegraphics[width=\linewidth]{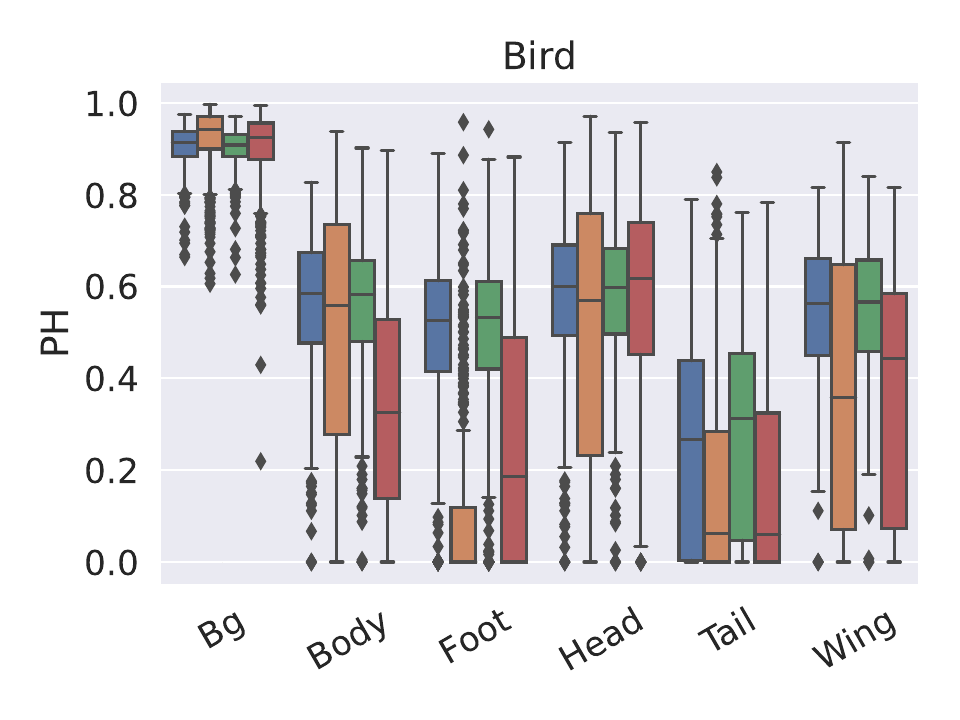}
      \end{subfigure}
      \hfill
      \begin{subfigure}{0.325\textwidth}
        \includegraphics[width=\linewidth]{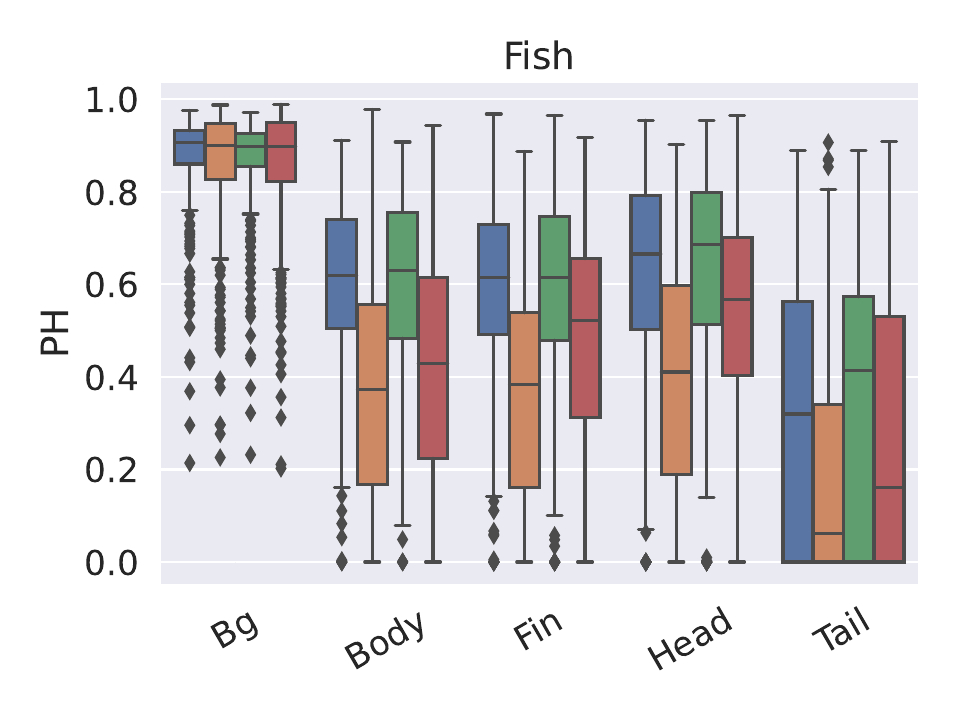}
      \end{subfigure}
  \end{minipage}
  \caption{Exp. 4: Comparison of Heatmap Extraction Methods with PQAH on PartImageNet.}
    \label{fig:PQAH-eval}
\end{subfigure}

\caption{Representative examples of PQAH analysis of Exp. 2-4. The backbone network is ResNet-50. On the X-axis, various parts are displayed, with 'Bg' denoting the background.}

\end{figure*}

\paragraph*{Exp. 2: The Influence of Data Augmentation}
Data augmentation techniques, such as Cutout \cite{devries2017improved} and CutMix \cite{yun2019cutmix}, have played a pivotal role in enhancing the generalisation capabilities of neural networks by encouraging the model to focus on distinct regions within input objects. Their effectiveness is typically measured using numerical metrics like Top-1 error rate as shown in Table \ref{tab:pqah-da}. While the Top-1 error provides an overall measure of the effectiveness of these data augmentation techniques, a more detailed comparison is often preferred, especially from an XAI perspective. To address this need, we applied the proposed methods to the aforementioned networks\footnote{Please visit https://github.com/clovaai/CutMix-PyTorch for accessing the pre-trained weights.} over the PartImageNet dataset, and the example results are presented in Fig. \ref{fig:dataaug_f1}. Based on these results, it is evident that the utilization of Cutout and CutMix techniques has led the network to enhance its focus on smaller components of the objects of interest, with better localisation of components such as ``Bottle Mouth'', ``Quadruped Tail'', and ``Car Mirror'' observed.

In conclusion, PQAH not only effectively captures subtle differences in heatmaps but also aligns with established metrics, such as the Top-1 Error Rate as given in Table \ref{tab:pqah-da}. 

\begin{table}[tbh!]
\centering
 \caption{Exp. 2: Summary Statistics of PQAH analysis of DNN models on the PartImageNet dataset.} %
	\begin{tabular}{c|c | c|c|c}
		\hline 
		\textbf{Data}& \textbf{Top-1} &\multicolumn{3}{c}{\textbf{PartImageNet}} \\\cline{3-5}
	\textbf{Augmentation}	& \textbf{Error $\downarrow$} & {$\mathbf{PH_{Q1}}$ $\uparrow$} & $\mathbf{PH_{Q2}} \uparrow$ & $\mathbf{PH_{Q3}} \uparrow$ \\
		\hline
		Normal \cite{he2016deep} & 23.68 & 0.46 & 0.60 & \bf 0.72 \\
		CutMix \cite{yun2019cutmix} & \bf 21.40 & \bf 0.47 & \bf 0.61 & \bf 0.72 \\
		Cutout \cite{devries2017improved} & 22.90 & 0.46 & 0.61 & \bf 0.72 \\ %
		\hline
	\end{tabular}
  \label{tab:pqah-da}
\end{table}

\paragraph*{Exp. 3: Impact of Saliency Enhancing Methods}
Unlike data augmentation techniques, studies such as Puzzle-CAM \cite{jo2021puzzle} and SESS explicitly compel models to focus on distinct regions either via re-training or a combination of pre- and post-processing. In these studies, the efficacy of the proposed techniques is assessed by their performance on downstream tasks like object localization and segmentation. However, a detailed comparison is often lacking in these studies. Here, we utilise PQAH for comparing Puzzle-CAM and SESS at two different scales (2 and 6). Here, ResNet-50 is the backbone and GradCAM is the heatmap visualisation method. Note that for Puzzle-CAM, ResNet-50 is initialised with the weights learned with Puzzle-CAM. Summary statistics are given in Table \ref{tab:pqah-puz}, and the selected comparative results are visualized in Fig. \ref{fig:sess-puz}. In general, Puzzle-CAM demonstrates superior performance when compared to SESS with a scale of 2. However, it's noteworthy that with a scale setting of 6, SESS exhibits better $\mathit{PH}$ scores than Puzzle-CAM on individual categories, namely ``aeroplane'' and ``horse''. 

Overall, the PQAH analysis results establish the effectiveness of Puzzle-CAM in enhancing heatmaps, while demonstrating that a larger number of scales significantly benefits SESS's performance. This conclusion is consistent with the findings of the Puzzle-CAM \cite{jo2021puzzle}  and SESS \cite{tursun2022sess}.

\paragraph*{Exp. 4: Evaluating Heatmaps with PQAH}

\begin{figure}[!ht]
    \centering
    \includegraphics[width=0.9\linewidth]{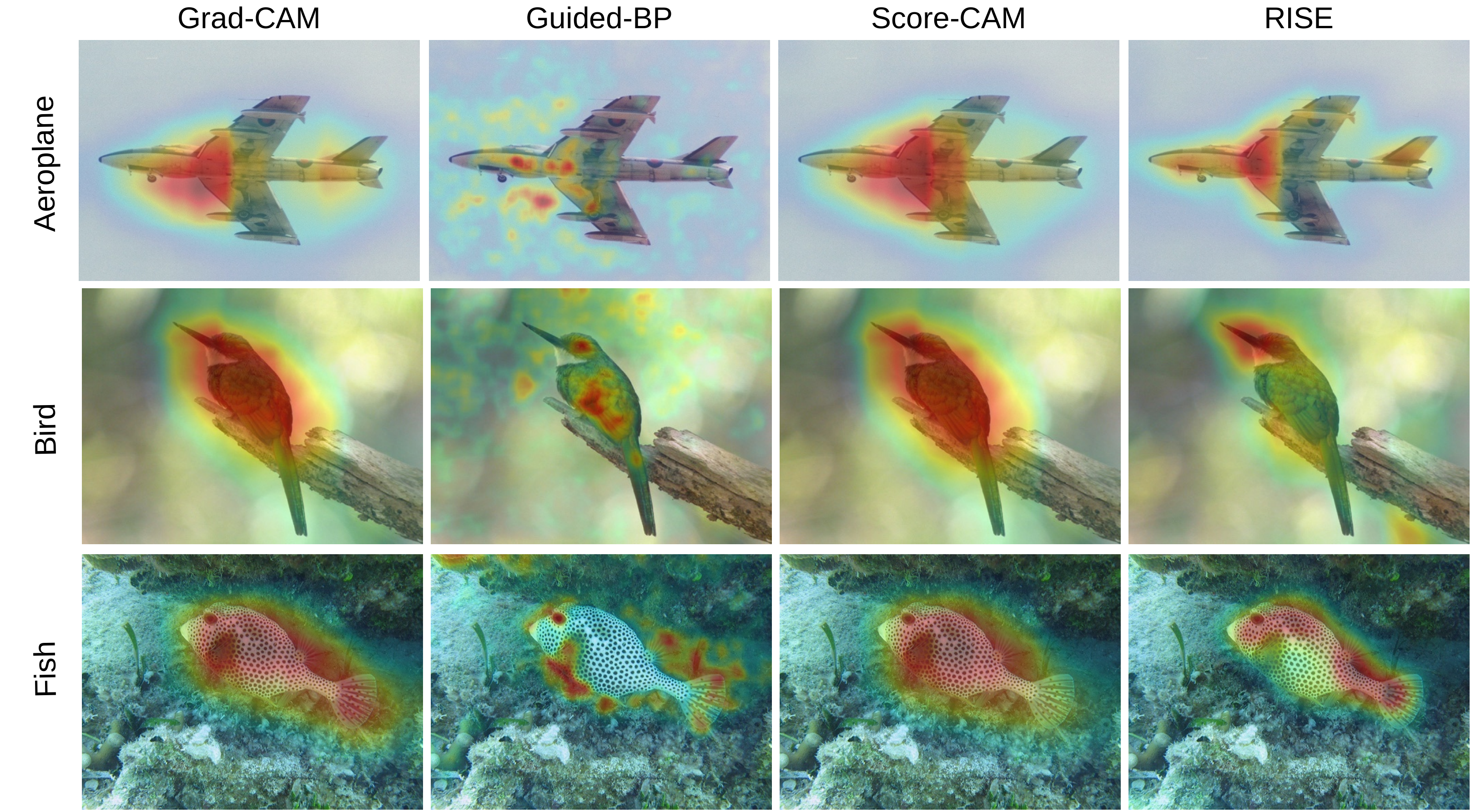}    
    \caption{Exp. 4: Example heatmap visualization results from various heatmap generation approaches.}
    \label{fig:eval_heatmap}
\end{figure}

We compare various heatmap extraction methods with PQAH to demonstrate PQAH's application in heatmap evaluation. We have chosen four well-established heatmap visualization techniques, namely Grad-CAM, Guided Backpropagation \cite{springenberg2014striving}, Score-CAM \cite{wang2020score}, and RISE \cite{petsiuk2018rise}, for this comparative study. An analysis of the summary statistics presented in Table \ref{tab:pqah-eval} reveals that Score-CAM followed by Grad-CAM achieves the highest $\mathit{PH}$ scores, while Guided Backpropagation attains the lowest $\mathit{PH}$ scores. Nonetheless, the detailed results depicted in Fig. \ref{fig:PQAH-eval} indicate that RISE and Guided Backpropagation outperform the other methods for specific parts of certain classes. For example, for the ``Aeroplane'' class, RISE returns the highest $\mathit{PH}$ scores. Example heatmap visualizations corresponding to the results in Fig. \ref{fig:PQAH-eval} are presented in Fig. \ref{fig:eval_heatmap}.

\begin{table}[!t]
	\centering
 \captionof{table}{Exp. 3: Summary Statistics of PQAH analysis of Puzzle-CAM and SESS on the PASCAL-VOC dataset. The backbone network is ResNet-50. }
 {
	\begin{tabular}{c|c|c|c}
			\hline 
		\textbf{Methods} &{$\mathbf{PH_{Q1}} \uparrow$} & $\mathbf{PH_{Q2}} \uparrow$ & $\mathbf{PH_{Q3}} \uparrow$ \\
			\hline
			SESS-2 \cite{tursun2022sess} & 0.37 & 0.56 & 0.69 \\
			SESS-6 \cite{tursun2022sess} & 0.40 & 0.59 & \textbf{0.72} \\
			Puzzle-CAM \cite{jo2021puzzle} & \bf{0.44} & \bf{0.60} & \bf{0.72} \\ %
			\hline
		\end{tabular}
 }
  \label{tab:pqah-puz}

\end{table}

\begin{table}
\centering
 \captionof{table}{Exp. 4: Comparison of Heatmap Extraction Methods with PQAH on PartImageNet. The backbone network is ResNet-50.} %
	\begin{tabular}{c|c|c|c}
		\hline 
		\textbf{Methods}
		& {$\mathbf{PH_{Q1}} \uparrow$} & $\mathbf{PH_{Q2}} \uparrow$ & $\mathbf{PH_{Q3}} \uparrow$ \\
		\hline
		Grad-CAM \cite{selvaraju2017grad} & 0.47 & 0.61 & 0.73 \\
		Guided-BP \cite{springenberg2014striving} & 0.23 & 0.37 & 0.53 \\
		Score-CAM \cite{wang2020score}& \textbf{0.48} & \textbf{0.63} & \textbf{0.74} \\ 
		RISE \cite{petsiuk2018rise} & 0.35 & 0.52 & 0.67 \\ 
		\hline
	\end{tabular}
  \label{tab:pqah-eval}

\end{table}

\paragraph*{Exp. 5: Run Time}
Excluding the time taken for heatmap generation and part-annotation, the PAQH system demonstrates a remarkably short runtime. It processes each image in approximately 0.015 seconds on a desktop machine with an Intel® Core™ i7-6700 CPU @ 3.40GHz.

\paragraph*{Exp. 6: Generating XAI reports}
\label{exp:xai}
As outlined in Section \ref{sec:report}, PQAH results can be utilised for generating end-user-friendly text-based XAI reports through the integration of state-of-the-art large language models, particularly GPT-4. We have compiled such reports for ResNet-50, VGG-16, and ViT (Tiny), utilizing the data derived from Exp. 1. Complete XAI reports will be included in the supplementary material. An example of a briefly summarized XAI report can be found in Table \ref{tab:xai_report}. A qualitative assessment of these reports reveals their insightful nature and structured format, contributing significantly to enhancing the network's performance. These enhancements are evident in terms of both accuracy and consistency, spanning various categories and parts. Furthermore, the reports strike a balance between technical rigor and clarity, rendering them beneficial for diverse audiences, ranging from expert technical teams to stakeholders with limited AI expertise.

\begin{table}[h]
\centering
\caption{A summary of the example GPT-4-generated XAI report for the ResNet-50 model, utilizing PQAH data on the PartImageNet dataset (Exp. 1).}
\begin{tabular}{l|l}
\hline
\textbf{Content} & \textbf{Details} \\ \hline
Advantages & \begin{tabular}[c]{@{}l@{}}1. High Background Detection Accuracy\\ 2. Strong in Certain Parts and Categories\end{tabular} \\ \hline
Disadvantages & \begin{tabular}[c]{@{}l@{}}1. Inconsistent Part Detection\\ 2. Challenges in Specific Categories\end{tabular} \\ \hline
Suggestions & \begin{tabular}[c]{@{}l@{}}1. Focused Training on Weak Areas\\ 2. Model Architecture Optimization\end{tabular} \\ \hline
\end{tabular}
\label{tab:xai_report}
\end{table}

\section{Limitation of PQAH Analysis}
PQAH is designed to be a human-perspective-based metric, as it measures the alignment between heatmaps and semantic part segments annotated by humans. While this characteristic could be seen as a limitation if we take the gap between human and machine perspective \cite{linsley2017visual,ullman2016atoms}. As a result, higher PQAH scores do not automatically imply superior model efficacy for specific tasks, and conversely, lower scores may not indicate inadequate model performance. Nevertheless, the human-centric nature of PQAH positions it as a valuable tool for Explainable AI (XAI) focused on bridging the understanding between AI outputs and human interpretability.

While PQAH metrics are computed through an automated process, the interpretation of these metrics necessitates human insight. As the number of classes and semantic components within the dataset increases, the analytical process becomes more complex. This limitation, however, could be mitigated with the assistance of a large language model, as demonstrated in Exp. 6 and the supplementary material.

\section{Conclusion}
\label{sec:con}

In this study, we introduced PQAH, a novel framework designed to conduct granular quantitative analysis on the heatmaps of DNNs. Through a series of comprehensive experiments, we have successfully demonstrated the efficacy of PQAH in providing objective, scalable, and granular analyses. Its utility extends to creating end-user-friendly XAI and offering nuanced, fine-grained evaluations of heatmaps. This dual capability highlights PQAH's potential as a significant tool for both XAI enhancement and in-depth heatmap analysis in the realm of deep learning.

\appendix

\section{Usecase: Using PQAH for Improving DNN Performance}
\label{app1}
The PQAH method is valuable not only for analyzing deep learning models but also for enhancing them. We specifically apply PQAH to the medical field, where we see significant potential, though we highlight that PQAH is designed for broad applicability across diverse computer vision domains.

The medical cases we have chosen involve COVID-19 classification based on Chest X-ray images. In particular, we would like to improve the deep learning model trained to classify if a patient has COVID based on Chest X-ray images. 

\subsection{Dataset, Network and Experiment Setup}
\label{app2} 
\begin{figure}[!t]
	\centering
	\includegraphics[width=0.9\linewidth]{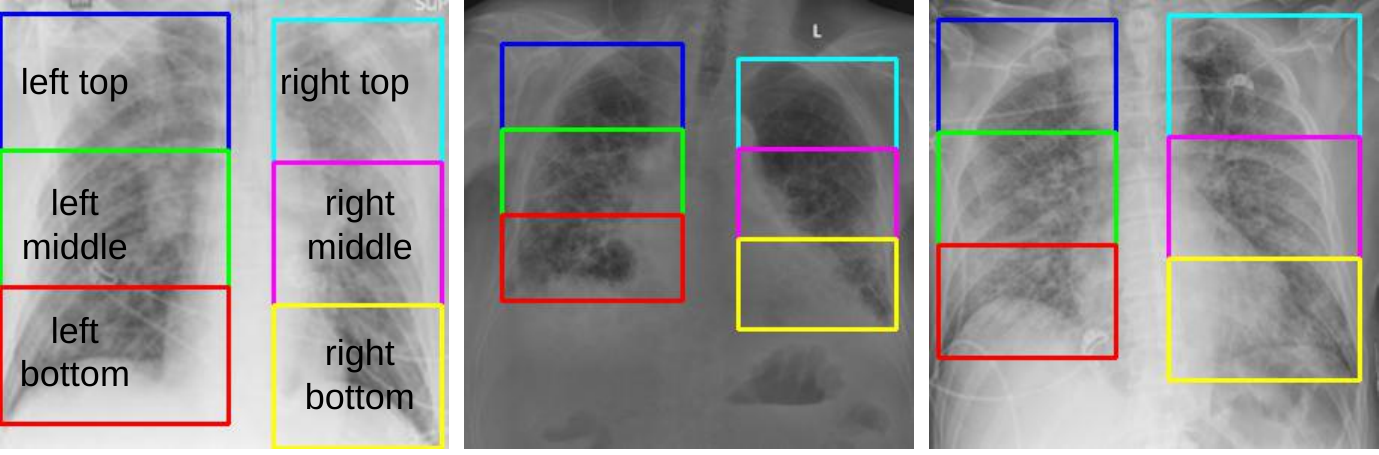}
	\caption{Examples of the six region-based segments of the lung.}
	\label{fig:clpart}
\end{figure}

For our study, we have selected the COVID-19 Chest X-ray dataset\footnote{https://www.kaggle.com/datasets/tawsifurrahman/covid19-radiography-database/data}. This dataset is divided into three categories: normal (699 cases), COVID-19 (2,015 cases), and Viral Pneumonia (275 cases). While segmentation masks for the entire lung are provided, no fine-grained part segmentation data is included. We syntactically divided the lung into six parts based on positions. We utilise a contour selection algorithm to locate the left and right lung bounding box based on the binary lung mask, and subsequently divide each lung into three parts vertically: top, middle and bottom. In total, each X-ray has six lung parts: left top (lt), left middle (lm), left bottom (lb), right top (rt), right middle (rm) and right bottom (rb). Fig. \ref{fig:clpart} demonstrates how a lung X-ray is divided into six parts. 

The COVIDNext50 model{\footnote{https://github.com/velebit-ai/COVID-Next-Pytorch}} that builds upon the ResNext50 \cite{xie2017aggregated} architecture with the training setup from the original repository is used.

\subsection{Using PQAH to Improve Network Performance}

\begin{figure}[!tbh]
	\centering
	\includegraphics[width=0.99\linewidth]{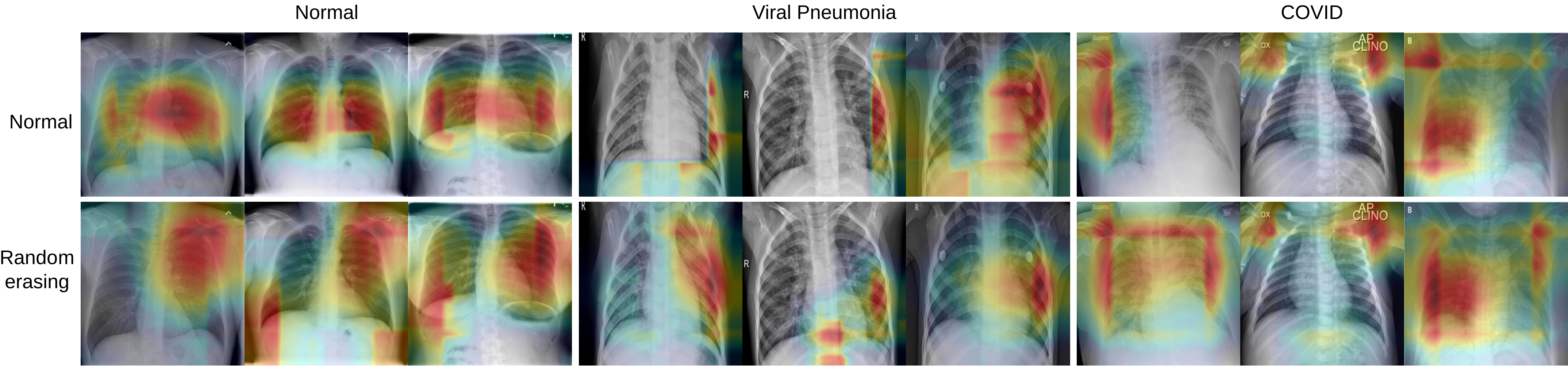}
	\caption{Examples of heatmaps extracted from the CovidNext50 model before (top row) and after (bottom row) PQAH analysis and data augmentation.}
	\label{fig:covid_heatmap}
\end{figure}

\begin{figure}[!tbh]
	\centering
	\includegraphics[width=0.99\linewidth]{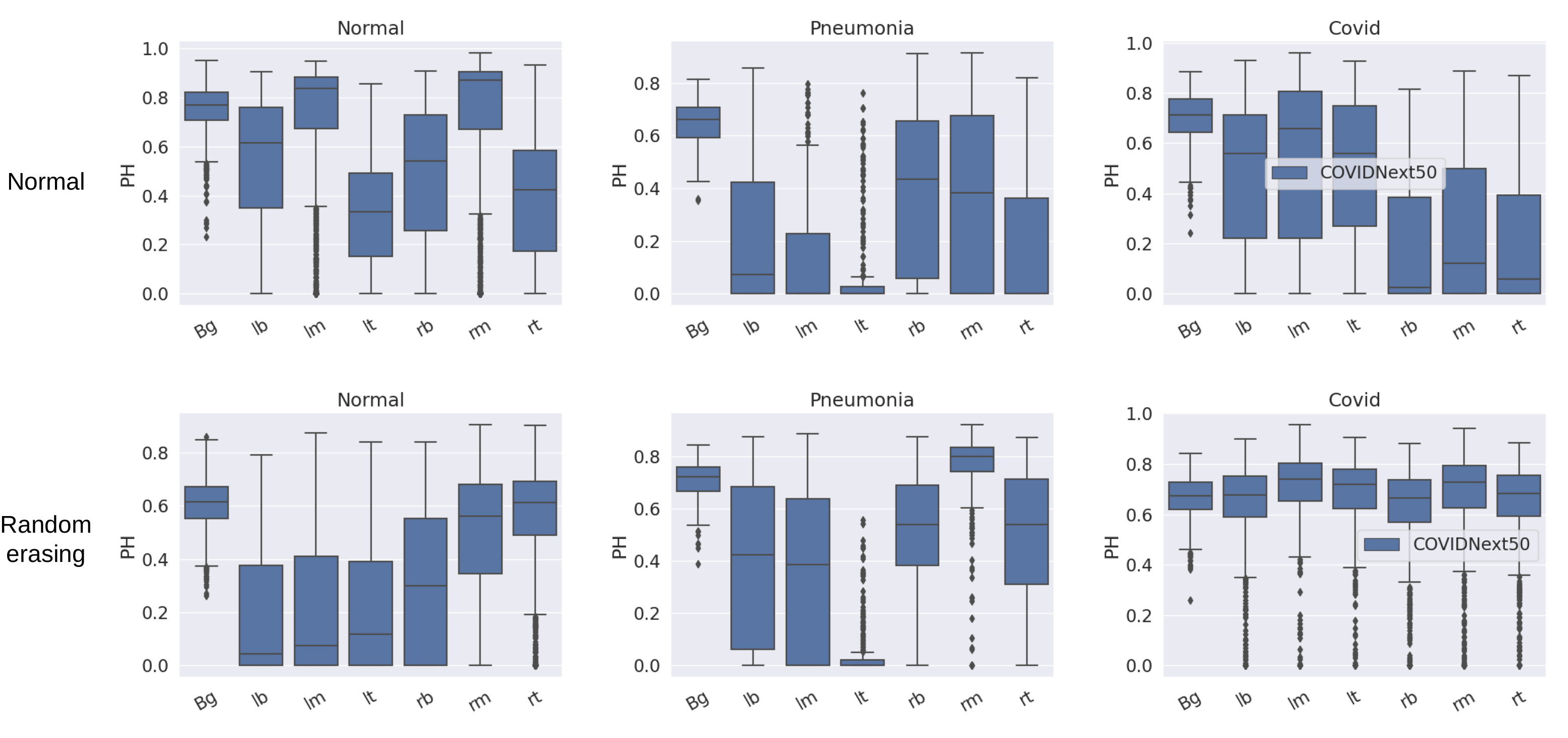}
	\caption{PQAH analysis for COVIDNext50 models on the COVID Chest x-ray dataset. The top row shows boxplots for the original COVIDNext50 model, while the bottom row shows the COVIDNext50 model trained with the random erasing data augmentation that was applied based on deficiencies identified from the initial PQAH analysis.}
	\label{fig:pqahbbox}
\end{figure}

After conducting a PQAH analysis on the original model, we identified a position-based bias based on the PQAH results depicted in Fig. \ref{fig:pqahbbox}. Specifically, the analysis shows a predominant focus on the left lung for COVID images, while for Viral Pneumonia images, the right lung is more frequently targeted. To mitigate this bias, we incorporated Random Erasing data augmentation \cite{zhong2020random} into the COVIDNext50 dataset. This adjustment reduced the position-based bias. For example, as shown in Fig. \ref{fig:pqahbbox}, the model better considers the both left and right lungs when classifying a case into the COVID category. Additionally, the model's accuracy is increased from $94.97\%$ to $96.03\%$.To visually demonstrate the improvement, we provide examples of heatmaps before and after the PQAH analysis and model retraining in Fig. \ref{fig:covid_heatmap}.

\subsection{Summary}

Through this practical example, we have shown that PQAH analysis can be effective in identifying overfitted and underfitted regions within models. This insight is helpful for developing improved training strategies that significantly enhance model performance.

\section{Generating Critiques from PQAH with a Large Language Model}
In Section 3.3 and Experiment 6 of the main manuscript, we discuss the generation of end-user-friendly Explainable Artificial Intelligence (XAI) reports using a large language model like GPT-4. This process involves sending PQAH (Part-based Question Answering for Humans) results (as a JSON file) along with a specific prompt, as illustrated in Fig. \ref{tab:prompt}. We created XAI reports for three neural networks: ResNet-50, VGG-16, and Vision Transformer (ViT) Tiny, utilizing PQAH data derived from the PartImageNet dataset as detailed in Fig. 1. These generated reports are given in Figures \ref{tab:report_resnet50}, \ref{tab:report_vgg16} and \ref{tab:report_vit}.

We have analysed these generated XAI reports and observed the following: 
\begin{itemize}
	\item{\bf Readability and Language}: The reports are user-friendly and easy to read.
	\item{\bf Coverage}: The reports effectively highlight the main advantages and disadvantages of the networks and offer technical suggestions with references.
	\item{\bf Depth of Analysis}: The analysis of the network's strengths and weaknesses lacks depth. This could be due to the absence of a standard reference for comparison.
	\item{\bf General Nature of Suggestions}: The technical suggestions provided are perceived as too general and not specifically tailored to the unique aspects of the analyzed network.
\end{itemize}

These findings suggest that the goal of automatically generating end-user-friendly and technically robust XAI reports is attainable. However, to further improve the quality of the report,  we need to take the following points into consideration:
\begin{itemize}
	\item Developing an LLM which is specialized in analyzing neural networks. The GPT-4 model, employed for generating reports, is a general large language model and is not specialized in analyzing neural networks. 
	\item Additional information including specific details like the network architecture and training logs alongside the PQAH results could be provided to enable richer analysis results.
	\item Providing a standard reference for comparison may help the LLM generate more meaningful suggestions.
\end{itemize}

\begin{table}[!tbh]
	\centering
	\begin{tabular}{|p{8cm}|}
		\hline
		
		Act as an AI expert specialized in creating user-friendly reports, and generate a brief report summarizing the main advantages and disadvantages of a network, and offer technical suggestions on how to improve it.  This report will be based on a part-based quantitative analysis of heatmaps (PQAH) data presented in JSON format. The PQAH data specifically pertains to part-based heatmap analysis, with these heatmaps being derived from a Deep Neural Network (DNN).\newline
		
		The provided data is structured as follows:\newline
		
		- Top-level keys in the JSON file represent different categories.
		- Sub-level keys within each category represent individual parts.
		- For each part in each category, we have access to three key performance metrics: Q1, Median, and Q3 F1 scores.
		- `Bg' represents background\newline
		
		It's important to note that the F1 score for each part is calculated based on a comparison between the heatmap generated by the network and the ground truth part annotation.  High F1 scores indicate a strong overlap between the heatmap and the ground-truth part annotation. Heatmap highlights the regions of an input image responsible for a network's classification result.\newline
		
		Now, based on the PQAH data provided in the following chat, proceed to analyze the main pros and cons of the network and provide technical suggestions with references (high-rank conferences and journal papers) to improve the network.\\
		\hline	
	\end{tabular}
	\captionof{figure}{The prompt for generating an XAI report based on PQAH data.}
	\label{tab:prompt}
\end{table}

\begin{table*}[!t]
	\begin{tabular}{|p{18cm}|}
		\hline
		
		\textbf{Main Advantages}:\newline
		\textit{High Background Discrimination}: The network shows strong performance in distinguishing background (Bg) across most categories, as indicated by high F1 scores. This suggests effective background-foreground segmentation.\newline
		
		\textit{Consistent Performance in Certain Parts}: Some parts, like `Car Body', `Reptile Head', and `Biped Head', have relatively high F1 scores, indicating reliable detection and segmentation in these areas.\newline
		
		\textbf{Main Disadvantages}:\newline
		\textit{Inconsistent Performance Across Parts}: Certain parts like `Bicycle Seat', `Biped Tail', and `Fish Tail' have low F1 scores, particularly in their Q1 metrics. This indicates inconsistent and potentially unreliable detection in these areas.\newline
		
		\textit{Underperformance in Complex Structures}: Parts with complex structures or less distinct features (e.g., `Bicycle Head', `Aeroplane Tail') show lower performance, suggesting difficulties in capturing intricate details.\newline
		
		\textbf{Technical Suggestions for Improvement}:\newline
		\textit{Enhanced Part-Based Modeling}: For parts with low F1 scores, consider integrating specialized sub-networks focusing on these complex structures. Refer to ``Part-based R-CNNs for Fine-grained Category Detection" by Ning Zhang et al., presented at ECCV 2014.\newline
		
		\textit{Data Augmentation for Underperforming Parts}: Increase the diversity and quantity of training data for underperforming parts. Techniques such as synthetic data generation can be beneficial. See ``Data Augmentation for Object Detection via Differentiable Neural Rendering" by Nikita Dvornik et al., presented at NeurIPS 2020.\newline
		
		\textit{Improved Feature Extraction for Complex Parts}: Implement advanced feature extraction techniques, such as attention mechanisms, to better capture details in complex parts. ``CBAM: Convolutional Block Attention Module" by Sanghyun Woo et al., published in ECCV 2018, can be a useful reference.\newline
		
		\textit{Regularization Techniques}: To address overfitting on certain parts, employ regularization techniques like dropout or batch normalization. Refer to ``Batch Normalization: Accelerating Deep Network Training by Reducing Internal Covariate Shift" by Sergey Ioffe and Christian Szegedy, presented at ICML 2015.\newline
		
		\textit{Ensemble Learning}: Combining predictions from multiple models can improve overall accuracy, especially for parts with lower F1 scores. The paper ``Ensemble Deep Learning: A Review" by M.A. Ganaie et al., published in Arxiv 2021, provides insights into this approach.\newline
		
		\textbf{Conclusion:}
		The network demonstrates strong background discrimination but struggles with complex part structures and certain specific parts. Addressing these issues through enhanced modeling, data augmentation, improved feature extraction, regularization, and ensemble learning can lead to significant improvements in network performance. The suggested papers provide a solid foundation for these improvements.\\
		\hline
	\end{tabular}
	\captionof{figure}{The XAI report generated for ResNet-50 Using PQAH data extracted from the PartImageNet dataset.}
	\label{tab:report_resnet50}
\end{table*}

\begin{table*}
	\begin{tabular}{|p{18cm}|}
		\hline

		\textbf{Advantages}:
		
		\textit{High Background Detection}: Across all categories, the network shows a consistently high F1 score for background detection (`Bg'). This indicates strong performance in distinguishing the main objects from the background, which is crucial for accurate part-based analysis. \newline
		
		\textit{Good Performance on Certain Parts}: Some parts, like the `Bird Head' and `Snake Head', show high median F1 scores, indicating that the network is effective in identifying these specific parts. \newline
		
		\textbf{Disadvantages}:
		
		\textit{Poor Performance on Small or Complex Parts}: Many parts such as `Aeroplane Tail', `Bicycle Seat', `Biped Tail', `Bird Tail', and `Boat Body' have low F1 scores, especially in their first quartile (Q1). This suggests difficulty in detecting smaller or more complex parts. \newline
		
		\textit{ Inconsistency Across Categories}: There is a significant variation in performance across different categories and parts. For example, `Bird' parts have higher F1 scores compared to `Boat' parts, indicating inconsistent learning across different object types. \newline
		
		\textbf{Technical Suggestions for Improvement}:
		
		\textit{Enhanced Feature Extraction}: Implement advanced convolutional neural network architectures that focus on feature extraction. Techniques like Inception modules (Szegedy et al., ``Going Deeper with Convolutions", CVPR 2015) or Residual Networks (He et al., "Deep Residual Learning for Image Recognition", CVPR 2016) can help in extracting richer features for complex parts. \newline
		
		\textit{Attention Mechanisms}: Integrate attention mechanisms (Vaswani et al., ``Attention Is All You Need", NeurIPS 2017) to focus on small and complex parts. This can guide the network to pay more attention to parts that are harder to detect. \newline
		
		\textit{Data Augmentation}: Use extensive data augmentation (Shorten and Khoshgoftaar, ``A survey on Image Data Augmentation for Deep Learning", Journal of Big Data 2019) to improve the network's ability to generalize across various categories. Techniques like rotation, scaling, and cropping can help the network learn to identify parts in different orientations and scales. \newline
		
		\textit{Region-Based Convolutional Neural Networks (R-CNNs)}: Employ R-CNNs (Girshick et al., ``Rich feature hierarchies for accurate object detection and semantic segmentation", CVPR 2014) or its variants like Fast R-CNN and Faster R-CNN for better part detection. These models are particularly effective in localizing and classifying object parts. \newline
		
		\textit{Multi-Scale Analysis}: Implement multi-scale analysis techniques (Lin et al., ``Feature Pyramid Networks for Object Detection", CVPR 2017) to better capture both small and large parts. \newline
		
		\textit{Fine-Tuning and Hyperparameter Optimization}: Regularly fine-tune the network and optimize hyperparameters based on performance metrics. This includes adjusting learning rates, batch sizes, and other network parameters to improve detection accuracy. \newline
		
		By implementing these suggestions, there should be a notable improvement in the network's ability to accurately detect and analyze various parts in different categories, leading to a more robust and efficient part-based heatmap analysis. \\
		\hline
	\end{tabular}
	\captionof{figure}{The XAI report generated for VGG-16 using PQAH data extracted from the PartImageNet dataset.}
	\label{tab:report_vgg16}
\end{table*}

\begin{table*}
	\begin{tabular}{|p{18cm}|}
		\hline
		
		Overall Performance Evaluation: A general assessment of the network's accuracy based on the F1 scores.
		Category-Specific Analysis: A deeper look into the performance for each category.
		Suggestions for Improvement: Based on the analysis, I'll provide technical suggestions with references to high-rank conferences and journal papers.\newline
		
		\textbf{Overall Performance Evaluation}\newline
		The F1 score is a measure of a test's accuracy and considers both the precision and the recall of the test. A higher F1 score indicates better performance. The scores are presented in terms of Q1 (lower quartile), Median, and Q3 (upper quartile). In general, a median F1 score closer to 1 suggests excellent performance, while lower scores indicate areas needing improvement.\newline
		\textbf{Category-Specific Analysis}\newline
		\textbf{Aeroplane}: Good performance in background detection, moderate in identifying aeroplane parts. The body and wing have relatively better scores than the head, tail, and engine.\newline
		\textbf{Bicycle}: Excellent in background, but struggles with specific parts like the seat and head.\newline
		\textbf{Biped}: Decent background detection, varied performance across parts, with the hand performing the best.\newline
		\textbf{Bird}: Strong background detection, moderate performance on body and wing, but poor in identifying the tail and foot.\newline
		\textbf{Boat}: Excellent in background, very poor in identifying the boat body, moderate in sail detection.\newline
		\textbf{Bottle}: Good background detection, moderate performance for the bottle body, very poor in mouth detection.\newline
		\textbf{Car}: Good background detection, moderate to good in identifying car parts, with the side mirror and tier having a wide range of scores.\newline
		\textbf{Fish}: Excellent background detection, good performance on the fish body, moderate on other parts.\newline
		\textbf{Quadruped}: Good background, weak performance on the body and foot, moderate on the head.\newline
		\textbf{Reptile}: Good background, moderate to good performance on body and head, poor on tail and foot.\newline
		\textbf{Snake}: Excellent background, moderate performance on the snake body, poor on the head.\newline
		
		\textbf{Suggestions for Improvement}\newline
		Based on the analysis, the following suggestions are proposed:
		
		\textit{Enhanced Part Segmentation}: The network could be improved by focusing on more accurate part segmentation, especially for categories where part detection is weak. This can be achieved by using more advanced segmentation algorithms like U-Net or DeepLab.\newline
		
		\textit{Data Augmentation}: For parts with low F1 scores, data augmentation techniques can help. This involves creating modified versions of the existing dataset to improve the network's robustness.\newline
		
		\textit{Fine-Tuning with Additional Annotations}: Incorporating additional annotated data for the poorly performing parts could help the network learn these parts better.\newline
		
		\textit{Attention Mechanisms}: Implementing attention mechanisms like the ones used in Transformer models might help the network focus better on relevant parts of the image.\newline
		
		\textit{Transfer Learning}: Leveraging pre-trained models on similar tasks could also improve the performance, especially for categories with lower scores.\newline
		
		\textbf{References}
		
		Ronneberger, O., Fischer, P., \& Brox, T. (2015). U-Net: Convolutional Networks for Biomedical Image Segmentation. MICCAI.\newline
		Chen, L. C., Papandreou, G., Kokkinos, I., Murphy, K., \& Yuille, A. L. (2018). DeepLab: Semantic Image Segmentation with Deep Convolutional Nets, Atrous Convolution, and Fully Connected CRFs. IEEE Transactions on Pattern Analysis and Machine Intelligence.\newline
		Vaswani, A., et al. (2017). Attention is All You Need. NIPS.\newline
		
		These references are from high-ranking conferences and journals in the field of AI and machine learning, providing a solid foundation for the suggested improvements.\\
		\hline
	\end{tabular}
	\captionof{figure}{The XAI report generated for ViT (tiny) using PQAH data extracted from the PartImageNet dataset.}
	\label{tab:report_vit}
\end{table*}

\section{PQAH Algorithm}
\label{app3}
Algorithm \ref{alg:PQAH} outlines the PQAH algorithm detailed in Section 3.1 of the main paper. For each image $I$, $PH$ scores are computed for both its parts and background. To calculate the $PH$ score, we utilize the $F1$ score, chosen for its intuitive interpretation, particularly when precise $Precision$ values for parts are challenging to determine. The algorithm first computes the overall $\mathit{Precision}$, which serves as an approximate precision for each part. Subsequently, the $\mathit{Recall}$ value is computed for each part. Leveraging the $\mathit{Recall}$ and $\mathit{Precision}$ values, the $F_1$ score is derived for each part. A similar process is applied to the background, with the distinction that the $\mathit{Precision}$ for the background can be directly calculated without approximation. This is because the $\mathit{PH}$ score for the background is computed at the object level.

\renewcommand{\algorithmicrequire}{\textbf{Input:}}
\renewcommand{\algorithmicensure}{\textbf{Output:}}

\begin{algorithm}[!tbh]
	\caption{PQAH}
	\label{alg:PQAH}
	\begin{algorithmic}[1]
		\Require
		$I$ - Image;
		$H$ - Heatmap;
		$M$ - Mask;
		$M^p$ - Mask of the part $p$;
		$\theta$ - Threshold;
		$Bg$ - Background;
		$TP$ - True positive;
		$FN$ - False negative;
		$FP$ - False positive;
		\Ensure
		$PH$ - Dictionary to store the $PH$ scores  of each part $p$ of image $I$;		
		\State $H \gets H > \theta$\\
		\textcolor{gray}{\# Calculate F1 score for each part}
		\State $\mathit{TP} \gets \sum (M \odot H)$
		\State $\mathit{FP} \gets \sum H - \mathit{TP}$
		\State $\widetilde{\mathit{Precision}} \gets \frac{\mathit{TP}}{\mathit{TP} + \mathit{FP}}$ \textcolor{gray}{\qquad \# Equation 3}
		\For{$p$ in $parts(I)$} 
		\State $\mathit{TP} \gets \sum (M^p \odot H) $
		\State $\mathit{FN} \gets \sum M^p - TP$
		\State $\mathit{Recall} \gets \frac{\mathit{TP}}{\mathit{TP} + \mathit{FN}}$ \textcolor{gray}{\qquad \#Equation 2}
		\State $PH^p \gets \frac{2 \cdot \widetilde{\mathit{Precision}} \cdot \mathit{Recall}}{{\widetilde{\mathit{Precision}}} + \mathit{Recall}}$ \textcolor{gray}{\qquad \# Equation 1}
		\EndFor
		\\
		\textcolor{gray}{\# Calculate F1 score for the background of the image $I$}
		\State $\mathit{TP} \gets \sum ((1-M) \odot (1- H))$
		\State $\mathit{FP} \gets \sum (1-H) - \mathit{TP}$
		\State $\mathit{FN} \gets \sum (1-M) - \mathit{TP}$
		\State $\mathit{Precision} \gets \frac{\mathit{TP}}{\mathit{TP} + \mathit{FP}}$
		\State $\mathit{Recall} \gets \frac{\mathit{TP}}{\mathit{TP} + \mathit{FN}}$
		\State $PH^{Bg} \gets \frac{2 \cdot \mathit{Precision} \cdot \mathit{Recall}}{{\mathit{Precision}} + \mathit{Recall}}$
		
	\end{algorithmic}
\end{algorithm}

\section{Complete PQAH Results}
\begin{figure*}[hbt!]
	\centering
	\begin{subfigure}{0.325\textwidth}
		\centering
		\includegraphics[width=\linewidth]{images/F1/partimagenet/Aeroplane.pdf}
	\end{subfigure}
	\begin{subfigure}{0.325\textwidth}
		\centering
		\includegraphics[width=\linewidth]{images/F1/partimagenet/Bicycle.pdf}
	\end{subfigure}
	\begin{subfigure}{0.325\textwidth}
		\centering
		\includegraphics[width=\linewidth]{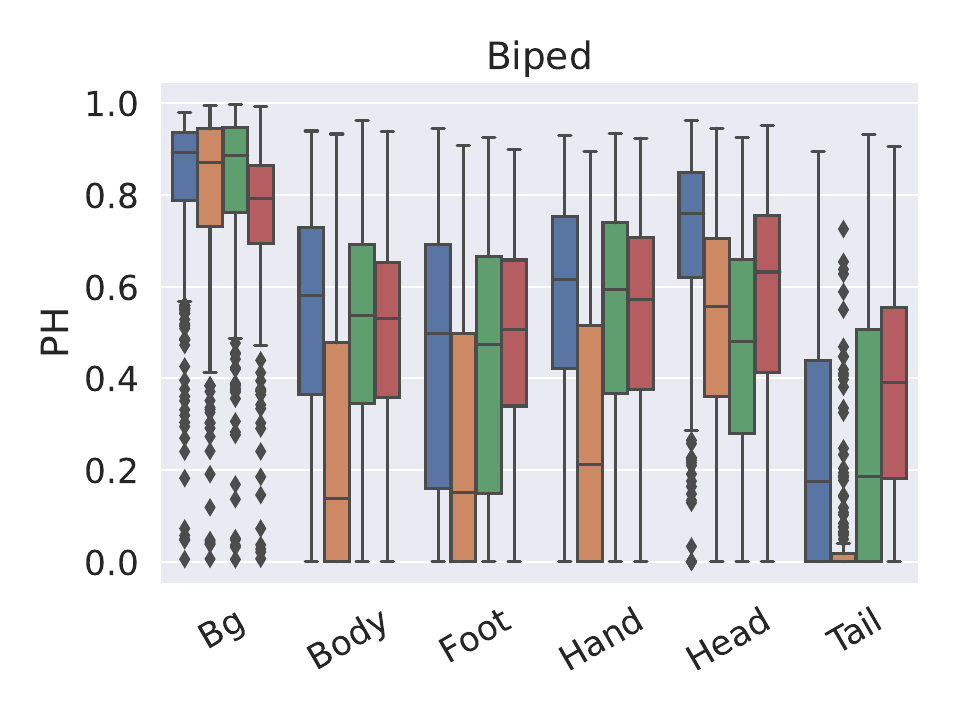}
	\end{subfigure}
	\\
	\begin{subfigure}{0.325\textwidth}
		\centering
		\includegraphics[width=\linewidth]{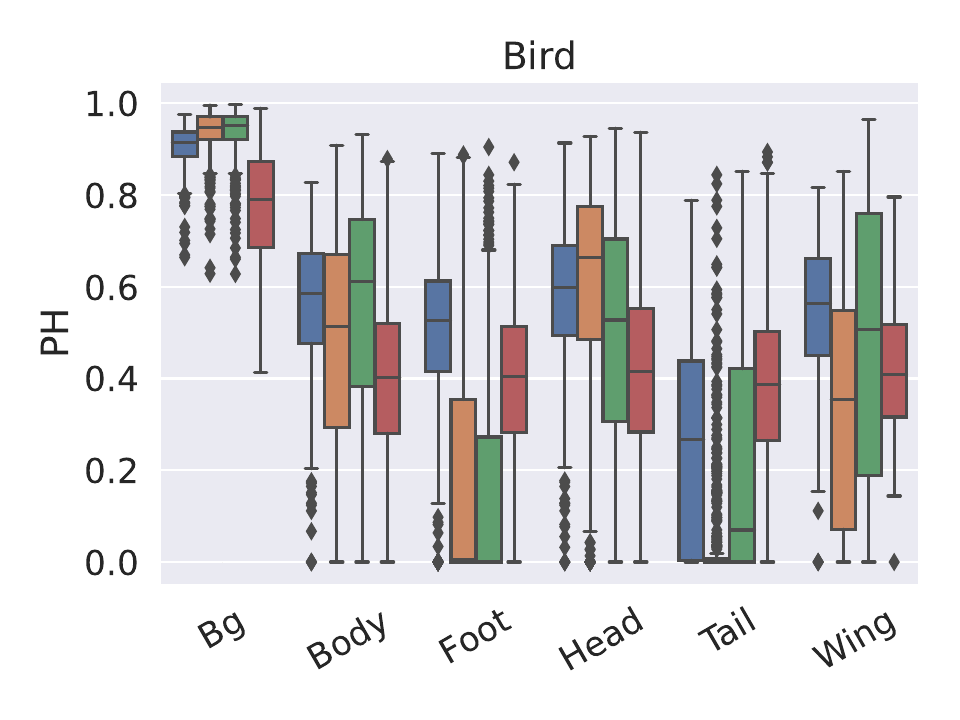}
	\end{subfigure}
	\begin{subfigure}{0.325\textwidth}
		\centering
		\includegraphics[width=\linewidth]{images/F1/partimagenet/Boat.pdf}
	\end{subfigure}
	\begin{subfigure}{0.325\textwidth}
		\centering
		\includegraphics[width=\linewidth]{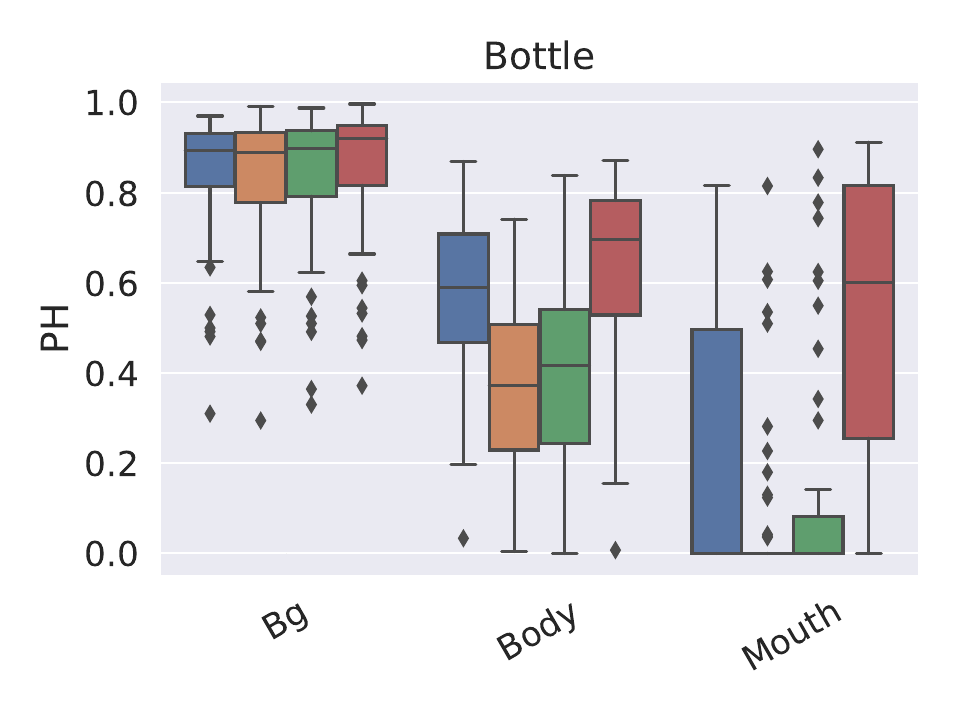}
	\end{subfigure}
	\begin{subfigure}{0.325\textwidth}
		\centering
		\includegraphics[width=\linewidth]{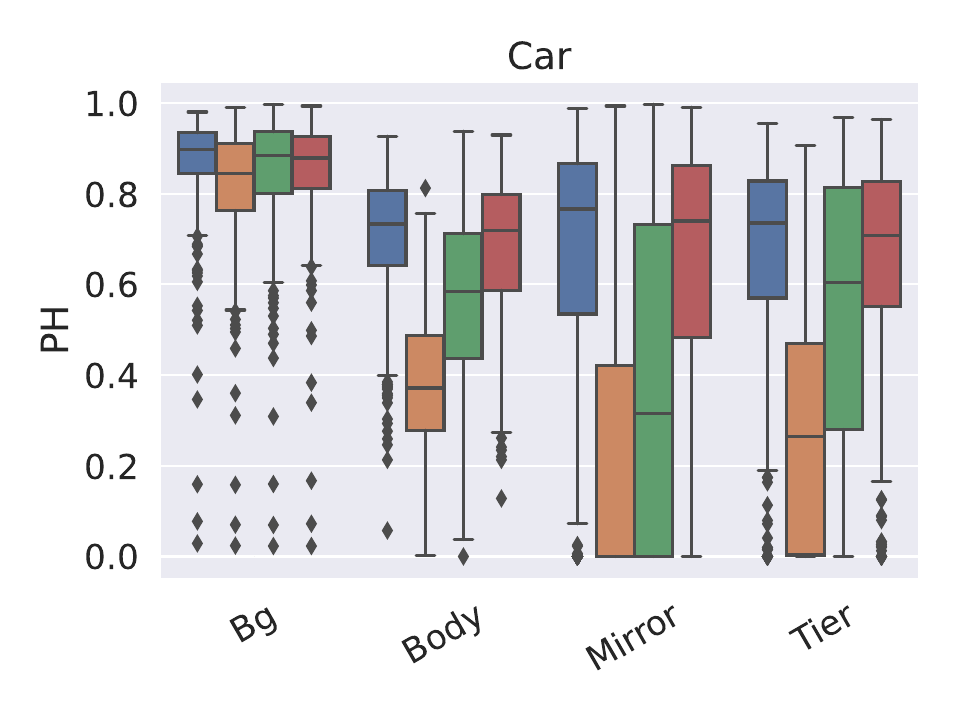}
	\end{subfigure}
	\begin{subfigure}{0.325\textwidth}
		\centering
		\includegraphics[width=\linewidth]{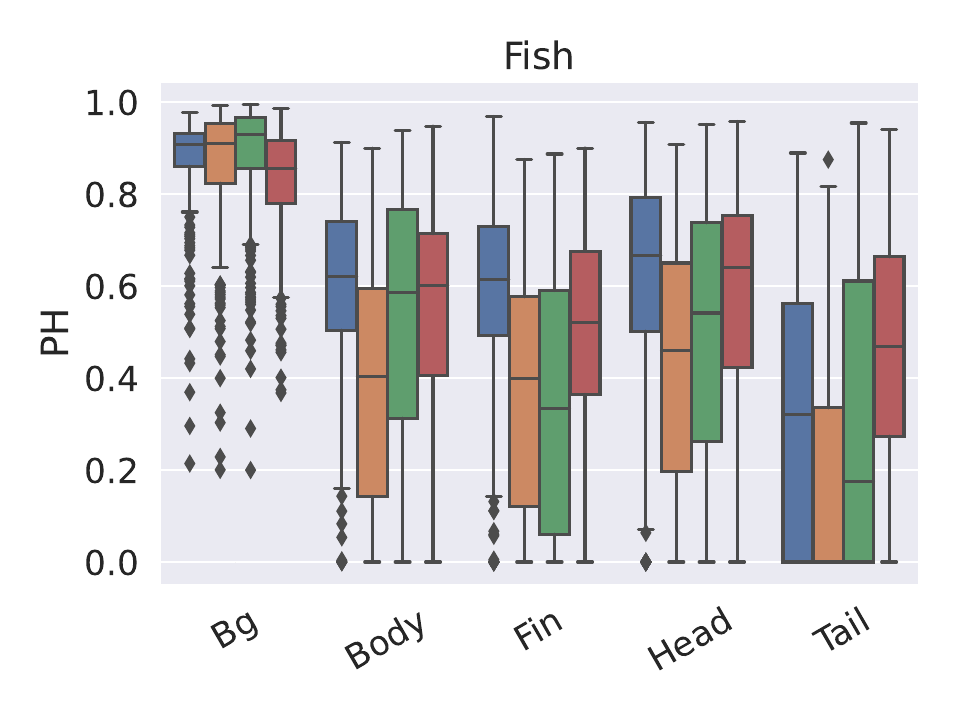}
	\end{subfigure}
	\begin{subfigure}{0.325\textwidth}
		\centering
		\includegraphics[width=\linewidth]{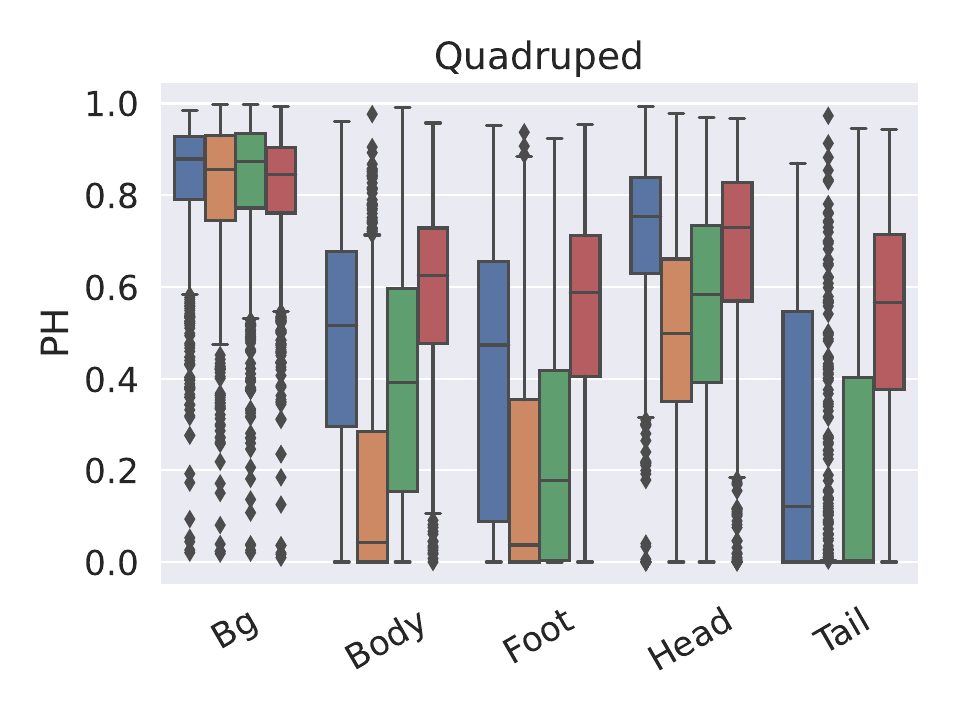}
	\end{subfigure}
	\\
	\begin{subfigure}{0.325\textwidth}
		\centering
		\includegraphics[width=\linewidth]{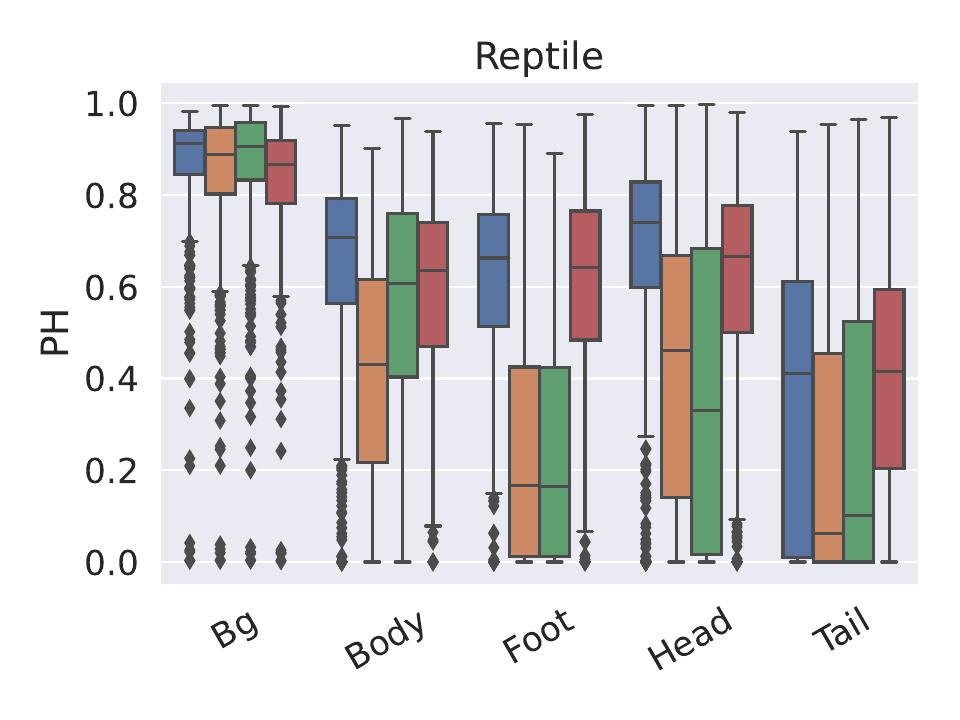}
	\end{subfigure}
	\begin{subfigure}{0.325\textwidth}
		\centering
		\includegraphics[width=\linewidth]{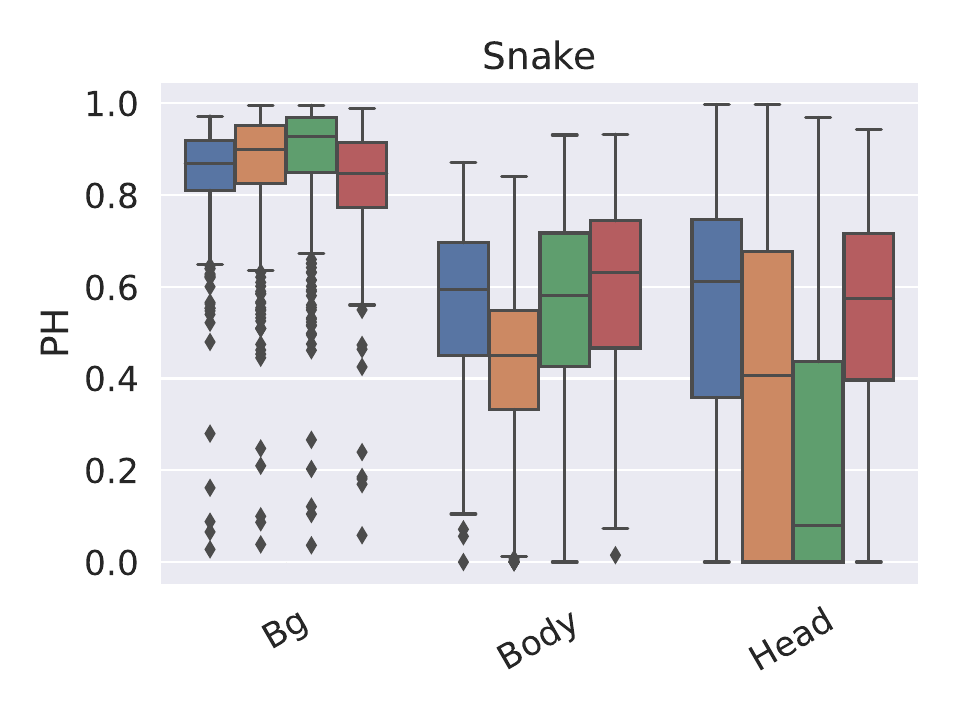}
	\end{subfigure}
	\caption{PQAH analysis for DNN models on the PartImageNet dataset (Exp. 1). On the X-axis, various parts are displayed, with 'Bg' denoting the background.}
	\label{fig:partimagenet_f1_fl}
\end{figure*}

\begin{figure*}[hbt!]
	\centering
	\begin{subfigure}{0.325\textwidth}
		\centering
		\includegraphics[width=\linewidth]{images/F1/pascalvoc/aeroplane.pdf}
	\end{subfigure}
	\begin{subfigure}{0.325\textwidth}
		\centering
		\includegraphics[width=\linewidth]{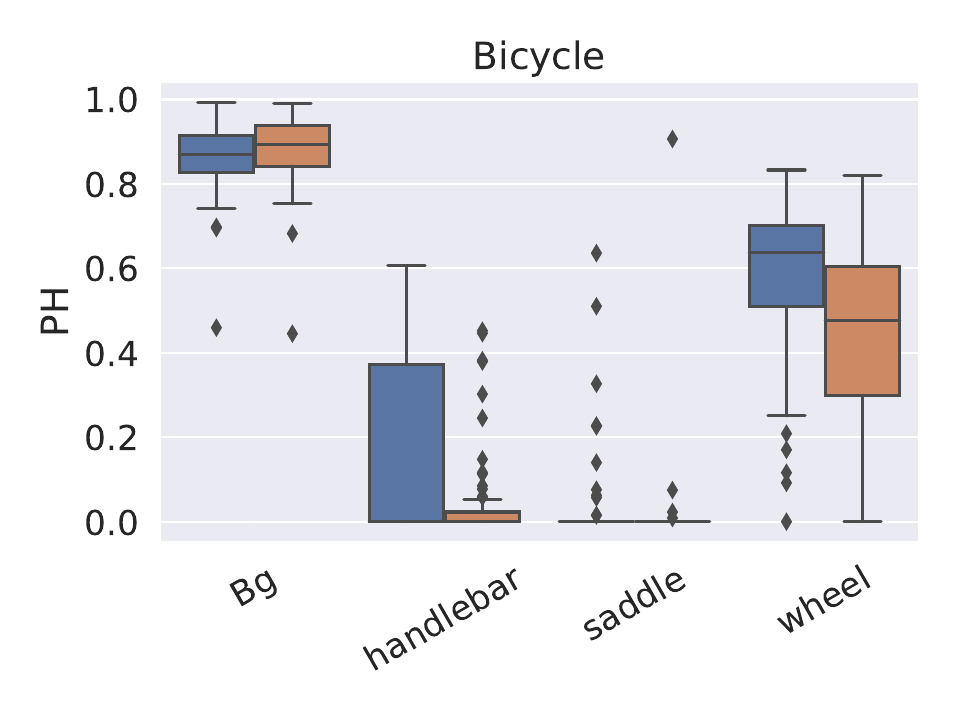}
	\end{subfigure}
	\begin{subfigure}{0.325\textwidth}
		\centering
		\includegraphics[width=\linewidth]{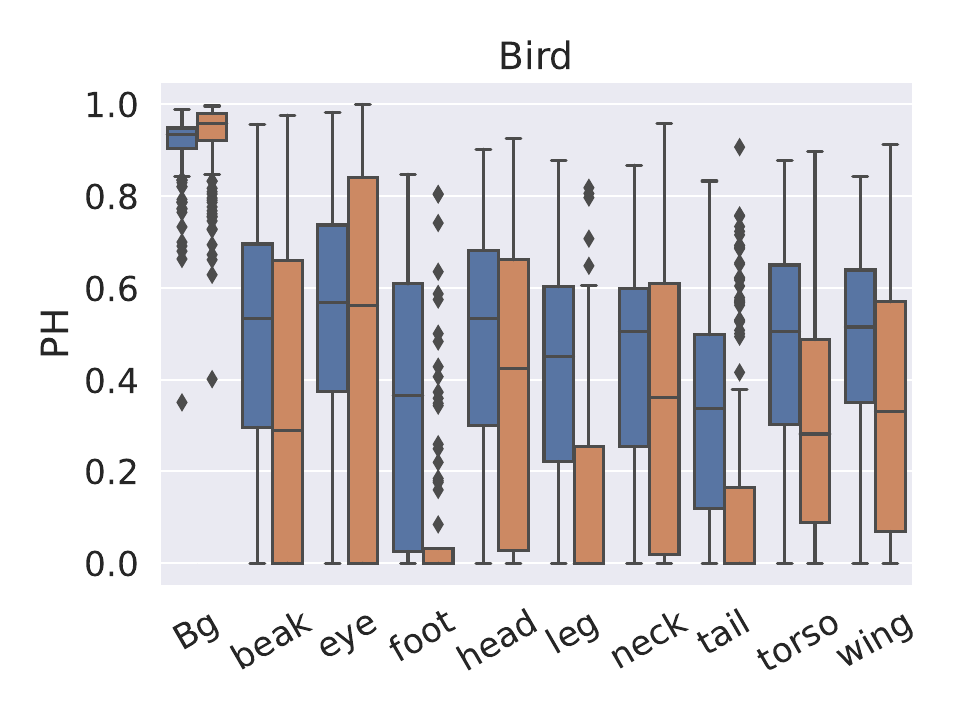}
	\end{subfigure}
	\\
	\begin{subfigure}{0.325\textwidth}
		\centering
		\includegraphics[width=\linewidth]{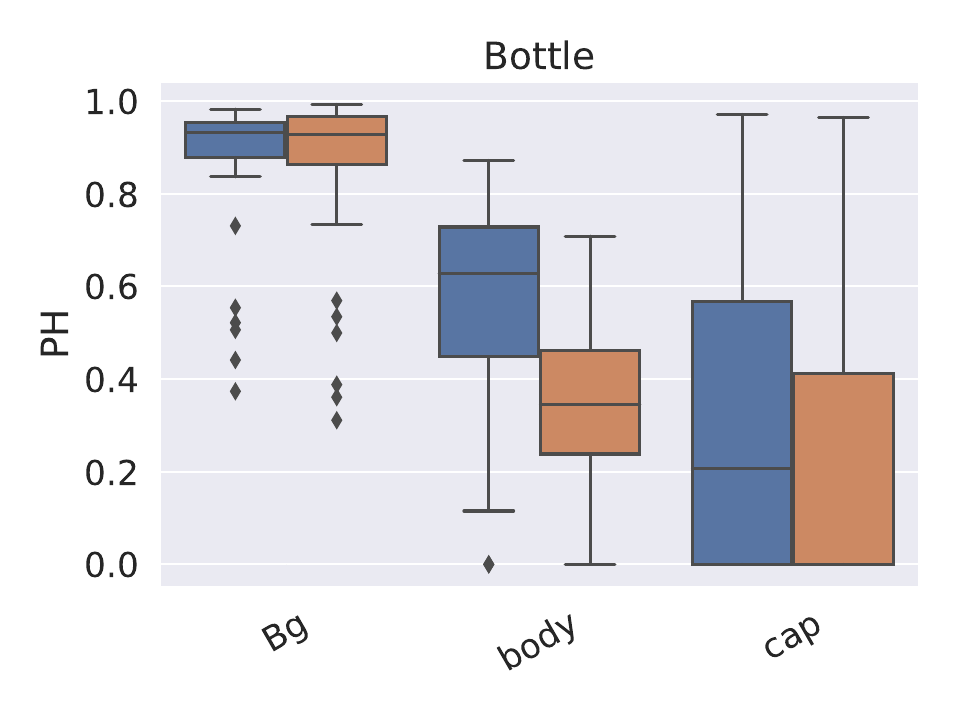}
	\end{subfigure}
	\begin{subfigure}{0.325\textwidth}
		\centering
		\includegraphics[width=\linewidth]{images/F1/pascalvoc/bus.pdf}
	\end{subfigure}
	\begin{subfigure}{0.325\textwidth}
		\centering
		\includegraphics[width=\linewidth]{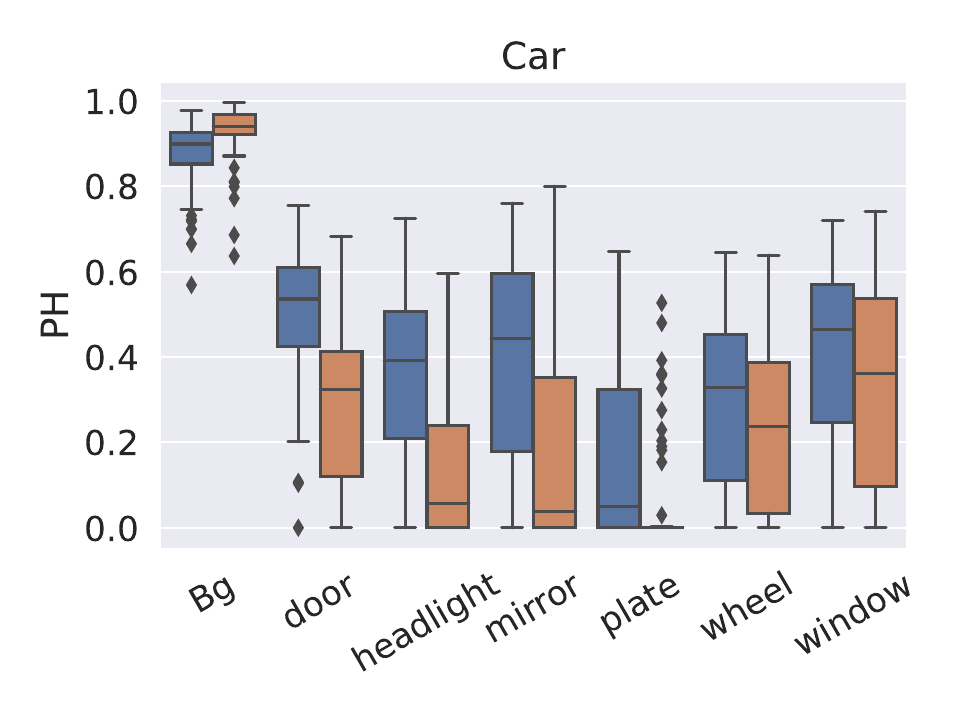}
	\end{subfigure}
	\begin{subfigure}{0.325\textwidth}
		\centering
		\includegraphics[width=\linewidth]{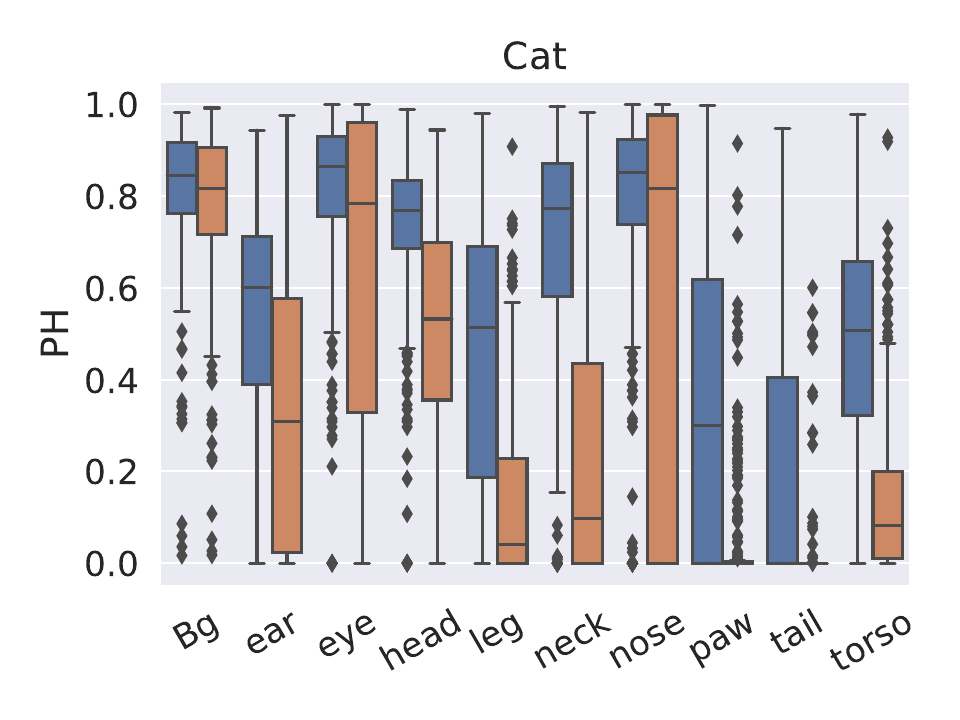}
	\end{subfigure}
	\begin{subfigure}{0.325\textwidth}
		\centering
		\includegraphics[width=\linewidth]{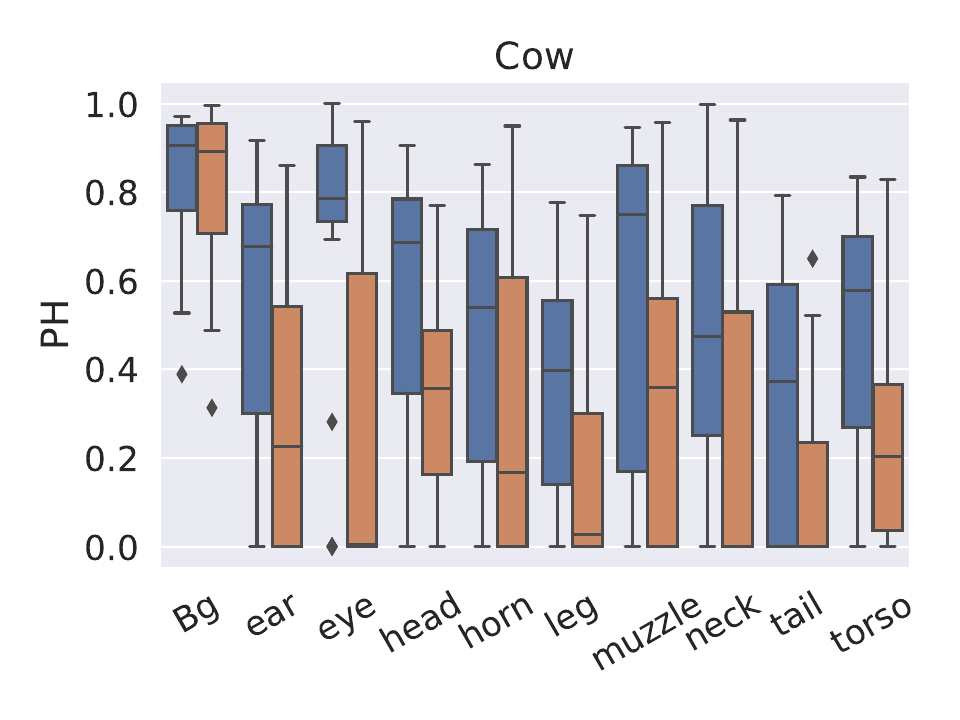}
	\end{subfigure}
	\begin{subfigure}{0.325\textwidth}
		\centering
		\includegraphics[width=\linewidth]{images/F1/pascalvoc/dog.pdf}
	\end{subfigure}
	\begin{subfigure}{0.325\textwidth}
		\centering
		\includegraphics[width=\linewidth]{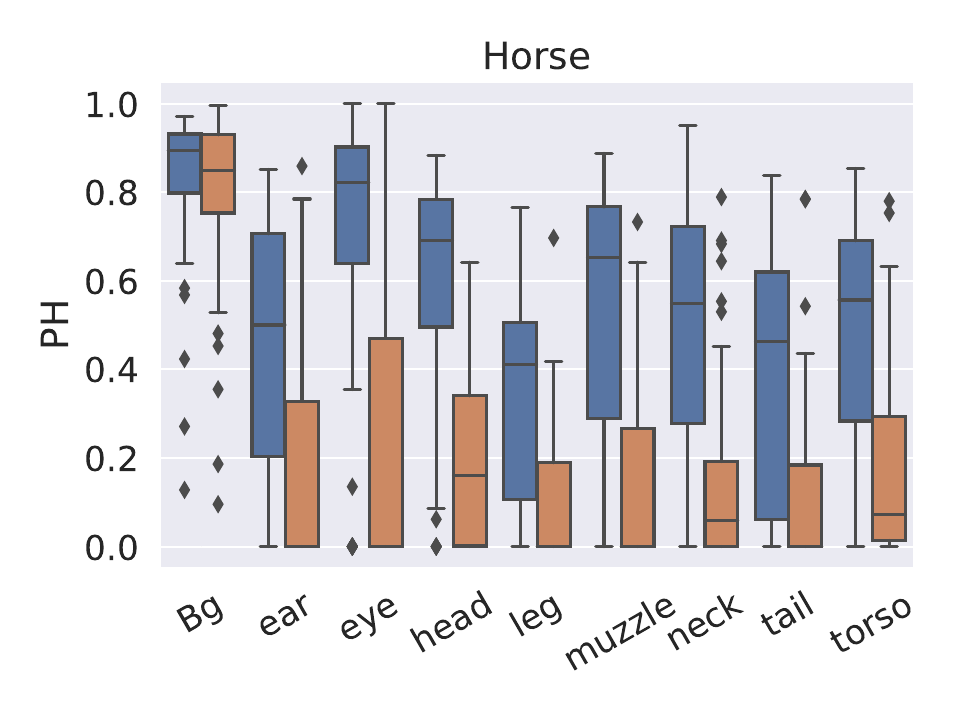}
	\end{subfigure}
	\begin{subfigure}{0.325\textwidth}
		\centering
		\includegraphics[width=\linewidth]{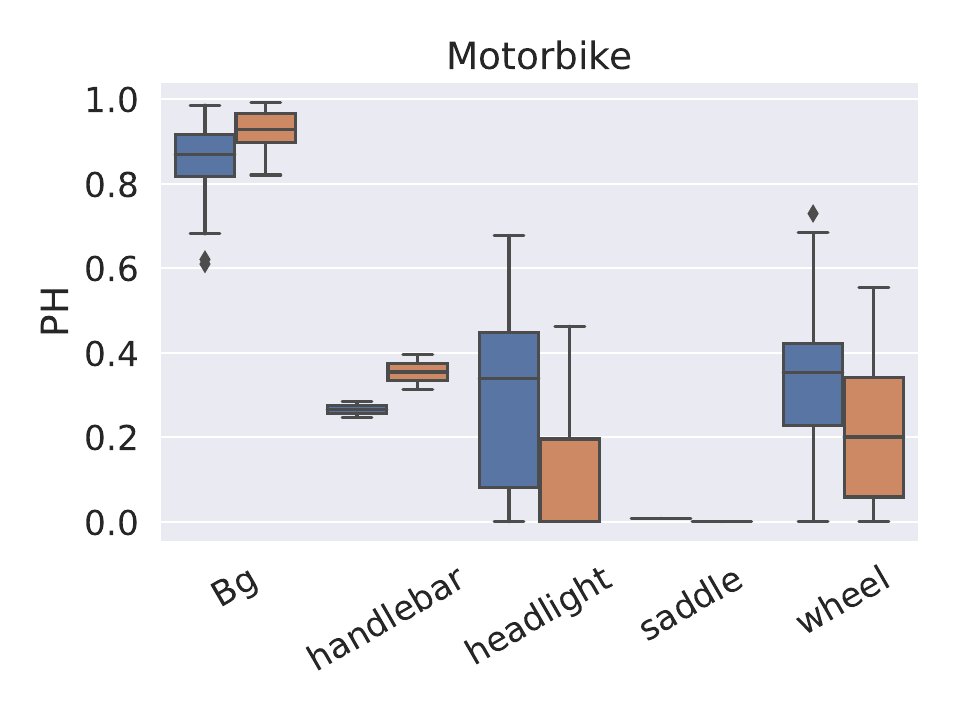}
	\end{subfigure}
	\begin{subfigure}{0.325\textwidth}
		\centering
		\includegraphics[width=\linewidth]{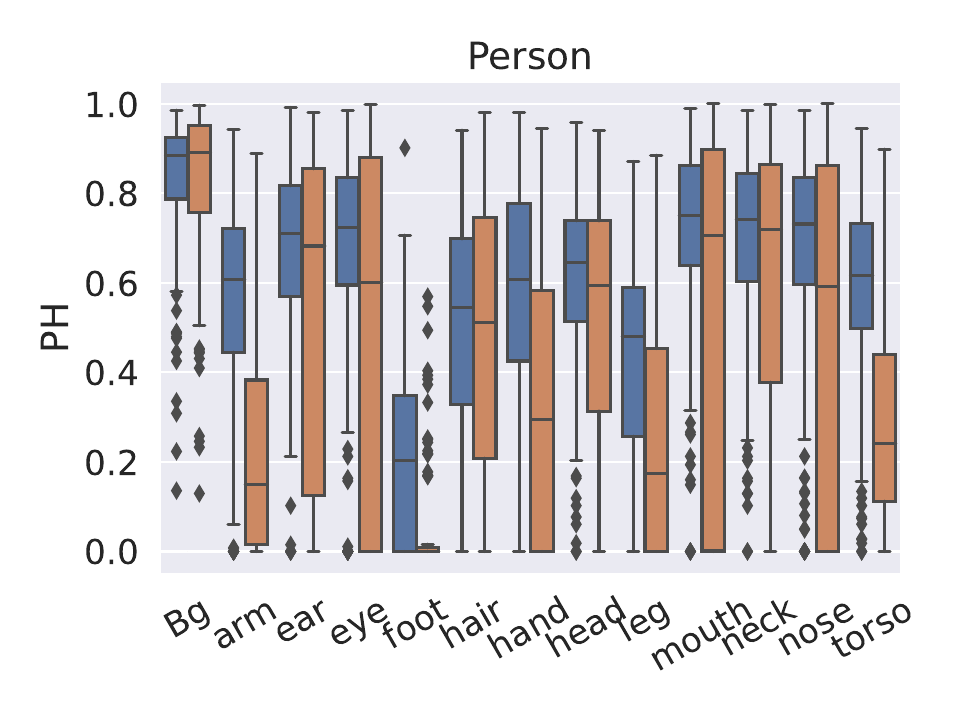}
	\end{subfigure}
	\begin{subfigure}{0.325\textwidth}
		\centering
		\includegraphics[width=\linewidth]{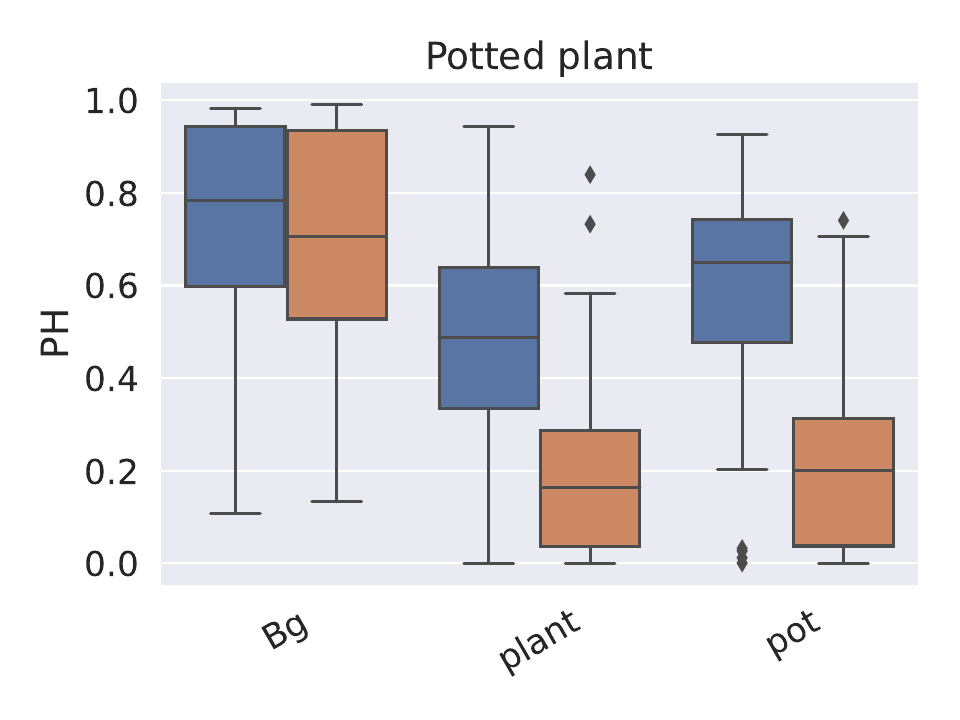}
	\end{subfigure}
	\begin{subfigure}{0.325\textwidth}
		\centering
		\includegraphics[width=\linewidth]{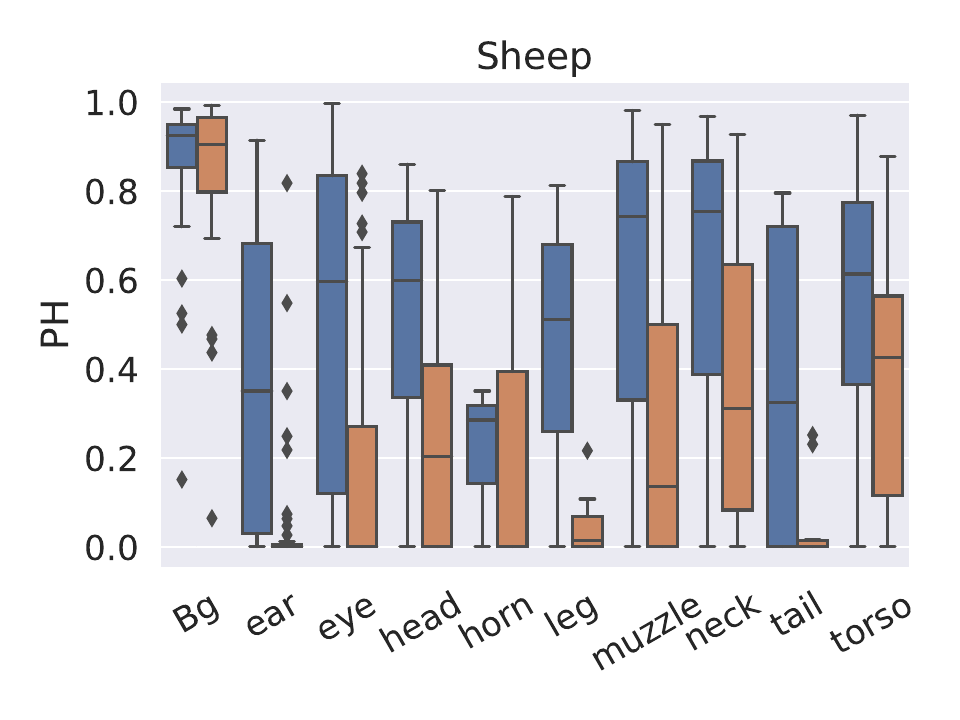}
	\end{subfigure}
	\caption{Exp. 1: PQAH analysis for DNN models on the Pascal-Part dataset. On the X-axis, various parts are displayed, with 'Bg' denoting the background.}
	\label{fig:pascalvoc_f1_fl}
\end{figure*}

\begin{figure*}[!tbh]
	\centering
	\begin{subfigure}{0.325\textwidth}
		\centering
		\includegraphics[width=\linewidth]{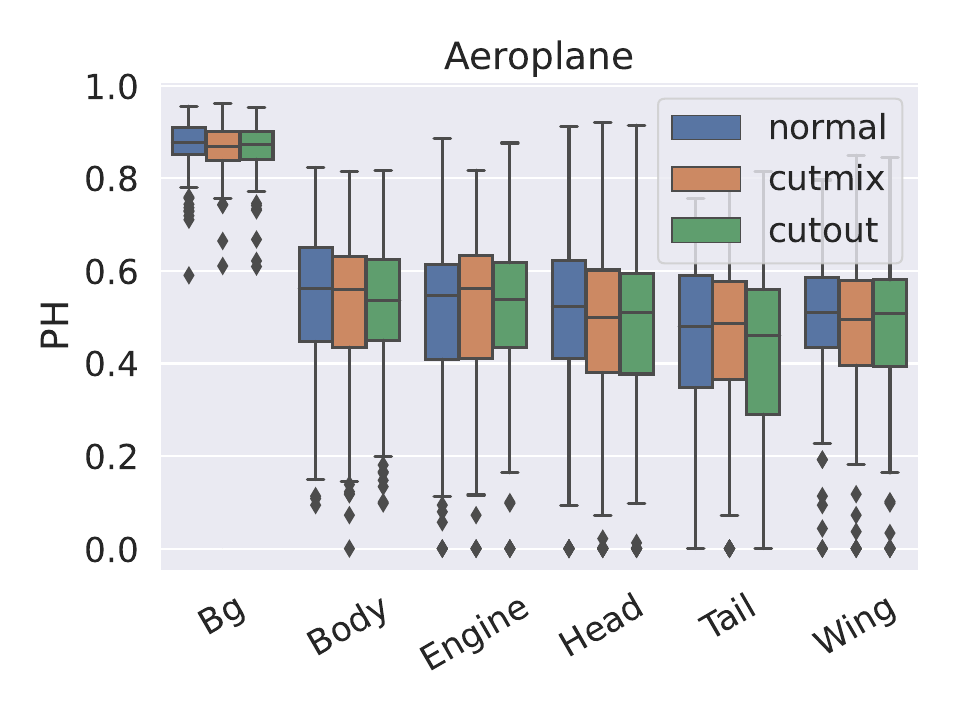}
	\end{subfigure}
	\begin{subfigure}{0.325\textwidth}
		\centering
		\includegraphics[width=\linewidth]{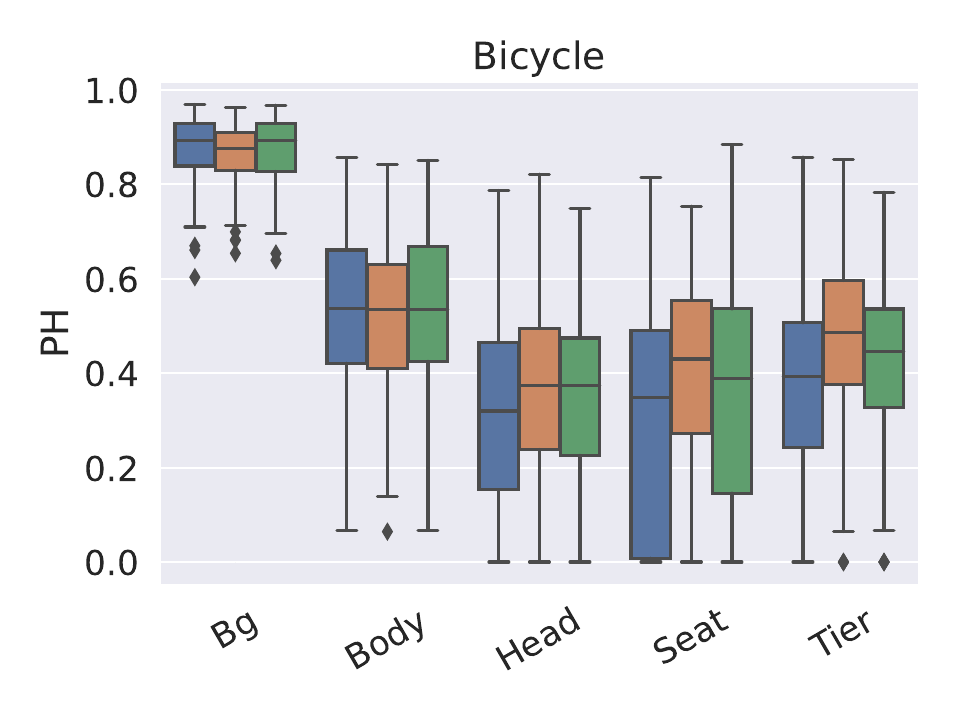}
	\end{subfigure}
	\begin{subfigure}{0.325\textwidth}
		\centering
		\includegraphics[width=\linewidth]{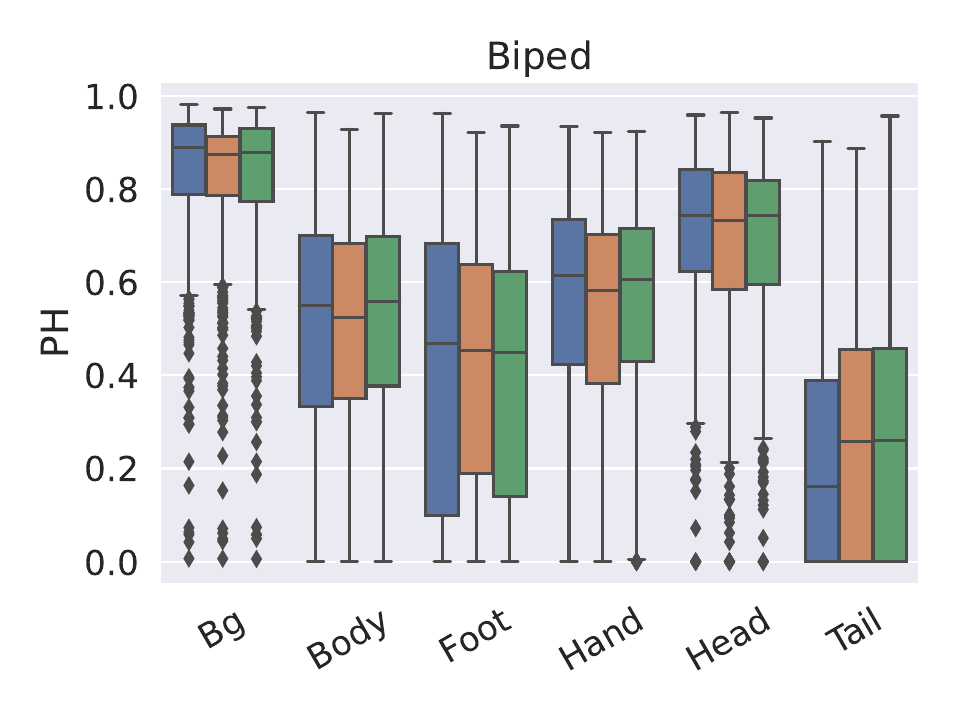}
	\end{subfigure}
	\\
	\begin{subfigure}{0.325\textwidth}
		\centering
		\includegraphics[width=\linewidth]{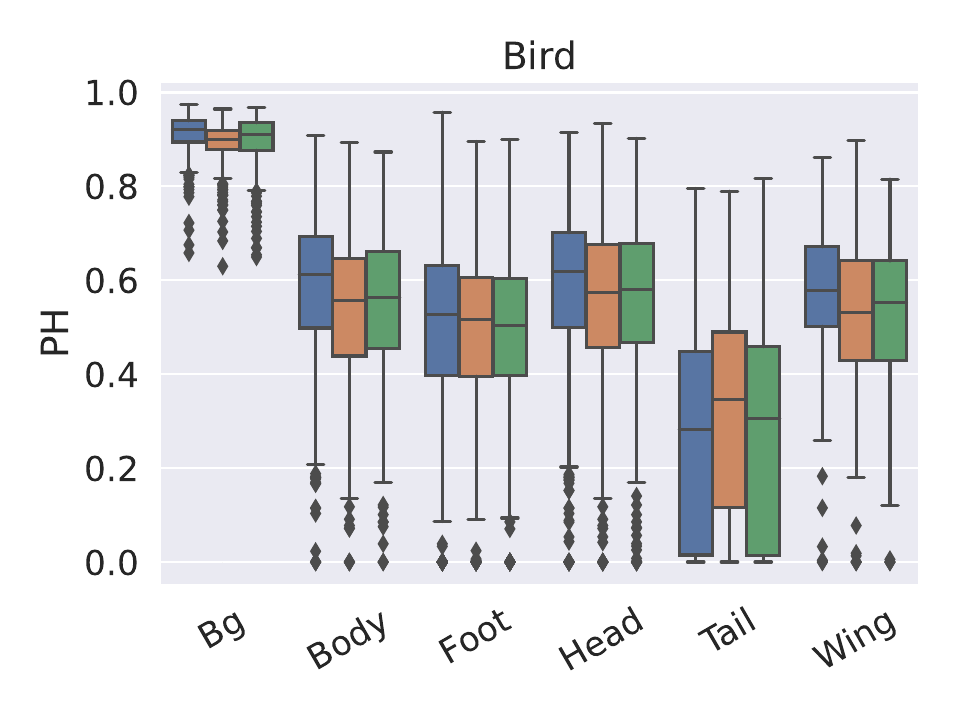}
	\end{subfigure}
	\begin{subfigure}{0.325\textwidth}
		\centering
		\includegraphics[width=\linewidth]{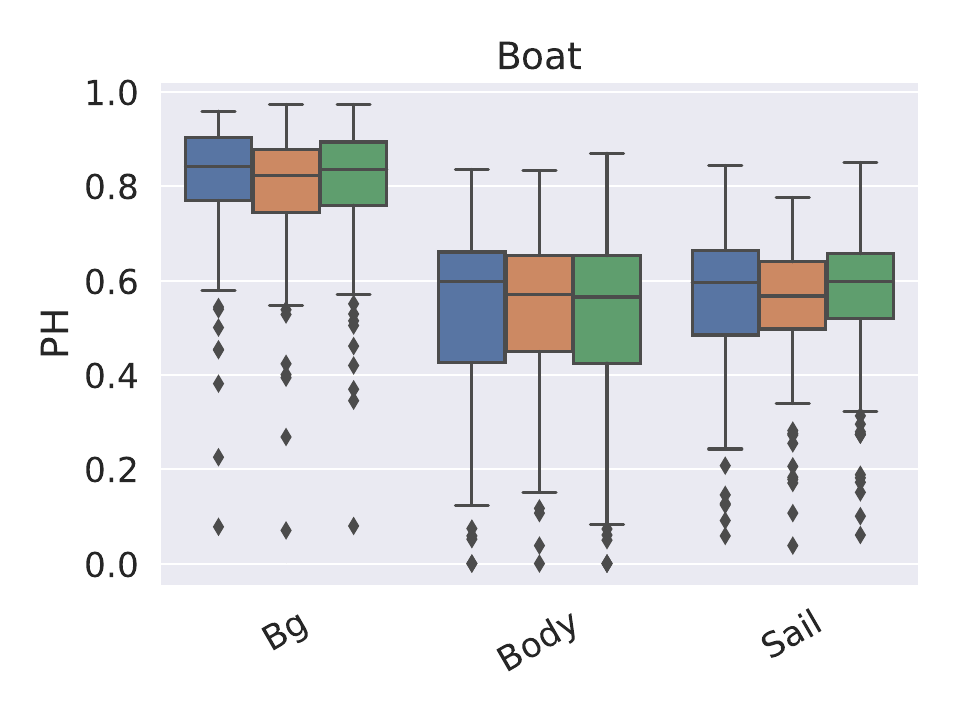}
	\end{subfigure}
	\begin{subfigure}{0.325\textwidth}
		\centering
		\includegraphics[width=\linewidth]{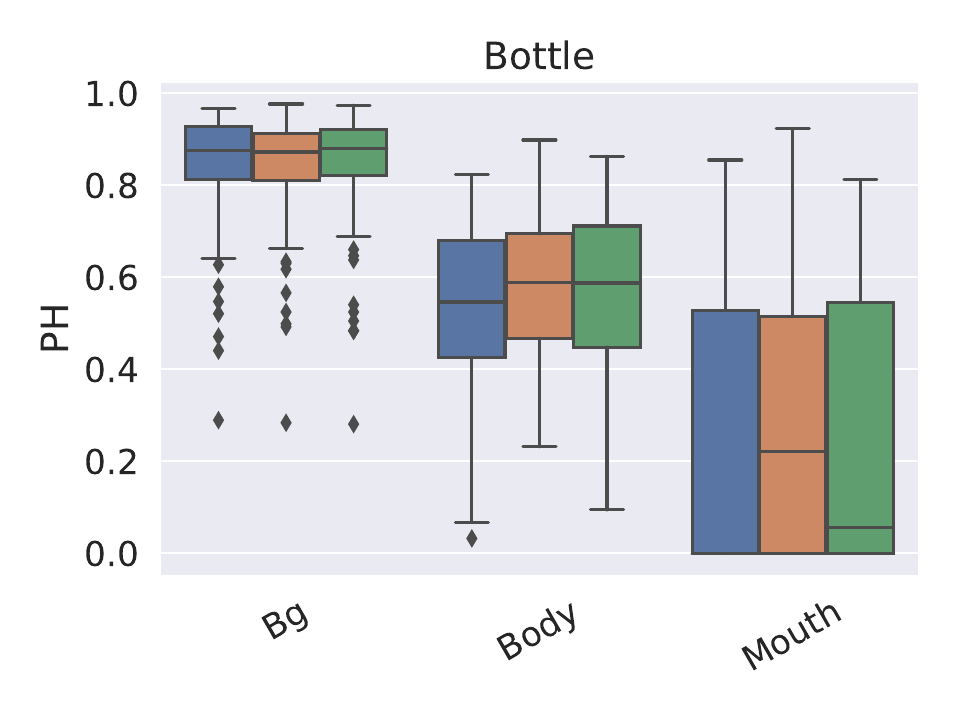}
	\end{subfigure}
	\begin{subfigure}{0.325\textwidth}
		\centering
		\includegraphics[width=\linewidth]{images/F1/partimagenet_dataaug_compare/Car.pdf}
	\end{subfigure}
	\begin{subfigure}{0.325\textwidth}
		\centering
		\includegraphics[width=\linewidth]{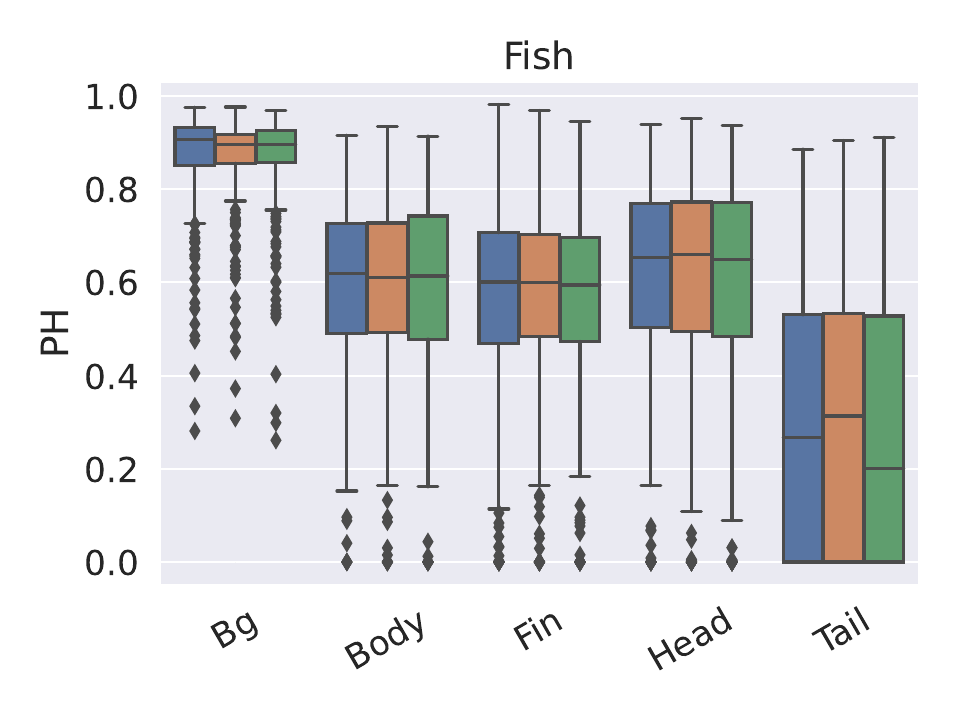}
	\end{subfigure}
	\begin{subfigure}{0.325\textwidth}
		\centering
		\includegraphics[width=\linewidth]{images/F1/partimagenet_dataaug_compare/Quadruped.pdf}
	\end{subfigure}
	\\
	\begin{subfigure}{0.325\textwidth}
		\centering
		\includegraphics[width=\linewidth]{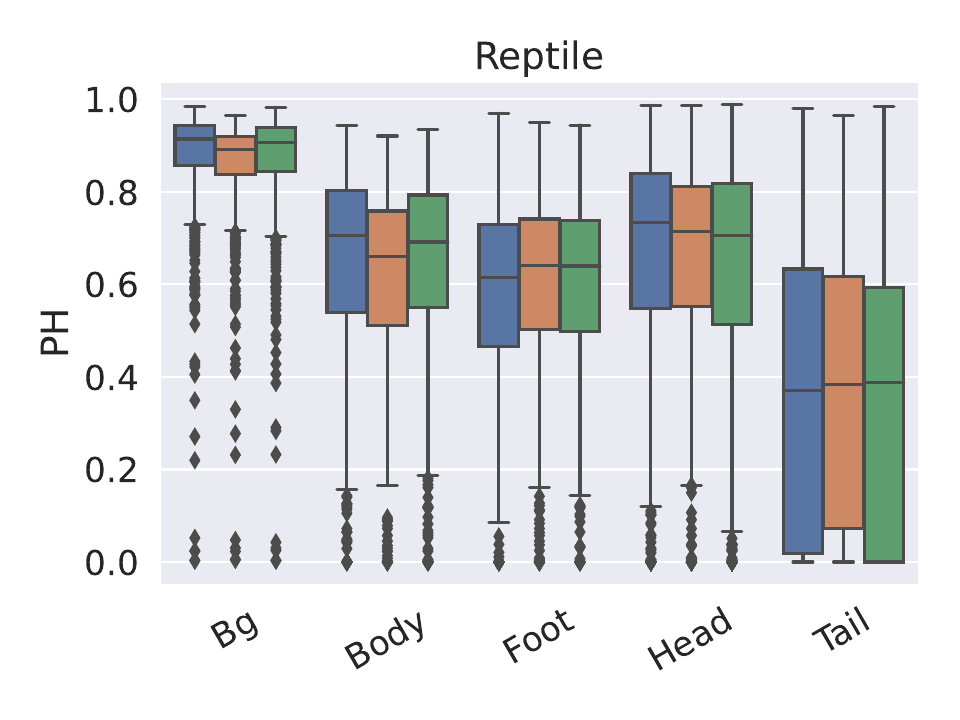}
	\end{subfigure}
	\begin{subfigure}{0.325\textwidth}
		\centering
		\includegraphics[width=\linewidth]{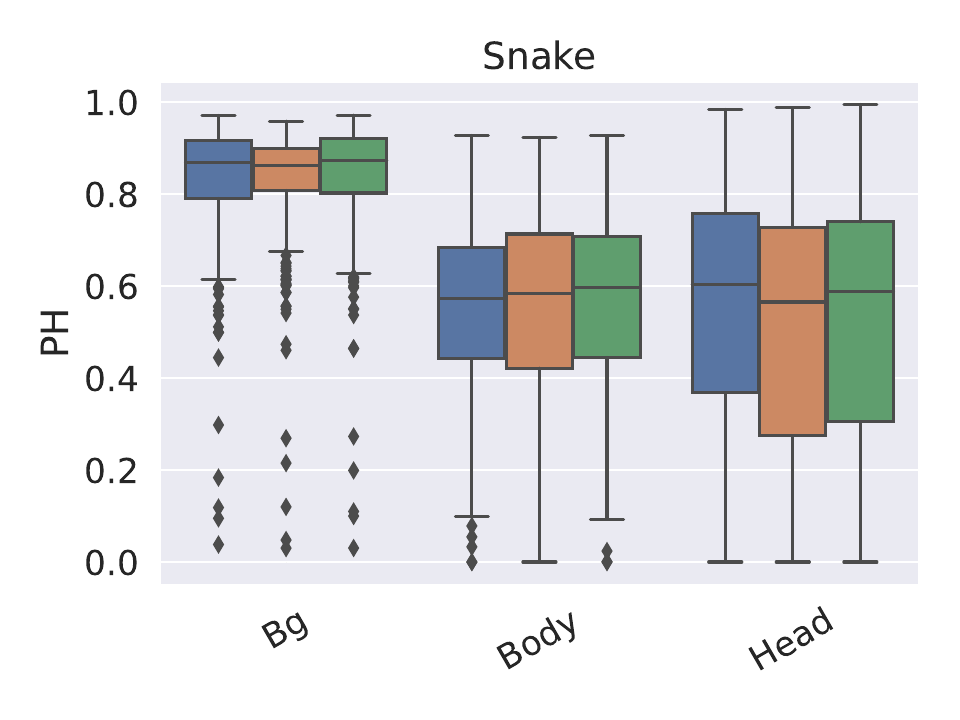}
	\end{subfigure}
	\caption{EXP. 2: PQAH analysis of data augmentation approaches, namely Cutout and CutMix, on the PartImageNet dataset. The backbone network is ResNet-50. On the X-axis, various parts are displayed, with 'Bg' denoting the background.}
	\label{fig:dataaug_f1_fl}
\end{figure*}

Figures \ref{fig:partimagenet_f1_fl}, \ref{fig:pascalvoc_f1_fl}, \ref{fig:dataaug_f1_fl}, \ref{fig:sess-puz}, and \ref{fig:PQAH-eval_fl} provide a comprehensive presentation of the PQAH results across Experiments 1 to 4 in the main manuscript. Additionally, Figures \ref{fig:partimagenetallnetworks}, \ref{fig:pascalallnetworks}, \ref{fig:pascal_dataaug}, \ref{fig:pascal_enhance}, and \ref{fig:partimagenet_heatmap} depict the corresponding heatmaps associated with the PQAH results from Experiments 1 through 4.

By analysing the PQAH results shown in Figures \ref{fig:partimagenet_f1}, \ref{fig:pascalvoc_f1}, \ref{fig:dataaug_f1}, \ref{fig:sess-puz}, and \ref{fig:PQAH-eval}, it is evident that these results offer a detailed analysis of the extracted heatmaps. This level of detail is invaluable for identifying the strengths and weaknesses of the networks and for assessing the effectiveness of different heatmap extraction methods. However, it is difficult to get a similar analysis by comparing the individual heatmaps as given in Figures \ref{fig:partimagenetallnetworks}, \ref{fig:pascalallnetworks}, \ref{fig:pascal_dataaug}, \ref{fig:pascal_enhance}, and \ref{fig:partimagenet_heatmap}. For instance, as illustrated in Figure \ref{fig:pascal_dataaug}, the heatmaps extracted using three distinct methods display striking visual similarities, making it very difficult to compare them based on these individual results. In contrast, the PQAH results, displayed in Figure \ref{fig:dataaug_f1}, capture fine-grained differences.

\begin{figure*}
	\centering	
	\begin{subfigure}{0.325\textwidth}
		\centering
		\includegraphics[width=\linewidth]{images/F1/pascalvoc_puzzle-cam-sess/aeroplane.pdf}
	\end{subfigure}
	\begin{subfigure}{0.325\textwidth}
		\centering
		\includegraphics[width=\linewidth]{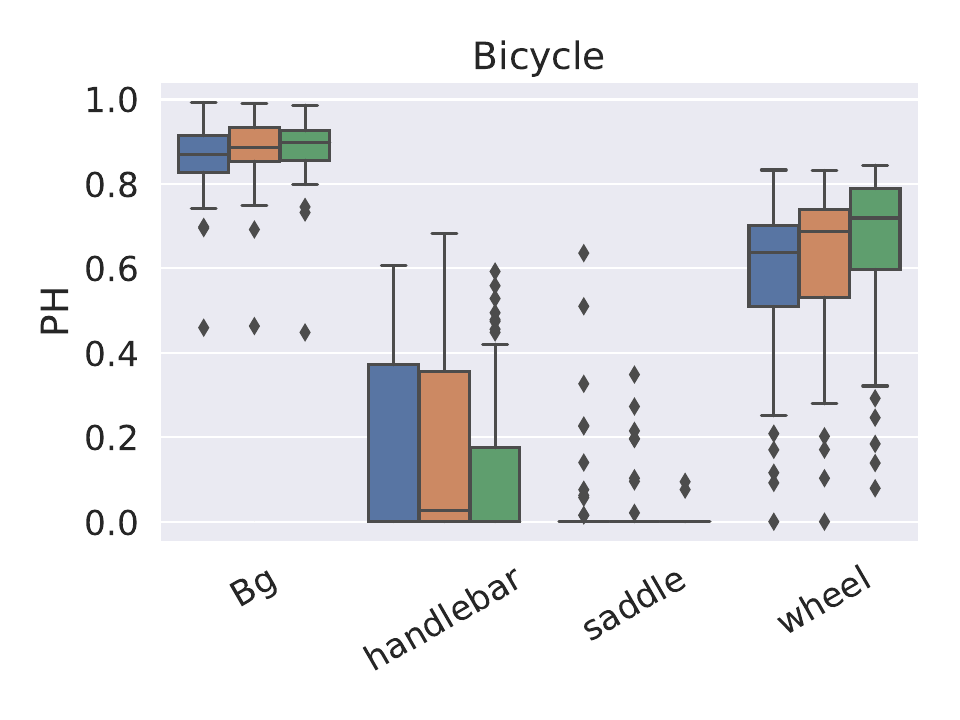}
	\end{subfigure}
	\begin{subfigure}{0.325\textwidth}
		\centering
		\includegraphics[width=\linewidth]{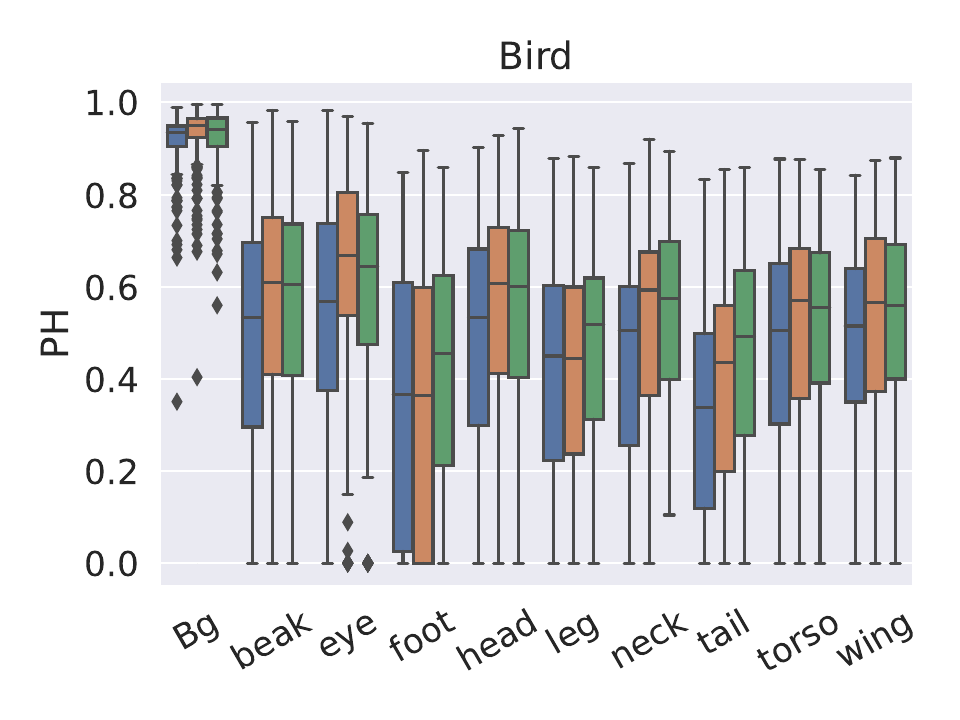}
	\end{subfigure}
	\\
	\begin{subfigure}{0.325\textwidth}
		\centering
		\includegraphics[width=\linewidth]{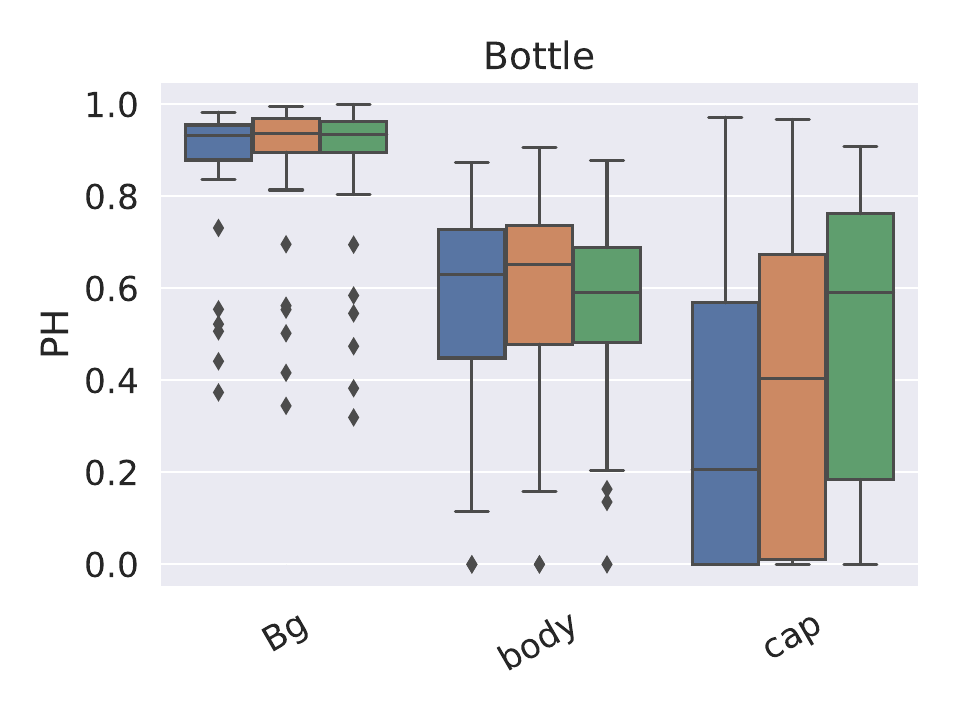}
	\end{subfigure}
	\begin{subfigure}{0.325\textwidth}
		\centering
		\includegraphics[width=\linewidth]{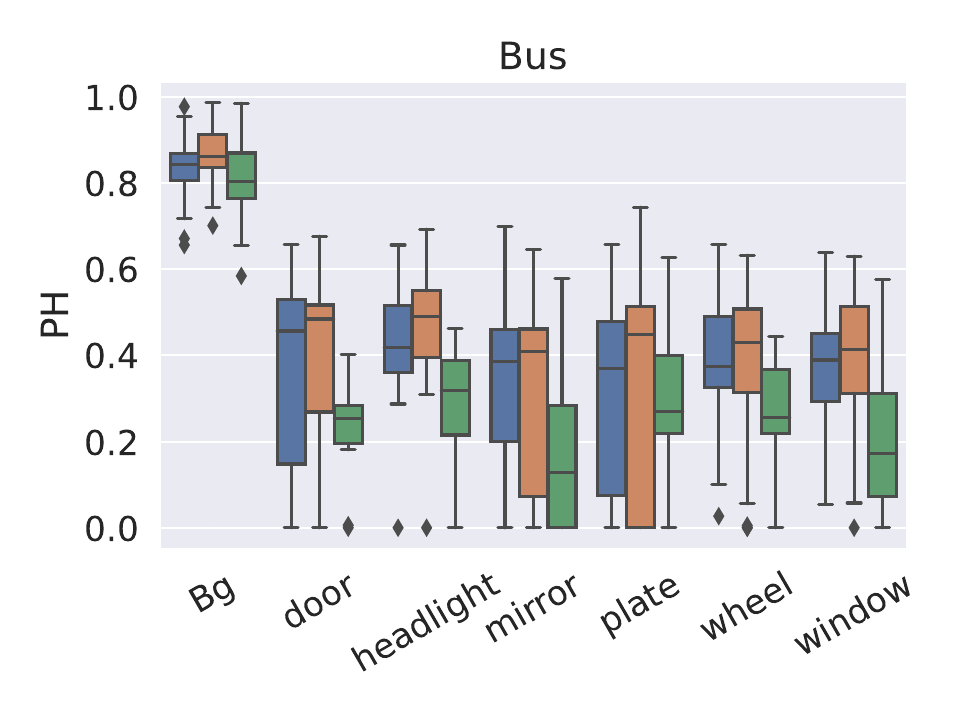}
	\end{subfigure}
	\begin{subfigure}{0.325\textwidth}
		\centering
		\includegraphics[width=\linewidth]{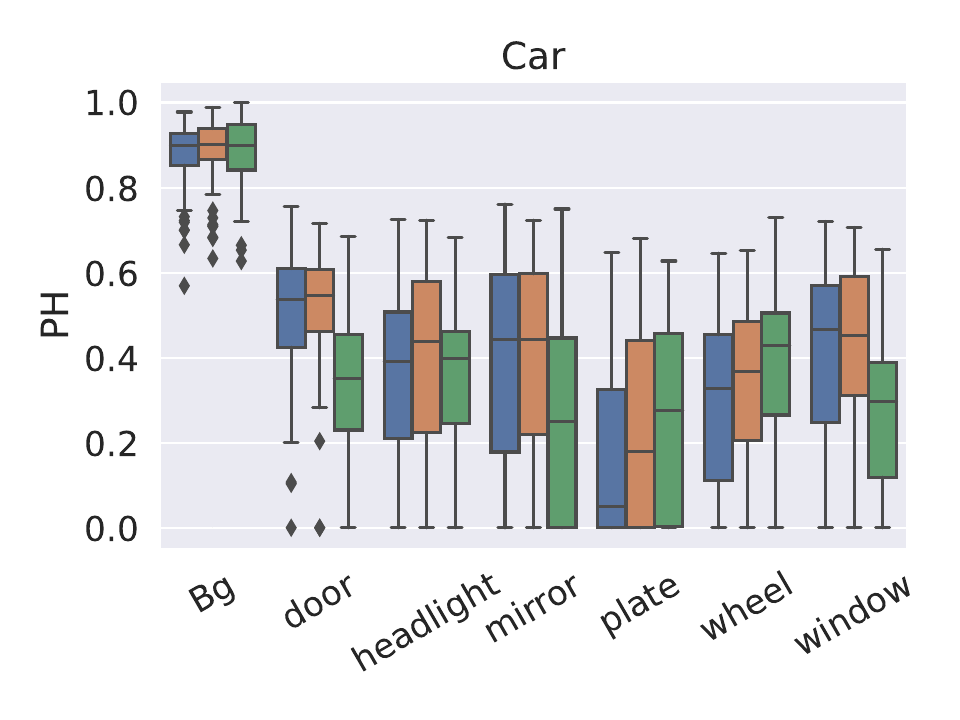}
	\end{subfigure}
	\begin{subfigure}{0.325\textwidth}
		\centering
		\includegraphics[width=\linewidth]{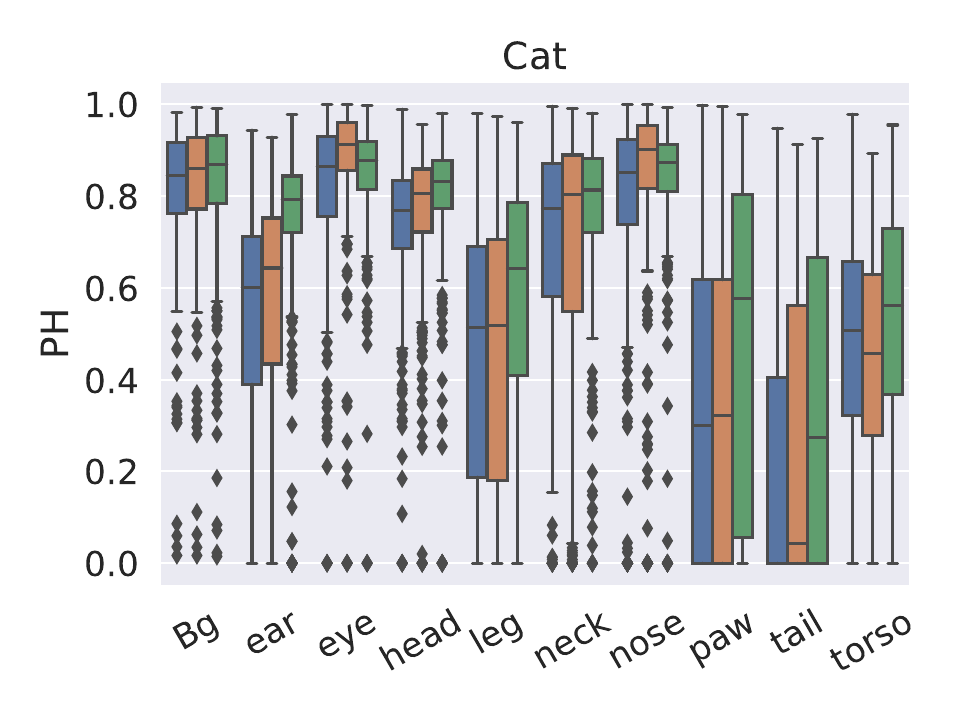}
	\end{subfigure}
	\begin{subfigure}{0.325\textwidth}
		\centering
		\includegraphics[width=\linewidth]{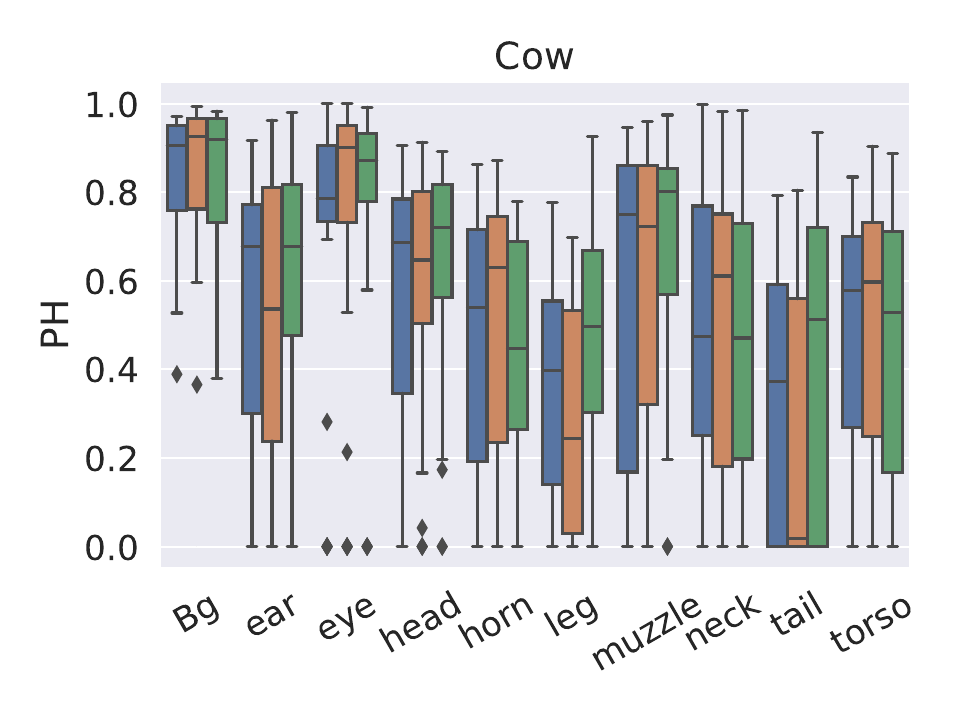}
	\end{subfigure}
	\begin{subfigure}{0.325\textwidth}
		\centering
		\includegraphics[width=\linewidth]{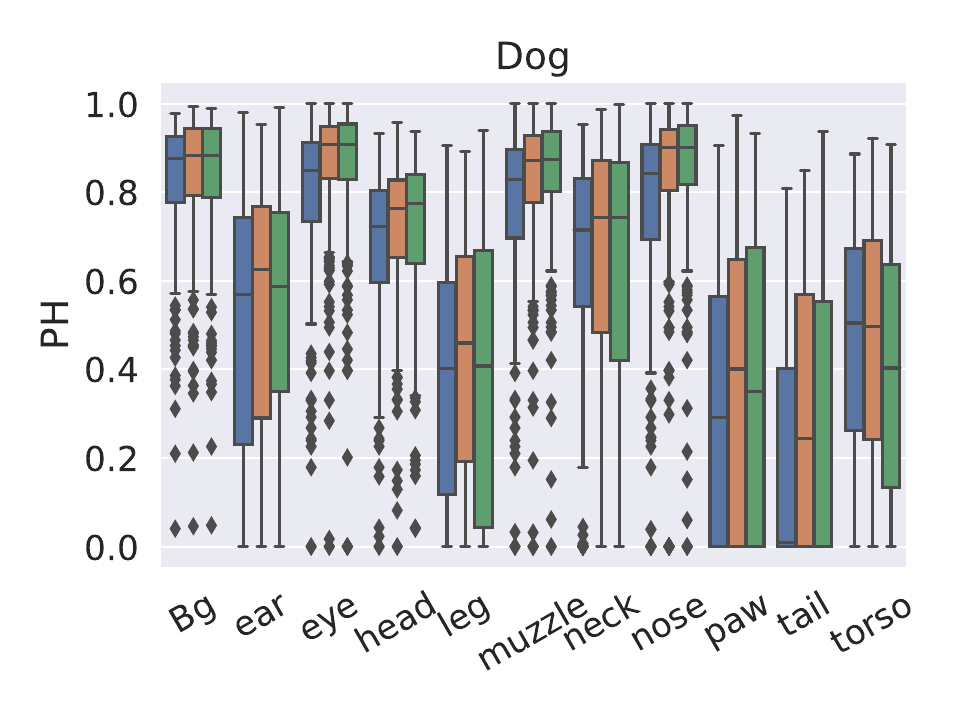}
	\end{subfigure}
	\begin{subfigure}{0.325\textwidth}
		\centering
		\includegraphics[width=\linewidth]{images/F1/pascalvoc_puzzle-cam-sess/horse.pdf}
	\end{subfigure}
	\begin{subfigure}{0.325\textwidth}
		\centering
		\includegraphics[width=\linewidth]{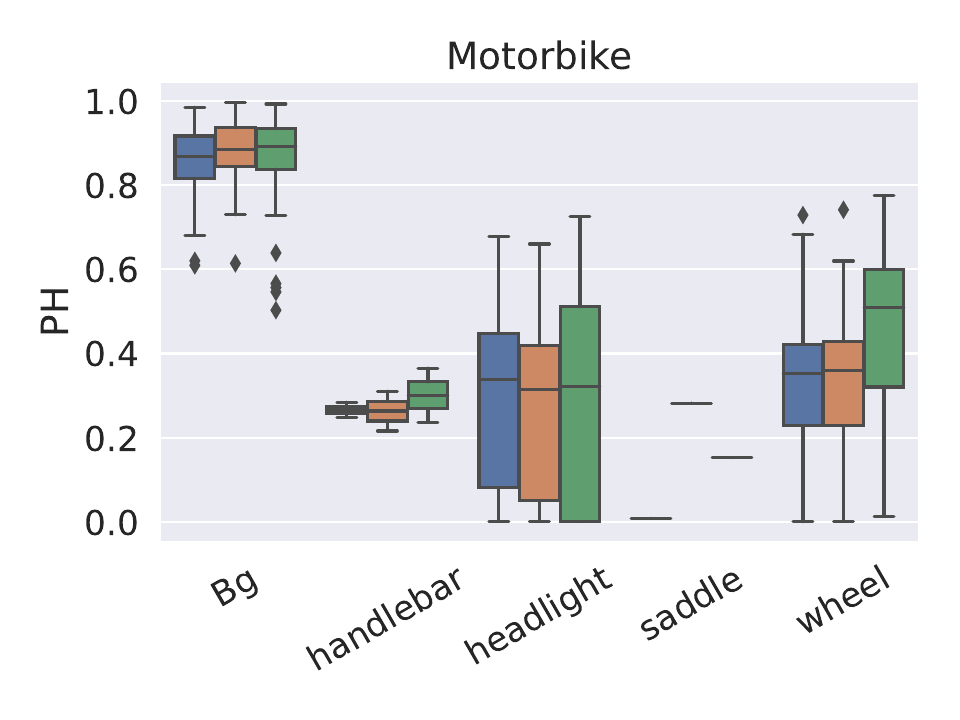}
	\end{subfigure}
	\begin{subfigure}{0.325\textwidth}
		\centering
		\includegraphics[width=\linewidth]{images/F1/pascalvoc_puzzle-cam-sess/person.pdf}
	\end{subfigure}
	\begin{subfigure}{0.325\textwidth}
		\centering
		\includegraphics[width=\linewidth]{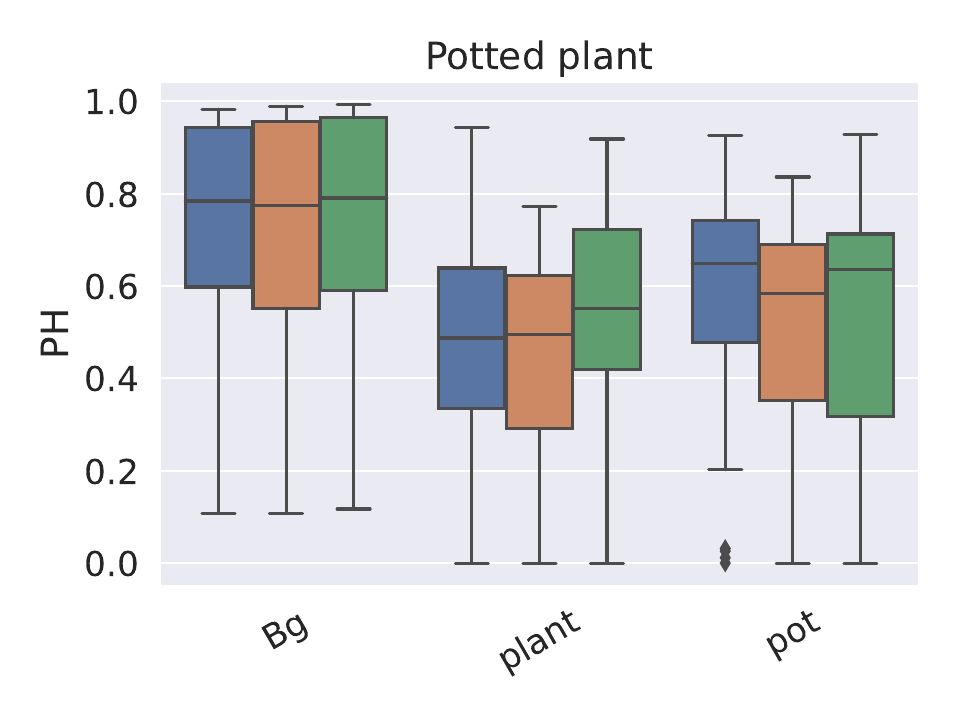}
	\end{subfigure}
	\begin{subfigure}{0.325\textwidth}
		\centering
		\includegraphics[width=\linewidth]{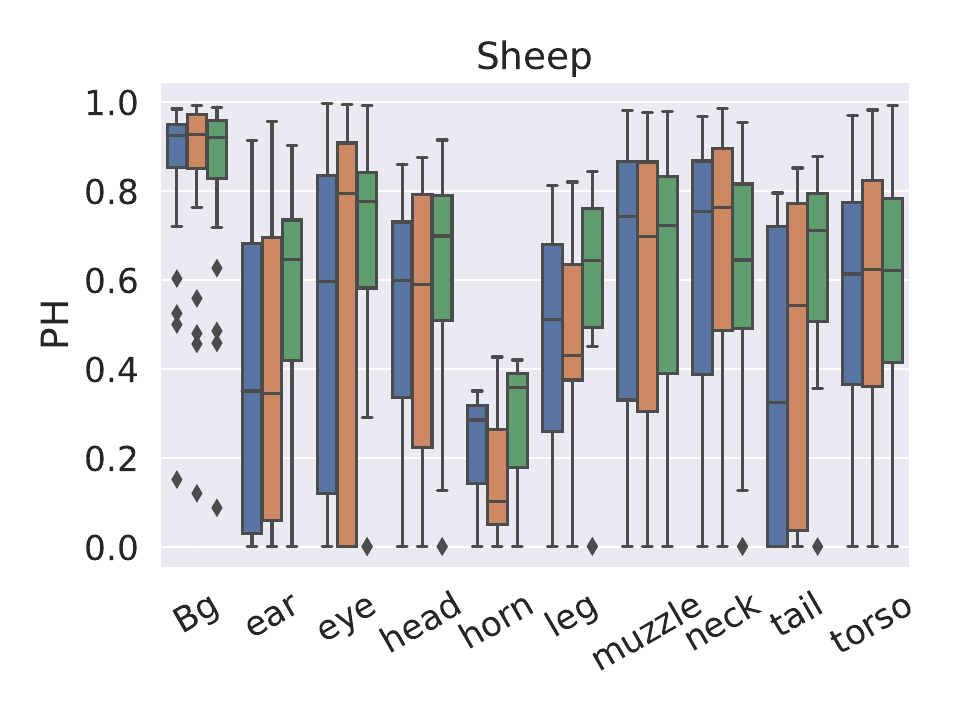}
	\end{subfigure}
	\caption{Exp. 3: PQAH analysis of saliency enhancing approaches, namely Puzzle-CAM and SESS, on the PASCAL-Part dataset. The backbone network is ResNet-50. On the X-axis, various parts are displayed, with 'Bg' denoting the background.}
	\label{fig:sess-puz_fl}
\end{figure*}

\begin{figure*}[tbh!]
	\centering
	\begin{subfigure}{0.3\textwidth}
		\centering
		\includegraphics[width=\linewidth]{images/F1/partimagenet_eval/Aeroplane.pdf}
	\end{subfigure}
	\begin{subfigure}{0.3\textwidth}
		\centering
		\includegraphics[width=\linewidth]{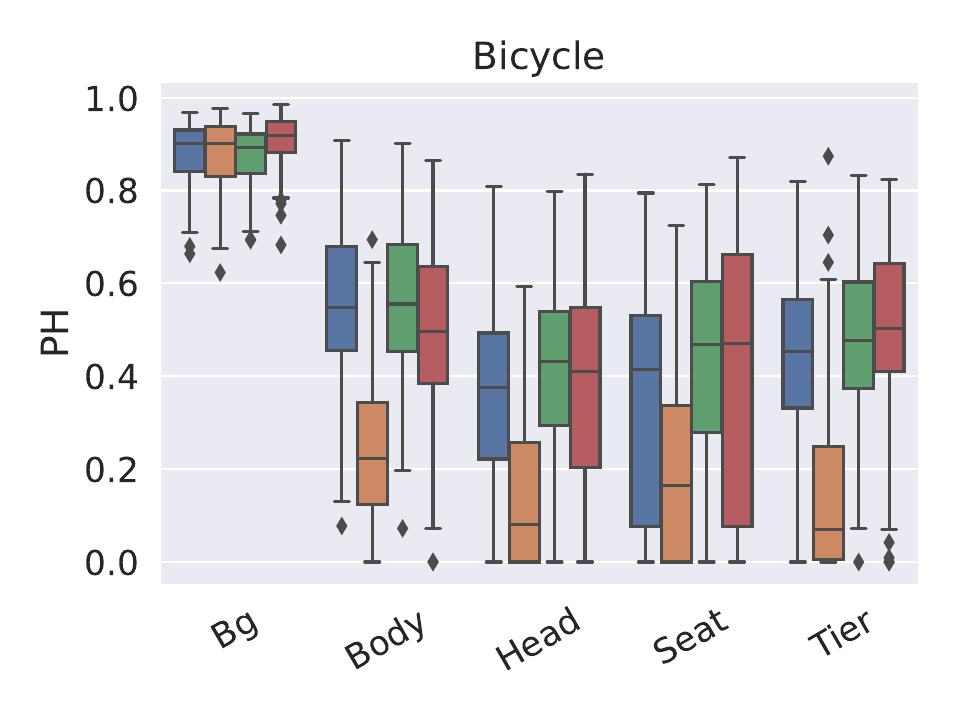}
	\end{subfigure}
	\begin{subfigure}{0.33\textwidth}
		\centering
		\includegraphics[width=\linewidth]{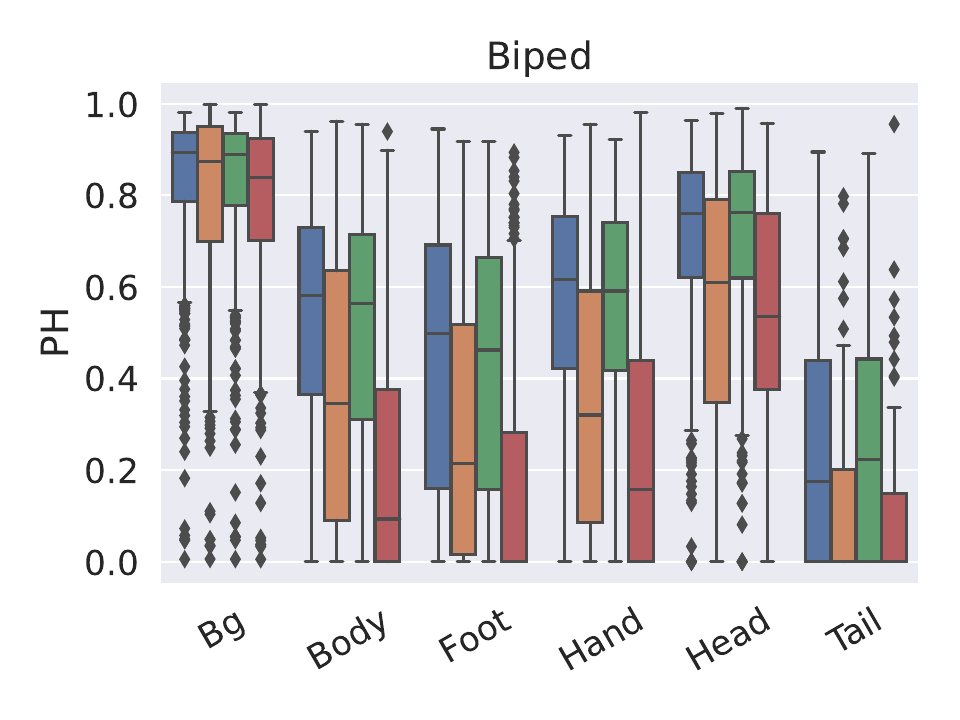}
	\end{subfigure}
	\\
	\begin{subfigure}{0.325\textwidth}
		\centering
		\includegraphics[width=\linewidth]{images/F1/partimagenet_eval/Bird.pdf}
	\end{subfigure}
	\begin{subfigure}{0.325\textwidth}
		\centering
		\includegraphics[width=\linewidth]{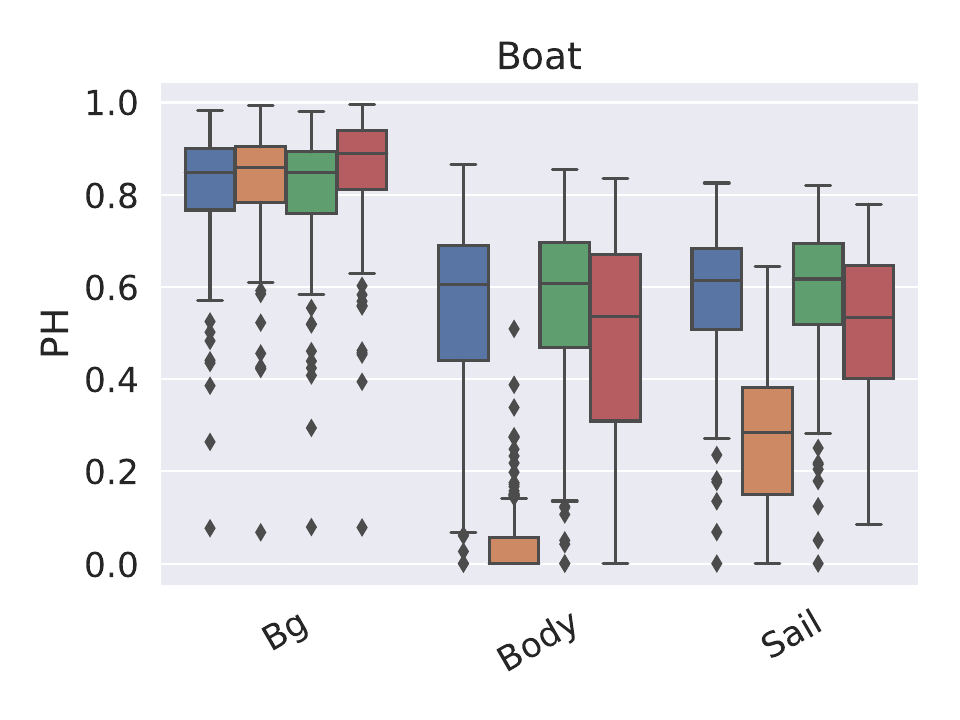}
	\end{subfigure}
	\begin{subfigure}{0.325\textwidth}
		\centering
		\includegraphics[width=\linewidth]{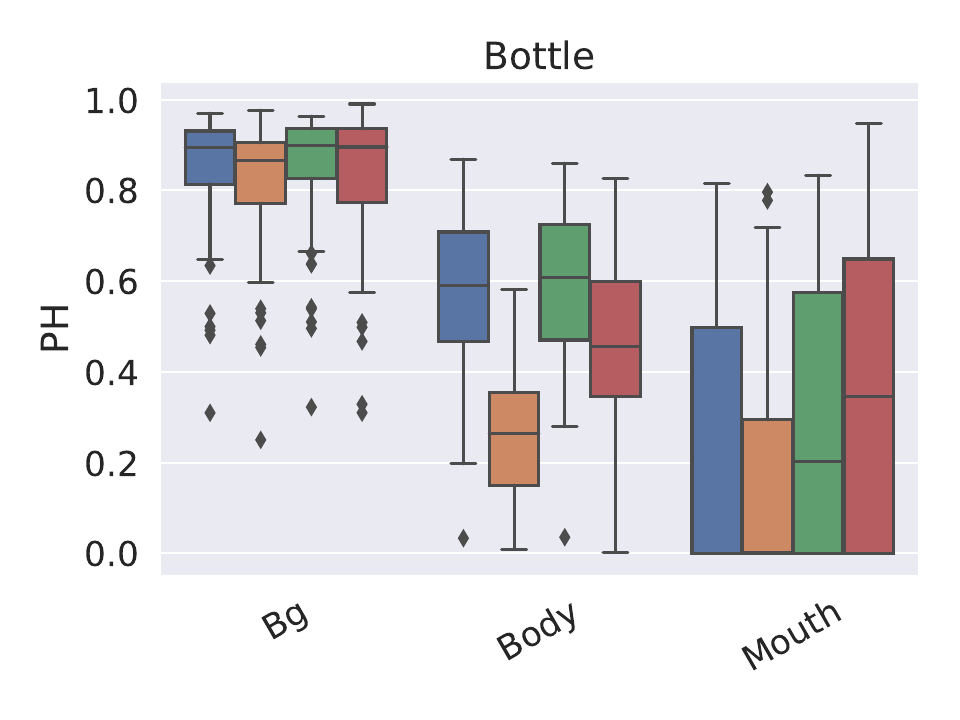}
	\end{subfigure}
	\begin{subfigure}{0.325\textwidth}
		\centering
		\includegraphics[width=\linewidth]{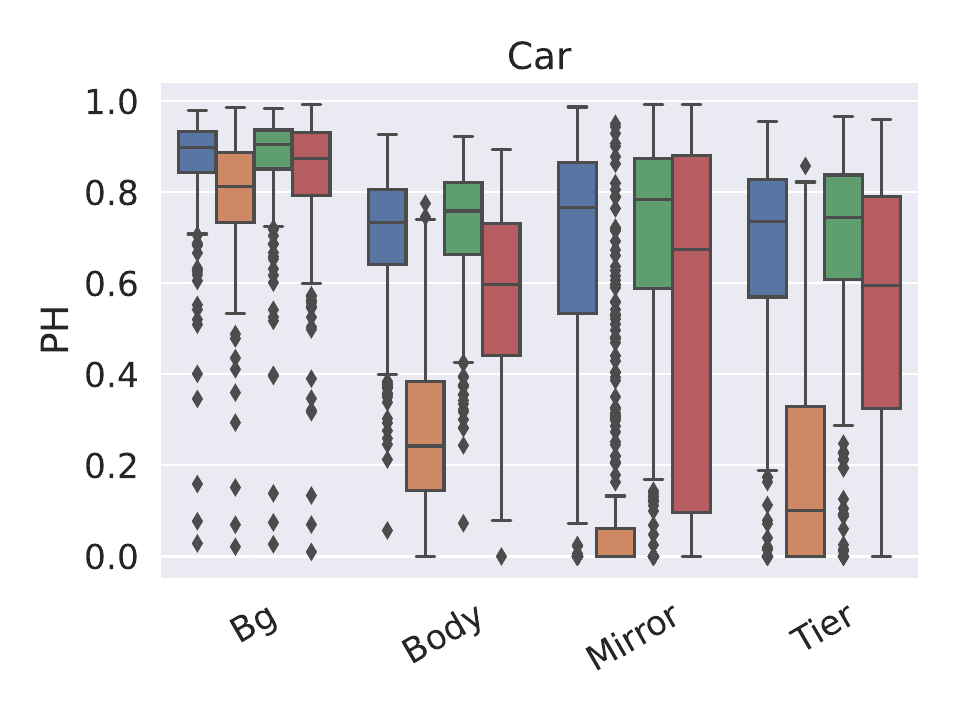}
	\end{subfigure}
	\begin{subfigure}{0.325\textwidth}
		\centering
		\includegraphics[width=\linewidth]{images/F1/partimagenet_eval/Fish.pdf}
	\end{subfigure}
	\begin{subfigure}{0.325\textwidth}
		\centering
		\includegraphics[width=\linewidth]{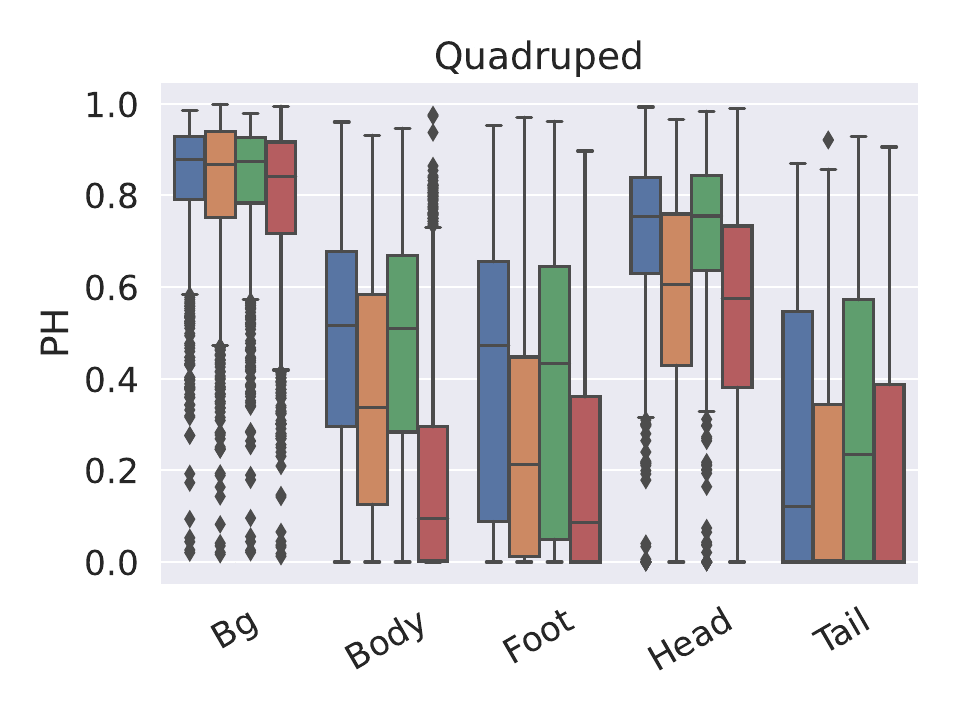}
	\end{subfigure}
	\\
	\begin{subfigure}{0.325\textwidth}
		\centering
		\includegraphics[width=\linewidth]{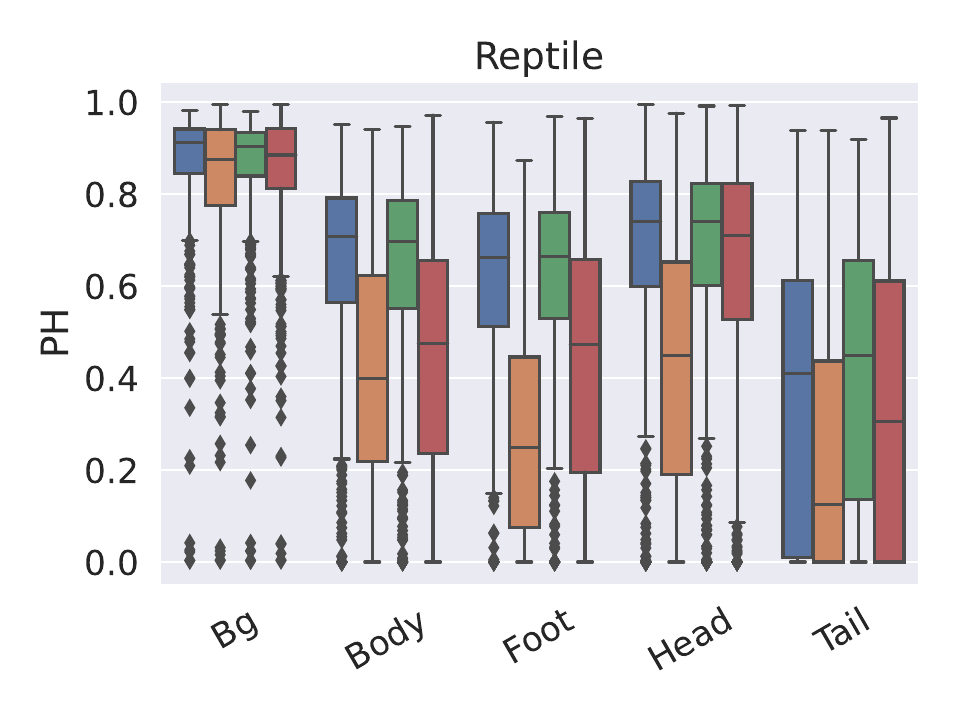}
	\end{subfigure}
	\begin{subfigure}{0.325\textwidth}
		\centering
		\includegraphics[width=\linewidth]{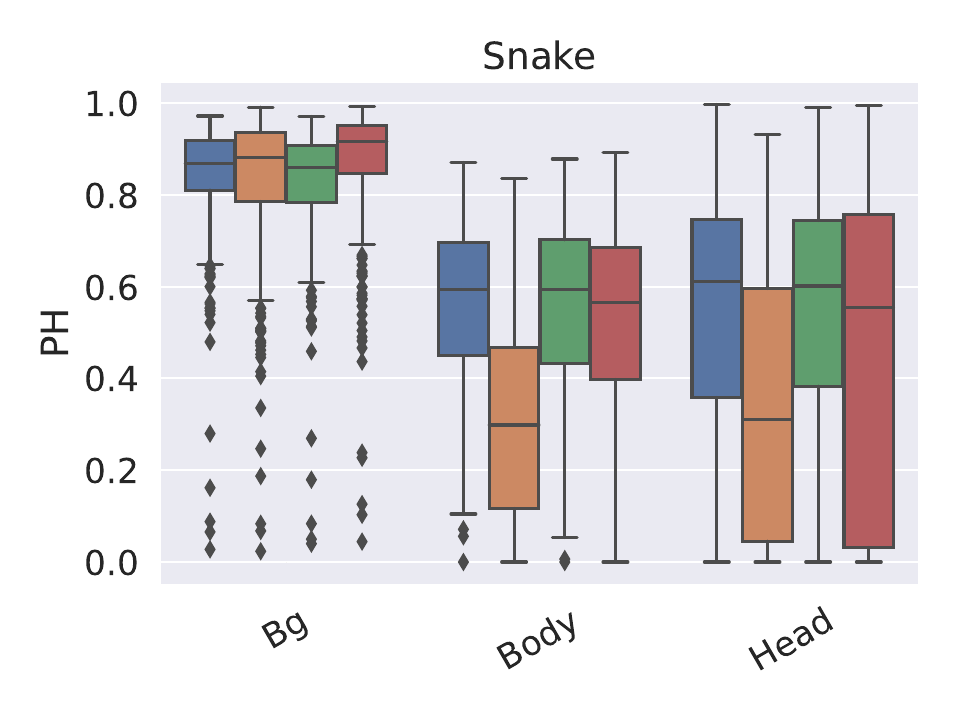}
	\end{subfigure}
	\caption{Exp. 4: Comparison of Heatmap Extraction Methods with PQAH on PartImageNet. The backbone network is ResNet-50. On the X-axis, various parts are displayed, with 'Bg' denoting the background.}
	\label{fig:PQAH-eval_fl}
\end{figure*}

\begin{figure}[!bh]
	\centering
	\includegraphics[width=0.95\linewidth]{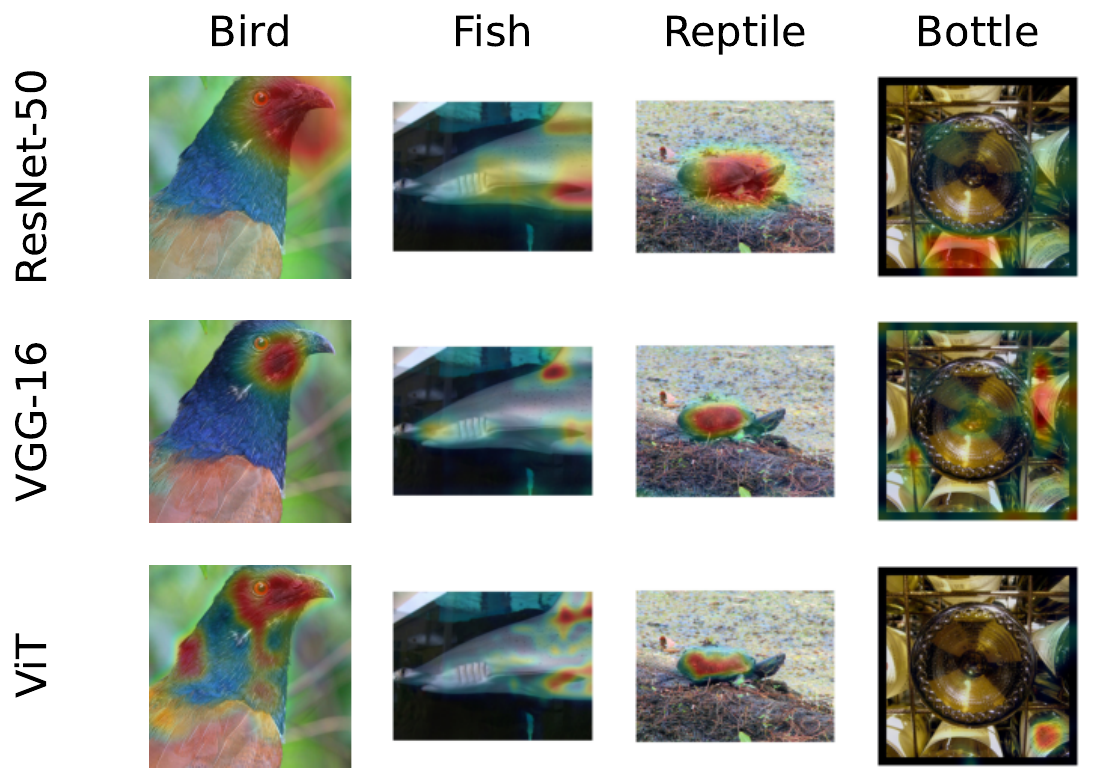}
	\caption{Visualisation of heatmaps on randomly selected images from the PartImageNet dataset  (EXP. 1). The visualisation method is GradCam + SESS.}
	\label{fig:partimagenetallnetworks}
\end{figure}

\begin{figure}[!tbh]
	\centering
	\includegraphics[width=0.95\linewidth]{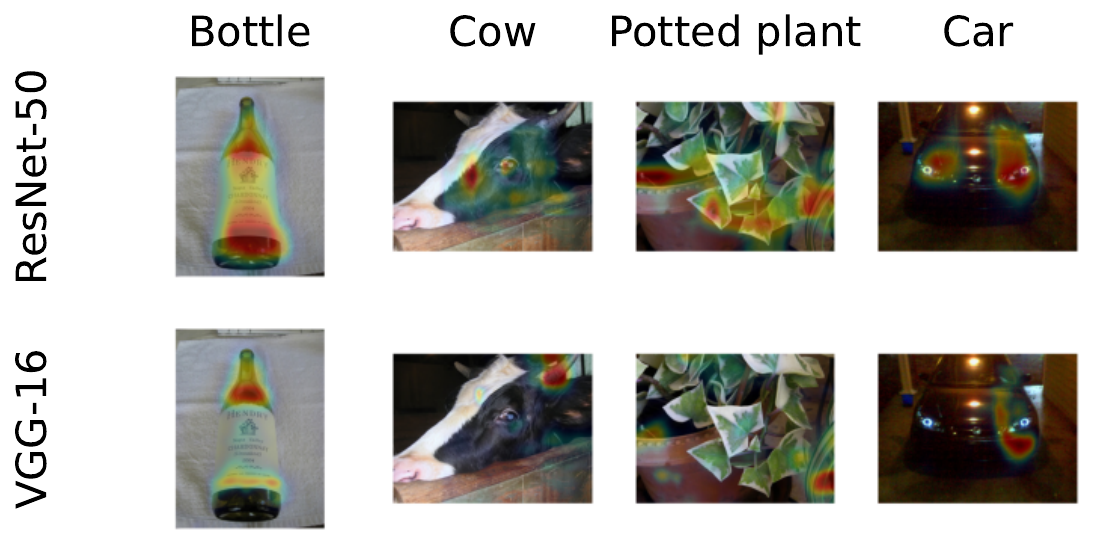}
	\caption{EXP. 1: Visualisation of heatmaps on randomly selected images from the Pascal-Part dataset. The visualisation method is GradCam + SESS.}
	\label{fig:pascalallnetworks}
\end{figure}

\begin{figure}[!tbh]
	\centering
	\includegraphics[width=0.95\linewidth]{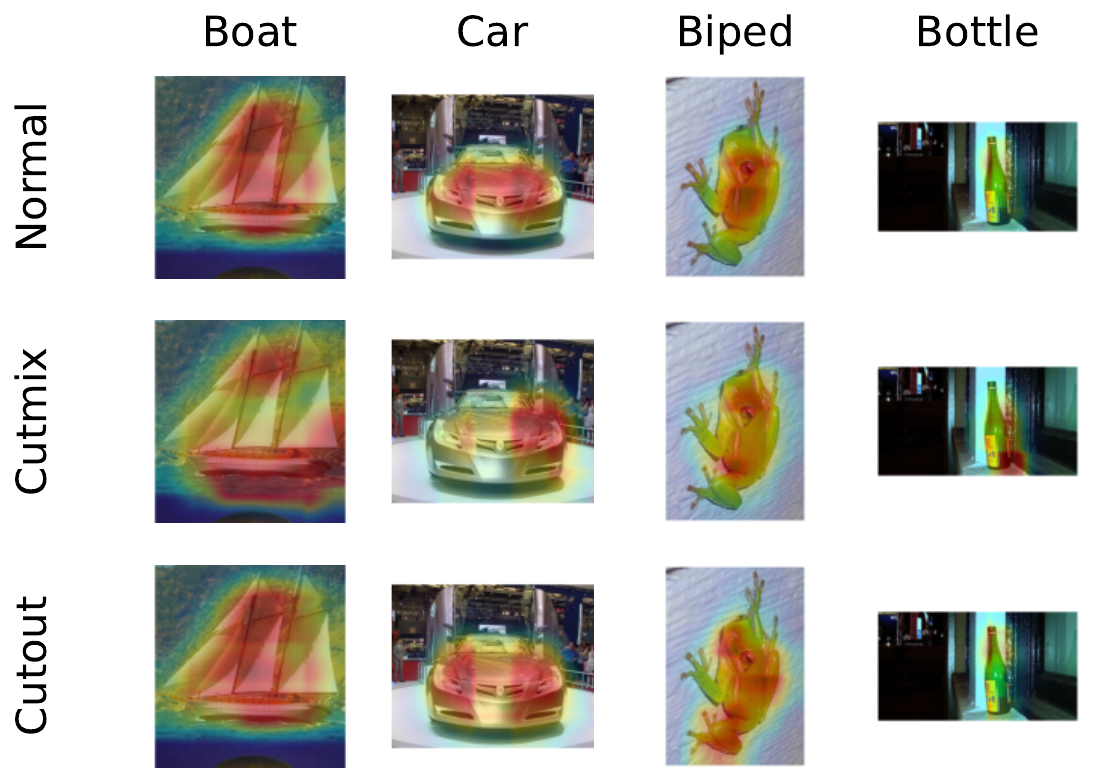}
	\caption{EXP. 2: Visualisation of heatmaps on randomly selected images from the PartImageNet dataset. The backbone network is ResNet-50. The visualisation method is GradCam + SESS.}
	\label{fig:pascal_dataaug}
\end{figure}

\begin{figure}[!tbh]
	\centering
	\includegraphics[width=0.95\linewidth]{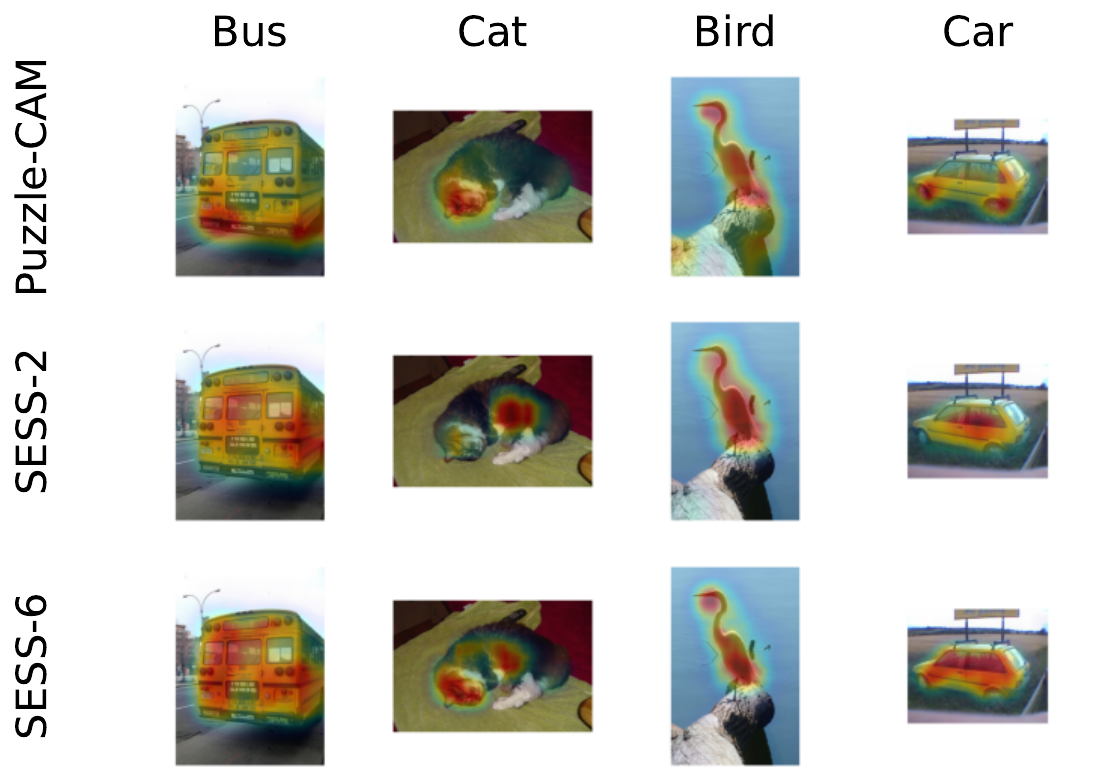}
	\caption{EXP. 3: Visualisation of heatmaps on randomly selected images from the Pascal-Part dataset. The backbone network is ResNet-50. The visualisation method is GradCam + SESS.}
	\label{fig:pascal_enhance}
\end{figure}

\begin{figure}[!tbh]
	\centering
	\includegraphics[width=0.95\linewidth]{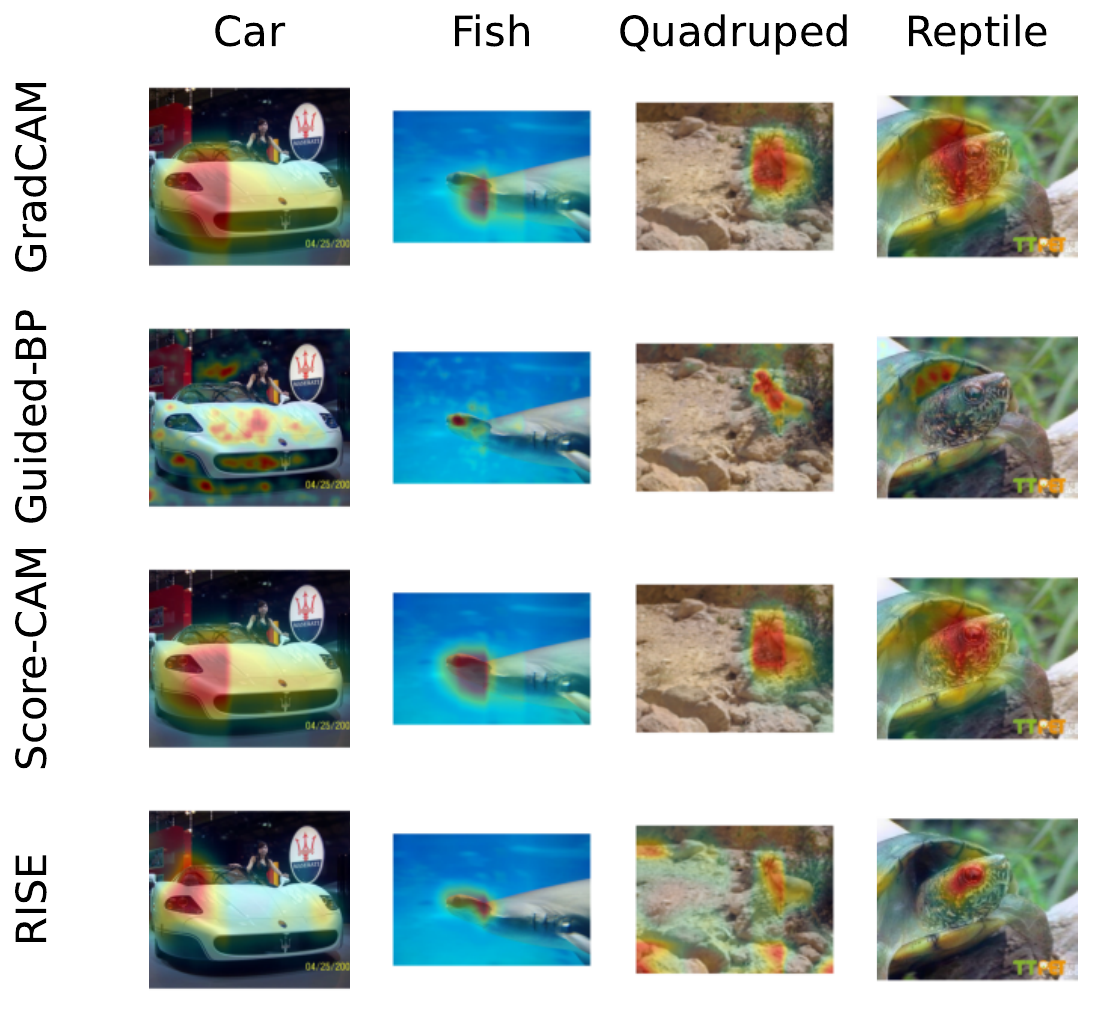}
	\caption{EXP. 4: Visualisation of heatmaps on randomly selected images from the PartImageNet dataset. The backbone network is ResNet-50.}
	\label{fig:partimagenet_heatmap}
\end{figure}

\clearpage
\bibliographystyle{elsarticle-num-names}
\bibliography{egbib}
\end{document}